\documentclass[12pt,a4paper,english,openany]{book}
\usepackage[T1]{fontenc}
\usepackage[utf8]{inputenc}
\usepackage[margin=2.25cm,headheight=26pt,includeheadfoot]{geometry}
\usepackage[english]{babel}
\usepackage{listings}
\usepackage[hyphens]{url}

\usepackage{color}
\usepackage{titlesec}
\usepackage{titling}
\usepackage[framed, numbered]{matlab-prettifier}
\usepackage{changepage}
\usepackage{amsmath}

\usepackage{enumitem}
\usepackage{fancyhdr}
\usepackage{lastpage}
\usepackage{setspace}
\usepackage{multirow}
\usepackage{titling}
\usepackage{float}
\usepackage{comment}
\usepackage{booktabs}
\usepackage{lscape}
\usepackage{booktabs,caption}
\usepackage[flushleft]{threeparttable}
\usepackage[english]{nomencl}
\usepackage{lipsum}
\usepackage{subcaption}
\usepackage{array}
\usepackage[toc,page]{appendix}
\usepackage{dirtytalk}
\usepackage{varioref}

\usepackage{xcolor}
\usepackage{pdfpages}
\usepackage{pifont}
\usepackage{tocloft}
\usepackage{hyperref}
\hypersetup{
    colorlinks=true,
    linkcolor=black,
    filecolor=purple, 
    citecolor=teal,
    urlcolor=cyan,
    pdftitle={Enabling Digitalization in Modular Robotic Systems Integration},
    linkbordercolor=red,
    }

\usepackage{background}

\usepackage[notlof,notlot,nottoc]{tocbibind}

\usepackage{cleveref}
\usepackage{bookmark} 
\usepackage{soul} 
\usepackage{csquotes} 
\usepackage{epigraph} 

\usepackage{framed}

\usepackage{tikz,bm,color}
\usetikzlibrary{shapes,arrows}
\usetikzlibrary{positioning,fit,calc}
\usetikzlibrary{arrows}
\usetikzlibrary{decorations.pathreplacing}
\usepackage{tikz}

\definecolor{orgred}{rgb}{0.8078,0.4471,0.2314}
\definecolor{darkgreen}{rgb}{0.4157,0.6,0.333}
\definecolor{darkblue}{rgb}{0.0,0.0,0.6}
\definecolor{cyan}{rgb}{0.0,0.6,0.6}
\definecolor{light-gray}{gray}{0.80}

\lstdefinestyle{xmlStyle}{
  basicstyle=\ttfamily\scriptsize,
  columns=fullflexible,
  showstringspaces=false,
  commentstyle=\color{darkgreen},
  numbers=left,                    
  numbersep=10pt, 
  numberstyle=\tiny,
  captionpos=b,
}
\backgroundsetup{contents={}}

\definecolor{codegreen}{rgb}{0,0.6,0}
\definecolor{codegray}{rgb}{0.5,0.5,0.5}
\definecolor{codepurple}{rgb}{0.58,0,0.82}
\definecolor{backcolour}{rgb}{0.95,0.95,0.92}
\newcommand\notsotiny{\@setfontsize\notsotiny\@vipt\@viipt}

\definecolor{kpblue}{rgb}{0.21, 0.46, 0.8}
\newcommand*\greencheck{\textcolor{codegreen}{\ding{52}}}
\newcommand*\redcross{\textcolor{red}{\ding{55}}}
\newcommand{\kp}[1]{{\textcolor{kpblue}{\ding{107} \textbf{KP#1:}}}}

\lstdefinestyle{mystyle}{
    language=ocl,
    backgroundcolor=\color{backcolour},   
    commentstyle=\color{codegreen},
    keywordstyle=\color{magenta},
    numberstyle=\tiny\color{codegray},
    stringstyle=\color{codepurple},
    basicstyle=\ttfamily\scriptsize,
    breakatwhitespace=false,         
    breaklines=true,                 
    captionpos=b,                    
    keepspaces=true,                 
    numbers=left,                    
    numbersep=5pt,                  
    showspaces=false,                
    showstringspaces=false,
    showtabs=false,                  
    tabsize=2,
    morekeywords={let,in}
}

\lstdefinestyle{aspstyle}{
    backgroundcolor=\color{backcolour},   
    commentstyle=\color{codegreen},
    keywordstyle=\color{magenta},
    numberstyle=\tiny\color{codegray},
    stringstyle=\color{codepurple},
    basicstyle=\ttfamily\footnotesize,
    breakatwhitespace=false,         
    breaklines=true,                 
    captionpos=b,                    
    keepspaces=true,                 
    numbers=left,                    
    numbersep=5pt,                  
    showspaces=false,                
    showstringspaces=false,
    showtabs=false,                  
    tabsize=2,
    morekeywords={not},
    morecomment=[l]{\%},
}

\usepackage[noredefwarn,acronym,toc,nopostdot]{glossaries}
\makeglossaries

\newacronym{fmu}{FMU}{Functional Mock-Up Unit}
\newacronym{fmi}{FMI}{Functional Mock-Up Interface}
\newacronym{cps}{CPS}{Cyber-Physical System}
\newacronym{oem}{OEM}{Original Equipment Manufacturer}
\newacronym{dt}{DT}{Digital Twin}
\newacronym{ds}{DS}{Digital Shadow}
\newacronym{dm}{DM}{Digital Model}

\newacronym{urdf}{URDF}{Unified Robot Description Format}
\newacronym{sdf}{SDF}{Simulation Description Format}
\newacronym{usd}{USD}{Universal Scene Description}
\newacronym{mjcf}{MJCF}{MuJoCo File}
\newacronym{xml}{XML}{Extensible Markup Language}
\newacronym{xsd}{XSD}{XML Schema Definition Language}
\newacronym{ros}{ROS}{Robot Operating System}
\newacronym{cad}{CAD}{Computer-Aided Design}

\newacronym{trl}{TRL}{Technology Readiness Level}
\newacronym{roi}{ROI}{Return on Investment}
\newacronym{pc}{PC}{Personal Computer}
\newacronym{cpu}{CPU}{Central Processing Unit}
\newacronym{asp}{ASP}{Answer Set Programming}
\newacronym{eecd}{EECD}{End-effector Coupling Device}
\newacronym{cnc}{CNC}{Computer Numerical Control}
\newacronym{ocl}{OCL}{Object Constraint Language}
\newacronym{uml}{UML}{Unified Modeling Language}

\usepackage[backend=biber,bibstyle=ieee,citestyle=numeric,maxbibnames=99]{biblatex} 

\preto\fullcite{\AtNextCite{\defcounter{maxnames}{99}}}

\makeatletter
\newcommand*{\resetnamehighlights}{\let\nhcbx@highlightlist\@empty}
\resetnamehighlights
\newcommand*{\highlightnames}{%
  \@ifstar{\resetnamehighlights\highlightnames@add}{\highlightnames@add}}
\newcommand*{\highlightnames@add@inner}[2]{%
  \listeadd{#1}{\the\numexpr#2\relax}}
\newcommand*{\highlightnames@add}{%
  \forcsvlist{\highlightnames@add@inner\nhcbx@highlightlist}}
\newcommand*{\mkbibhighlightnthname}[1]{%
  \xifinlist{\the\value{listcount}}{\nhcbx@highlightlist}
    {\underline{#1}}
    {#1}}
\makeatother
\DeclareNameWrapperFormat{highlightnthname}{%
  #1}
\DeclareCiteCommand{\fullcite}
  {\usebibmacro{prenote}}
  {\usedriver
     {\DeclareNameWrapperAlias{author}{highlightnthname}%
      \DeclareNameWrapperAlias{editor}{highlightnthname}%
      \DeclareNameWrapperAlias{translator}{highlightnthname}%
      \DeclareNameWrapperAlias{sortname}{highlightnthname}}
     {\thefield{entrytype}}}
  {\multicitedelim}
  {\usebibmacro{postnote}}
  \preto\fullcite{\AtNextCite{\defcounter{maxnames}{99}}}

\renewbibmacro*{cite}{%
  \printtext[bibhyperref]{%
    \printfield{labelprefix}%
    \ifkeyword{publication}{P}{}%
    \printfield{labelnumber}%
    \ifbool{bbx:subentry}
      {\printfield{entrysetcount}}
      {}}}

\defbibenvironment{bibliography}
  {\list
     {\printtext[labelnumberwidth]{%
      \printfield{labelprefix}%
      \ifkeyword{publication}{P}{}%
      \printfield{labelnumber}}}
     {\setlength{\labelwidth}{\labelnumberwidth}%
      \setlength{\leftmargin}{\labelwidth}%
      \setlength{\labelsep}{\biblabelsep}%
      \addtolength{\leftmargin}{\labelsep}%
      \setlength{\itemsep}{\bibitemsep}%
      \setlength{\parsep}{\bibparsep}}%
      }
  {\endlist}
  {\item}

\definecolor{mygreen}{RGB}{112,173,71}
\definecolor{myblue}{RGB}{91,155,213}

\definecolor{acqcolor}{RGB}{197, 90, 17}
\definecolor{intcolor}{RGB}{191, 144, 0}
\definecolor{depcolor}{RGB}{0, 176, 80}

\usepackage{outlines}
\usepackage{tcolorbox}
\newtcolorbox{hypothesis}[1]{colback=myblue!30!white,colframe=myblue!100!black,fonttitle=\bfseries,title=#1}
\newtcolorbox{mybox2}[1]{colback=orange!30!white,colframe=orange!70!white,fonttitle=\bfseries,title=#1}
\newtcolorbox{contr}[1]{colback=mygreen!40!white,colframe=mygreen!100!white,fonttitle=\bfseries,title=#1}

\assignpagestyle{\chapter}{fancy}
\assignpagestyle{\tableofcontents}{fancy}

\title{} 
\makeatletter\let\Title\@title\makeatother

\renewcommand{\headrulewidth}{1pt}
\renewcommand{\footrulewidth}{1pt}

\newlist{underlinedescription}{description}{1}
\setlist[underlinedescription]{style=nextline, font=\normalfont\ul}

\begin{document}
\pagenumbering{roman}
\begin{titlingpage}
\begin{center}
\vspace{2cm}
\includegraphics[width=0.45\textwidth]{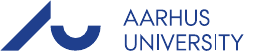}~\\[1cm]
\vspace{3.5cm}

\textsc{\textbf{PhD Dissertation}}\\
\vspace{.5cm}
\centering

\hrule
\vspace{.5cm}
{ \huge \bfseries Enabling Digitalization in Modular Robotic Systems Integration} 
\vspace{.5cm}

\hrule
\vspace{1.5cm}
\textsc{{Author: Daniella Tola}}\\
\vspace{0.3cm}
\textsc{Main Supervisor: Prof. Peter Gorm Larsen}\\
\vspace{0.8cm}
\textsc{Aarhus University}\\
\vspace{0.2cm}
\textsc{Department of Electrical and Computer Engineering}\\
\vspace{0.8cm}
\textsc{31st October 2023}\\

\end{center}
\end{titlingpage}


\chapter*{Abstract}
\addcontentsline{toc}{chapter}{Abstract}

Integrating robot systems into manufacturing lines is a time-consuming process, spanning days, months, or even years.
In the era of digitalization, the research and development of new technologies is crucial for improving integration processes.
Numerous challenges, including the lack of standardization and documentation, as well as intricate stakeholder relationships, complicate the process of robotic systems integration.
This process typically consists of acquisition, integration, and deployment of the robot systems, where it is common that robotic system integrators partially neglect the deployment.
This thesis examines if addressing challenges associated with acquisition and integration advances deployment.

To examine this statement, a mixed-method approach is used through empirical, applied, and inferential methods. 
These were based on surveys and discussions with experts, a literature review, and the development of proof-of-concepts.
This thesis focuses on three areas that help to automate and simplify the robotic systems integration process.

In the first area, related to acquisition, a constraint-based configurator is demonstrated that resolves compatibility challenges between robot devices, and automates the configuration process.
This reduces the risk of integrating incompatible devices and decreases the need for experts during the configuration phase.

In the second area, related to integration, the interoperable modeling format, \acrfull{urdf}, is investigated, where a detailed analysis is performed, revealing significant inconsistencies and critical improvements.
This format is widely used for kinematic modeling and 3D visualization of robots, and its models can be reused across simulation tools.
Improving this format benefits a wide range of users, including robotics engineers, researchers, and students, as the format is used in other industries.

In the third area, related to deployment, \acrfullpl{dt} for robot systems are explored, as these improve efficiency and reduce downtime.
A comprehensive literature review of \acrshortpl{dt} is conducted, and a case study of modular robot systems is developed.
This research can accelerate the adoption of \acrshortpl{dt} in the robotics industry.

These insights and approaches improve the process of robotic systems integration, offering valuable contributions that future research can build upon, ultimately driving efficiency, and reducing costs.

\newpage

\chapter*{Resumé}

\addcontentsline{toc}{chapter}{Resumé}

Integration af robotsystemer i produktionslinjer er en tidskrævende proces, der spænder over dage, måneder, eller endda år.
I digitaliseringsæraen er forskning og udvikling af nye teknologier afgørende for at forbedre integrationsprocesserne.
Adskillige udfordringer, herunder manglen på standardisering og dokumentation, samt indviklede relationer mellem interessenterne, komplicerer integrationsprocessen af robotsystemer.
Denne proces består typisk af anskaffelse, integration, og anvendelse af robotsystemerne, hvor det er almindeligt, at integratorerne af robotsystemerne delvist forsømmer andvendelsen.
Denne afhandling undersøger, om andvendelsen bliver forbedret ved at adressere de udfordringer, der er forbundet med anskaffelse og integration.

For at undersøge dette udsagn er en blandet metodetilgang blevet brugt gennem empiriske, anvendte,
og inferentielle metoder.
Disse var baseret på undersøgelser og diskussioner med eksperter, en litteraturundersøgelse og udvikling af proof-of-concepts. 
Denne afhandling fokuserer på tre områder, der hjælper med at automatisere og forenkle integrationsprocessen for robotsystemer.

Inden for det første område, som er relateret til anskaffelse, bliver der demonstreret en constraint-baseret konfigurator, som løser kompatibilitetsudfordringer mellem robotenheder og automatiserer konfigurationsprocessen. 
Det reducerer risikoen for at integrere inkompatible enheder, og mindsker behovet for eksperter i konfigurationsfasen.

På det andet område, der er relateret til integration, bliver det interoperable modelleringsformat \acrfull{urdf} undersøgt, hvor en detaljeret analyse bliver foretaget, som afslører betydelige uoverensstemmelser og kritiske forbedringer. 
Dette format bruges i vid udstrækning til kinematisk modellering og 3D-visualisering af robotter, og dets modeller kan genbruges på tværs af simuleringsværktøjer. 
Forbedring af dette format gavner en bred vifte af brugere, herunder robotingeniører, forskere og studerende, da formatet bruges i andre brancher. 

I det tredje område, der er relateret til anvendelse, bliver digitale tvillinger (DT'er) til robotsystemer udforsket, da disse forbedrer effektiviteten og reducerer driftsstop. 
Der udføres en omfattende litteraturgennemgang af DT'er, og der bliver udviklet et casestudie af modulære robotsystemer. 
Denne forskning kan fremskynde indførelsen af DT'er i robotindustrien.

Disse indsigter og tilgange forbedrer integrationsprocessen af robotsystemer og giver værdifulde bidrag, som fremtidig forskning kan bygge videre på, som i sidste ende øger effektiviteten og reducerer omkostningerne.

\newpage
\chapter*{Acknowledgements}
\addcontentsline{toc}{chapter}{Acknowledgements}

I am deeply grateful to everyone who has supported me during my PhD journey. 
I would not have been able to achieve this without your help and encouragement.
I would especially like to thank my main supervisor, Peter Gorm Larsen, for his invaluable support and mentorship during both my Masters and PhD, and his dedication to my academic and professional growth.
I am grateful to have had a supervisor that listens to your needs, and encourages you to do your best.
I would also like to thank my co-supervisors, Lukas Esterle for always supporting and being available for discussions, and Cláudio Gomes for inspiring to be meticulous about our work.
Furthermore, I would like to thank all of my collaborators, especially Carl Schultz for being such a great role model and for being so patient.

I would like to thank my family and friends for being so supportive.
Especially my father, Thomas, for teaching me the importance of high standards and hard work, and my dear friend, Helle Degnbol Østergaard, for all the support with many aspects of this journey.
My flatmates, Laura Swift, Rasmus Weile, and especially Poul Kruse-Hansen for the interesting and invaluable discussions, moral support, and uplifting mood.
A special thanks to my colleagues that have become my close and dear friends, especially Fateme Kakavandi, Mirgita Frasheri, and Arian Bakhtiarnia for always supporting me and being there for me.
Thanks to Till Böttjer and Christian Wewer for creating a supportive and collaborative office environment where we can thrive, by discussing our work, but also have time to laugh at dad jokes together.
Most of all, thanks to Mehdi Rafiei for supporting with all of the work-related and personal challenges throughout the whole PhD, and for all of the great experiences we had that made the journey easier.
I appreciate you.

A huge thanks to Peter Corke for welcoming me at Queensland University of Technology and for being such an inspiring role model.
Of the many things I have learned from him are the importance of maintaining high research standards and the value of asking questions.

Lastly, I am grateful to Aarhus University, MADE FAST, and Technicon for the opportunity to pursue my PhD and for their financial support, which made this academic journey possible.

\newpage

\chapter*{Preface}
\addcontentsline{toc}{chapter}{Preface}

This PhD project was conducted from November 1, 2020, to October 31, 2023, in accordance with the 3-year (corresponding to 180 ECTS points) duration defined for Danish PhD projects\footnote{\url{https://international.au.dk/about/organisation/index/6/64/ministerial-order-on-the-phd-degree-programme-at-the-universities-and-certain-higher-artistic-educational-institutions-the-phd-order/}}. 
During this period, the PhD student, as required, completed approximately 30 ECTS points of relevant coursework and contributed 840 hours to teaching and dissemination activities.
As part of the research requirements, the PhD student also undertook a research environment change, spending time at Queensland University of Technology with Distinguished Professor Peter Corke from September to December 2022. 
The main supervisor for this thesis was Professor Peter Gorm Larsen, with co-supervisors Associate Professor Lukas Esterle and Assistant Professor Cláudio Gomes.

This thesis follows the format of a `thesis by publication', comprising \textit{``a collection of scientific articles including a summary accounting for the relation between the publications and their individual contribution to the complete PhD project''}\footnote{\url{https://bss.au.dk/en/research/phd/phd-degree-structure/phd-dissertation}}.

\newpage

\chapter*{Thesis Details}
\addcontentsline{toc}{chapter}{Thesis Details}

\textbf{Thesis Title:} Enabling Digitalization in Modular Robotic Systems Integration

\vspace{0.3cm}
\noindent\textbf{PhD Student:} Daniella Tola

\vspace{0.3cm}
\noindent\textbf{Supervisor:} Professor Peter Gorm Larsen, Aarhus University, Denmark

\vspace{0.5cm}
\noindent The main body of this thesis is based on the following publications:

\begin{itemize}
    \item[\textbf{P1}] \highlightnames*{1}\fullcite{Tola2021}
    \item[\textbf{P2}] \highlightnames*{1}\fullcite{Tola2022}
    \item[\textbf{P3}] \highlightnames*{1}\fullcite{Tola&22b}
    \item[\textbf{P4}] \highlightnames*{1}\fullcite{Tola&2023d}
    \item[\textbf{P5}] \highlightnames*{2}\fullcite{Bottjer2023}
    \item[\textbf{P6}] \highlightnames*{1}\fullcite{Tola&2023b}
    \item[\textbf{P7}] \highlightnames*{1}\fullcite{Tola&2023a}
\end{itemize}

\noindent In addition to the main publications, the following publications that are related to the PhD study have also been made:

\begin{itemize}
    \item[\textbf{A}] \highlightnames*{2}\fullcite{Leporowski&22}
    \item[\textbf{B}] \highlightnames*{2}\fullcite{Legaard2021}
    \item[\textbf{C}] \highlightnames*{2}\fullcite{Madsen2022}
    \item[\textbf{D}] \highlightnames*{3}\fullcite{Schranz2021}
    \item[\textbf{E}] \highlightnames*{3}\fullcite{GithubJens}
    \item[\textbf{F}] \highlightnames*{3}\fullcite{SantiagoGil2023}
    \item[\textbf{G}] \highlightnames*{5}\fullcite{HaoFeng}
\end{itemize}

\noindent Additional tools developed that have not been released yet are:
\begin{itemize}
    \item \acrfull{urdf} analyzer tool\footnote{\url{https://github.com/Daniella1/urdf_analyzer}}.
    \item Modified Denavit-Hartenberg to \acrshort{urdf} converter\footnote{\url{https://github.com/Daniella1/robot_models/tree/main/urdf_converter}}.
\end{itemize}
\clearpage
\hypersetup{
    linkcolor=violet,
    }
\printglossary[type=\acronymtype]
\printglossary
\hypersetup{
    linkcolor=black,
    }

\addtocontents{toc}{\protect{\pdfbookmark[0]{\contentsname}{toc}}}
\tableofcontents
\cleardoublepage
\thispagestyle{fancy}

\newpage
\pagenumbering{arabic}
\renewcommand{\headrulewidth}{1pt}
\renewcommand{\footrulewidth}{1pt}
\definecolor{ref_f}{RGB}{1,121,111} 
\hypersetup{
    linkcolor=ref_f,
    }

\newpage
\chapter{Introduction}

\backgroundsetup{
  scale=1,
  color=black,
  opacity=1,
  angle=90,
  position=current page.north,
  vshift=-250mm,
  hshift=-10mm,
  contents={
    \textcolor{gray}{\rule{5mm}{\paperheight}}
  }
}

\section{Motivation}

The recent pandemic caused by COVID-19 clearly affected the health and daily lives of humans all over the world.
However, its negative impact did not stop there. 
Supply chains were disrupted resulting in increased production time and increased costs of manufacturing companies, due to delayed transportation of components abroad~\cite{SupplyChainDisruptionBacidore}.
Although costly, this lead many companies to move their production lines locally to avoid similar risks in the future.
Reasons, such as high labor costs in Western countries such as Denmark, have resulted in an increasing automation of manufacturing lines~\cite{IFRstats}.
On top of this, the applications being automated are becoming more complex, requiring more innovative designs of the robot systems.
The costs of integrating robots into manufacturing lines is often more expensive than the cost of the hardware itself~\cite{Sannemann&2020}.
The main motivation for this project is to improve modular robotic systems integration by enabling digitalization.

\subsection{Industry Research Funded Through MADE FAST}

This PhD project is part of the research platform called MADE FAST, short for \textit{Manufacturing Academy of Denmark} focusing on \textit{Flexible, Agile, and Sustainable production enabled by Talented employees}\footnote{\url{https://www.made.dk/en/made-fast/}}.
The platform includes stakeholders from academic institutions, research organizations, and industry that collaborate to advance manufacturing technologies and capabilities in Denmark.
One of the objectives of MADE FAST is to improve manufacturing operations using digitalization, aligned with the goals of this PhD project.
This initiative encompasses a range of part projects that have been conducted from 2020 and will complete in 2024.

The MADE FAST platform comprises five strategic initiatives, also called workstreams, where the fourth workstream\footnote{\url{https://www.made.dk/en/made-fast/digitalization-of-processes/}} encompasses this PhD project, and aims at sustainably enabling products and components to be developed faster through digitalization technologies such as \acrfullpl{dt}.
One of the goals of this workstream is to develop models of the manufacturing systems that can be used for simulation and optimization, and combine these into \acrshortpl{dt} or \acrfullpl{ds} that can be used prior to and also during production.
By simulating various designs and parameters, different options can be explored before committing to hardware investments, ultimately leading to a reduction in production expenses.

\subsection{Initial Goal of Project}
\begin{figure}
    \centering
    \includegraphics[width=\columnwidth]{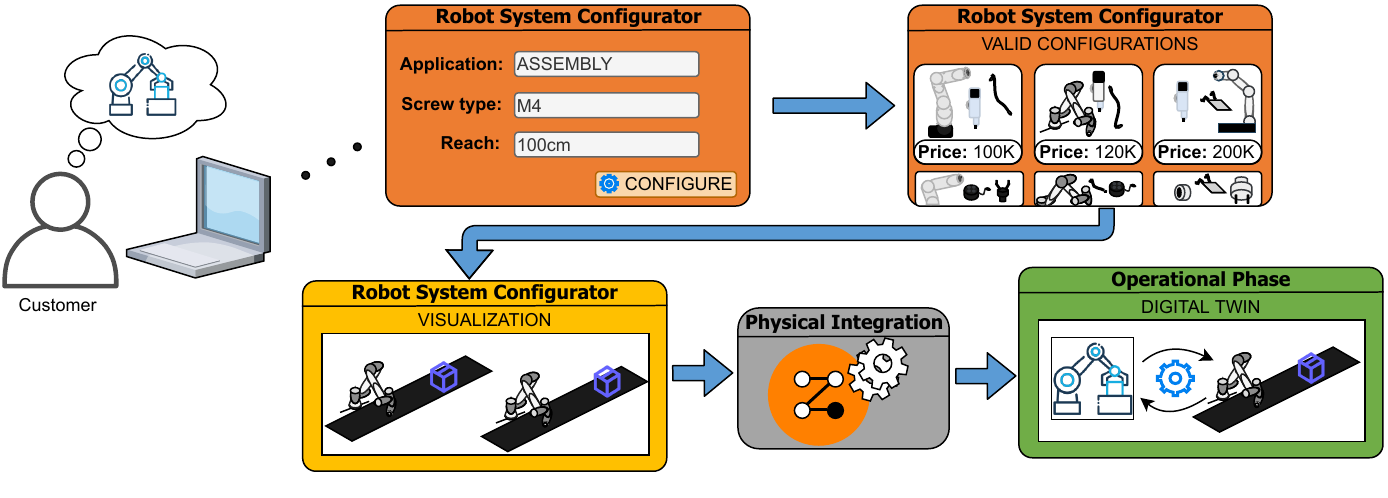}
    \caption{Illustration of the initial goal of the project, showing an e-commerce platform for purchasing modular robot systems, followed by the deployment and maintenance during operation of the purchased system.}
    \label{fig:project_overview}
\end{figure}

\backgroundsetup{
  scale=1,
  color=black,
  opacity=1,
  angle=90,
  position=current page.north,
  vshift=+250mm,
  hshift=-10mm,
  contents={
    \textcolor{gray}{\rule{5mm}{\paperheight}}
  }
}
This PhD project is in collaboration with the Danish robotic system integrator, Technicon\footnote{\url{https://technicon.dk/}}, that has been in the market for 9 years, and carried out more than 400 robotic system integrations. 
The goal of this PhD project was initially defined together with Technicon to tackle real-world problems they have encountered and to disrupt their own integration business.
Their main goal was to develop an e-commerce platform, similar to \textit{vention.io}\footnote{\url{https://vention.io/}}, for digitalizing the process of robotic systems integration, from design to operation, see \cref{fig:project_overview}.
The company desired to create \acrshortpl{dt} of the robot systems, that can be purchased, and use them to improve maintenance.

The collaboration with Technicon during the PhD project was limited, due to the high demand for robotic systems integration at the time, resulting in less employees available for supporting this PhD project.
This PhD project is therefore a combination of academic research and industry related research, with the former being more pronounced.
Moreover, one of the requirements from Technicon was to use the 3D game-engine, Unity\footnote{\url{https://unity.com/}}, to perform visualizations of the robots.

\subsection{Scope}
\begin{figure}
    \centering
    \includegraphics[width=0.6\textwidth]{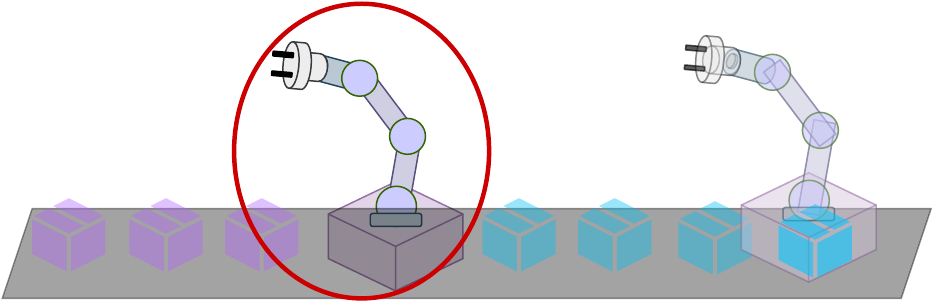}
    \caption{The project scope is limited to single robot systems, as shown with the red oval.}
    \label{fig:scoping}
\end{figure}

This project focuses on improving integration for a single robot system, and does not consider its surroundings in a manufacturing line, see \cref{fig:scoping}.
Such a system can also be defined as a machine at the unit level.
Furthermore, it focuses on creating advances through digitalization efforts, and does not include improving hardware-related aspects.

\section{Research Questions}
\backgroundsetup{contents={}}
We have specified a main hypothesis for this PhD project and Research Questions that aim at assisting to confirm or disprove this.
The main hypothesis is:

\begin{hypothesis}{Hypothesis}
    Solving relevant barriers related to the acquisition and integration of modular robot systems in the industry advances the deployment of such systems.
\end{hypothesis}

One of the major issues within robotic systems integration is that robotic system integrators mainly focus on the acquisition and integration of modular robot systems, and often neglect parts of the deployment, resulting in a number of complications~\cite{Sannemann&2020}.
Therefore, we propose a hypothesis where we can investigate if the advancement of the acquisition and integration can lead to advancements in deployment.

This thesis addresses the hypothesis by creating and advancing digitalization efforts in modular robotic systems integration.
The overall technical process of robotic systems integration consists of four main phases which are: design and configuration, visualization and simulation, physical integration, and an operational phase.
These phases and their association with acquisition, integration, and deployment are shown in \cref{fig:chapters_overview}, where acquisition entails design, configuration, and visualization; integration involves visualization, simulation, and physical integration of the system; and deployment includes physical integration and operation of the system.
We distinguish between \textit{robotic systems integration} and \textit{integration}, where robotic systems integration involves the complete process from acquisition to deployment (all phases in \cref{fig:chapters_overview}), while integration itself involves visualization, simulation, and physical integration of the system (phases~2 and~3 in \cref{fig:chapters_overview}).

The Research Questions are defined to investigate different parts of the robotic systems integration process through digitalization, with an exception of the physical integration phase, as there is limited digitalization possibilities within that phase.
To reach to a conclusion on the hypothesis, the following Research Questions are defined:

\begin{mybox2}{Research Question 1 (RQ1):} 
\textit{What are the main barriers of acquiring and integrating modular robot systems, and how can they be addressed?}
\end{mybox2}

The first Research Question (\textbf{RQ1}) is exploratory~\cite{kothari2004research} and involves understanding the process of robotic systems integration and identifying its associated challenges.
This RQ is used as the preliminary investigation for the hypothesis, where the barriers that need to be solved are identified.
A number of configuration-related challenges were identified in this RQ, which lead to defining \textbf{RQ2}.
\textbf{RQ1} is highly related to industry-specific information and research.

\begin{mybox2}{Research Question 2 (RQ2):}
\textit{Is it possible to define a configurator of modular robot systems that takes application requirements and compatibility of devices into account?}
\end{mybox2}

\textbf{RQ2} is applied~\cite{kothari2004research} as it aims at finding a solution to digitalize the process of configuring modular robot systems.
The proof-of-concept should find a feasible method to create a configurator that addresses the configuration-related challenges found in \textbf{RQ1}.
This question focuses on the design and configuration phase of robotic systems integration, and tackles the acquisition aspects of the hypothesis.
\textbf{RQ2} is based on industrial challenges, though it is solved through academic research.

\begin{mybox2}{Research Question 3 (RQ3):}
\textit{What are the advantages and complications of defining an interoperable robot modeling format for visualization and simulation of robots?}
\end{mybox2}
\textbf{RQ3} is descriptive and aims to gain a deeper understanding of the advantages and challenges that follow when using an interoperable  modeling format.
This RQ was derived from challenges related to vendor lock-in that were found in \textbf{RQ1}.
To answer this RQ, fundamental knowledge on such a format should be surveyed, together with its usability, limitations, and potential improvements.
Understanding and improving such a format is beneficial for robotic systems integration when visualizing or simulating the robot systems, especially since models created using such a format can be reused across different phases of the process.
This RQ tackles the integration aspects of the hypothesis.
\textbf{RQ3} is highly related to academic research, though such a format would be used in the industry.

\begin{mybox2}{Research Question 4 (RQ4):}
\textit{Is it possible to produce \acrfullpl{dt} of modular robot systems?}
\end{mybox2}
\textbf{RQ4} is a combination of exploratory and applied research, as the concept of \acrshortpl{dt} is fairly new and the research can provide fundamental knowledge in this area.
Application-wise, the research can be used to find potential techniques to create \acrshortpl{dt} of modular robot systems, making this RQ addressable through a combination of fundamental and applied methods.
This RQ is strongly related to the operational phase of robotic systems integration for improving quality control and maintenance of the systems.
This RQ tackles the deployment aspects of the hypothesis.
\textbf{RQ4} is highly related to academic research, as the concept of \acrshortpl{dt} is not yet fully industrialized within manufacturing processes.

\begin{figure}
    \centering
    \includegraphics[width=0.65\textwidth]{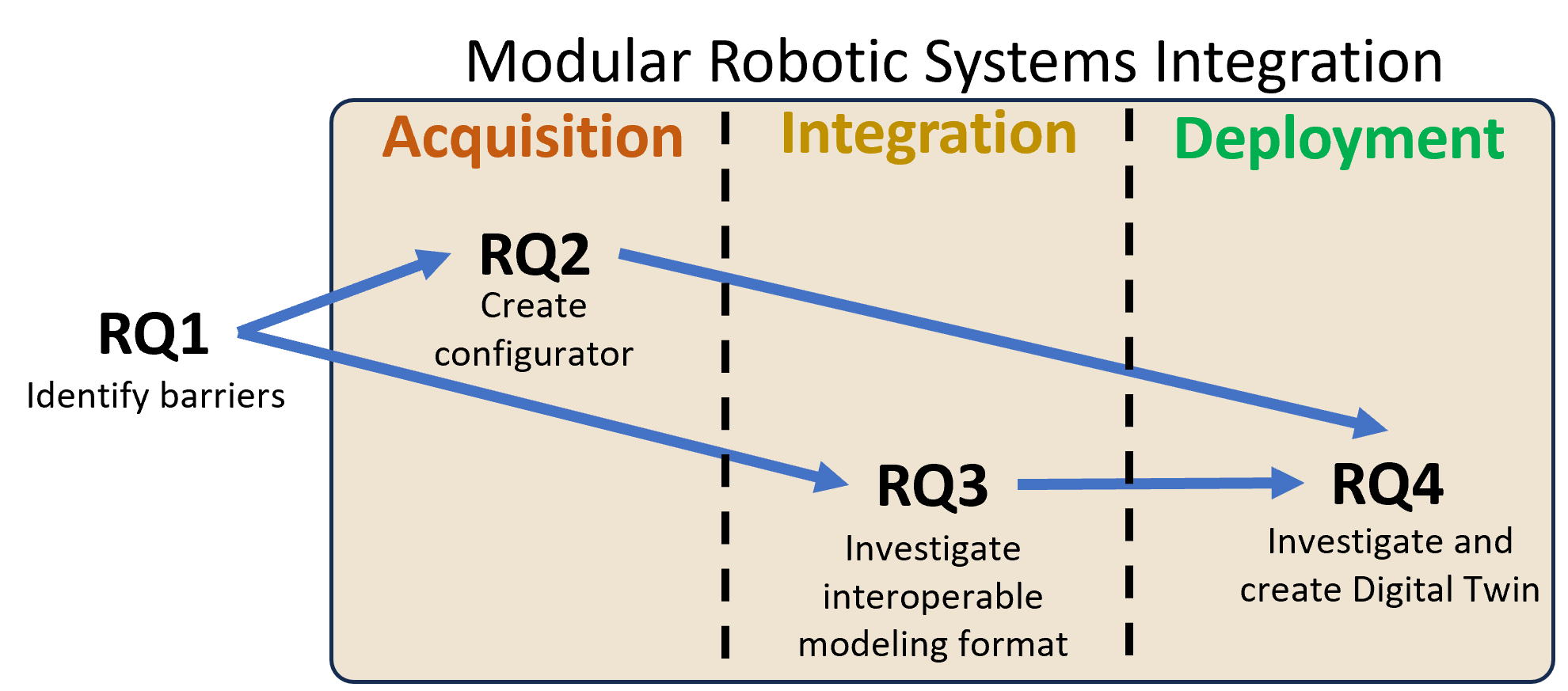}
    \caption{The connections assumed between the output of the RQs (illustrated by the solid arrows) and how this can be used in the addressing of other RQs. The RQs are mapped to the part of modular robotic systems integration that they address.}
    \label{fig:rqs_relation}
\end{figure}

\Cref{fig:rqs_relation} illustrates the assumed relations between the RQs and highlights how outputs of one RQ feeds into additional RQs.
\textbf{RQ1} is the preliminary basis for identifying relevant barriers, that based on the hypothesis, \textbf{RQ2} and \textbf{RQ3} can tackle.
The hypothesis assumes, as the arrows in \cref{fig:rqs_relation} show, that there is a direct relationship between the output from \textbf{RQ2} and \textbf{RQ3} to \textbf{RQ4}.

\textbf{RQ1} and \textbf{RQ4} are both chosen as \textit{exploratory} questions, as these two areas, of robotic systems integration and \acrshortpl{dt} of robot systems, have not been studied in depth before and therefore the most suitable research approach is to achieve new insights.
Conducting \textit{applied} research in \textbf{RQ2} is beneficial, as this PhD project was initially defined in collaboration with industry, and this question can be used to propose a technique to solve some of the identified problems from \textbf{RQ1}.
Understanding the capabilities and limitations of an interoperable robot description format is essential for developers and researchers to build upon and tackle these issues in the future, which is the reason we chose \textbf{RQ3} to be \textit{descriptive}.

\section{Research Contributions} \label{sec:intro_contributions}
The main contributions of this PhD project are presented below, and will be described in more detail in \cref{chapter:conclusion}.
Furthermore, the research contributions are presented in each relevant chapter following their related text.
\begin{itemize}
    \item[\textbf{C1:}] Characterized the main stakeholders involved in robotic systems integration and detailed the process.

    \item[\textbf{C2:}] Outlined current challenges in acquiring and integrating modular robot systems and presented directions to tackle them.

    \item[\textbf{C3:}] Proposed a formal approach for constraint-based configuration of robot systems to address a subset of the identified challenges. 

    \item[\textbf{C4:}] Identified the usage, current advantages, and challenges of \acrfull{urdf} through a user survey.

    \item[\textbf{C5:}] Created a dataset of \acrshort{urdf} files and highlighted significant discrepancies in the format and associated tools, and provided common guidelines on developing \acrshortpl{urdf}.

    \item[\textbf{C6:}] Classified the key components of \acrshort{dt} approaches in manufacturing based on unit level processes, identified critical barriers that complicate the development of such \acrshortpl{dt}, and highlighted future opportunities.

    \item[\textbf{C7:}] Demonstrated an approach for creating modular \acrshortpl{ds} of robot systems.
\end{itemize}


\section{Research Methods}

The research in this PhD project has been conducted in an iterative manner, where research problems were determined both through industrial needs (from Technicon) and through discoveries while researching these problems.

As the research conducted in this PhD project has various directions and techniques, the methodologies used are also different across research topics.
The research methods utilized span across the following~\cite{kothari2004research}:
\begin{itemize}
    \item Quantitative and qualitative data analysis in the publications~\cite{Tola&2023d,Tola&2023b} related to contributions \textbf{C4} and \textbf{C5}.
    \item Applied research in publications~\cite{Tola2021,Tola&22b}, where proof-of-concepts were developed, related to contributions \textbf{C3} and \textbf{C7}.
    \item Descriptive research in the \acrshort{dt} literature review~\cite{Bottjer2023}, related to contribution \textbf{C6}.
    \item Empirical research was conducted in the publications~\cite{Tola2022,Tola&2023a}, where evidence was gathered through empirical studies and exchanging knowledge with relevant stakeholders, related to contributions \textbf{C1} and \textbf{C2}.
\end{itemize}
The detailed research methodologies applied to each research topic and contributions are explained in each chapter below.

\section{Assessment of Thesis} \label{sec:intro_assessment}

A self-assessment of this thesis is provided in \cref{chapter:conclusion}, by evaluating the research conducted to address the Research Questions, and evaluating the research contributions. 
This assists in better evaluating the successful areas and deficiencies of this thesis.
Objectively evaluating how a thesis advances the state-of-the-art in research is difficult, as research is very exploratory where we are constantly trying to gain more knowledge in our research fields.
Although performing a self-assessment is subjective, it enables a form for evaluation of the contributions of this thesis.
The contributions are ranked based on the following three criteria, and are described in more detail in \cref{sec:assessment_contributions}:
\begin{itemize}
    \item Advancement of Acquiring Robot Systems
    \item Advancement of Integrating Robot Systems
    \item Advancement of Deploying Robot Systems
    \item Reducing Vendor-Lock-in
\end{itemize}

As the hypothesis of this thesis looks at the advancement of acquiring, integrating, and deploying modular robot systems, we evaluate how each contribution advances each of these areas.
Furthermore, an inherent issue found in robotics is vendor lock-in that affects robotic systems integration in various ways.
For instance, vendor lock-in increases the risk of integration, as integrators become dependent on the specific vendors' products, restricting the integrators to these.
Moreover, if not dependent on a specific vendor, robot models for visualization and simulation can be reused throughout the different phases of design and configuration, virtual commissioning (which is out of the scope of this thesis), and the operational phase in the context of \acrshortpl{dt}.
These models can be reused in engineering tasks and for providing customers with a better understanding of the robot systems.
Therefore, evaluating how each contribution reduces vendor lock-in is crucial in this thesis.

\section{Outline and Reading Guide}

\begin{figure}
    \centering
    \includegraphics[width=0.8\textwidth]{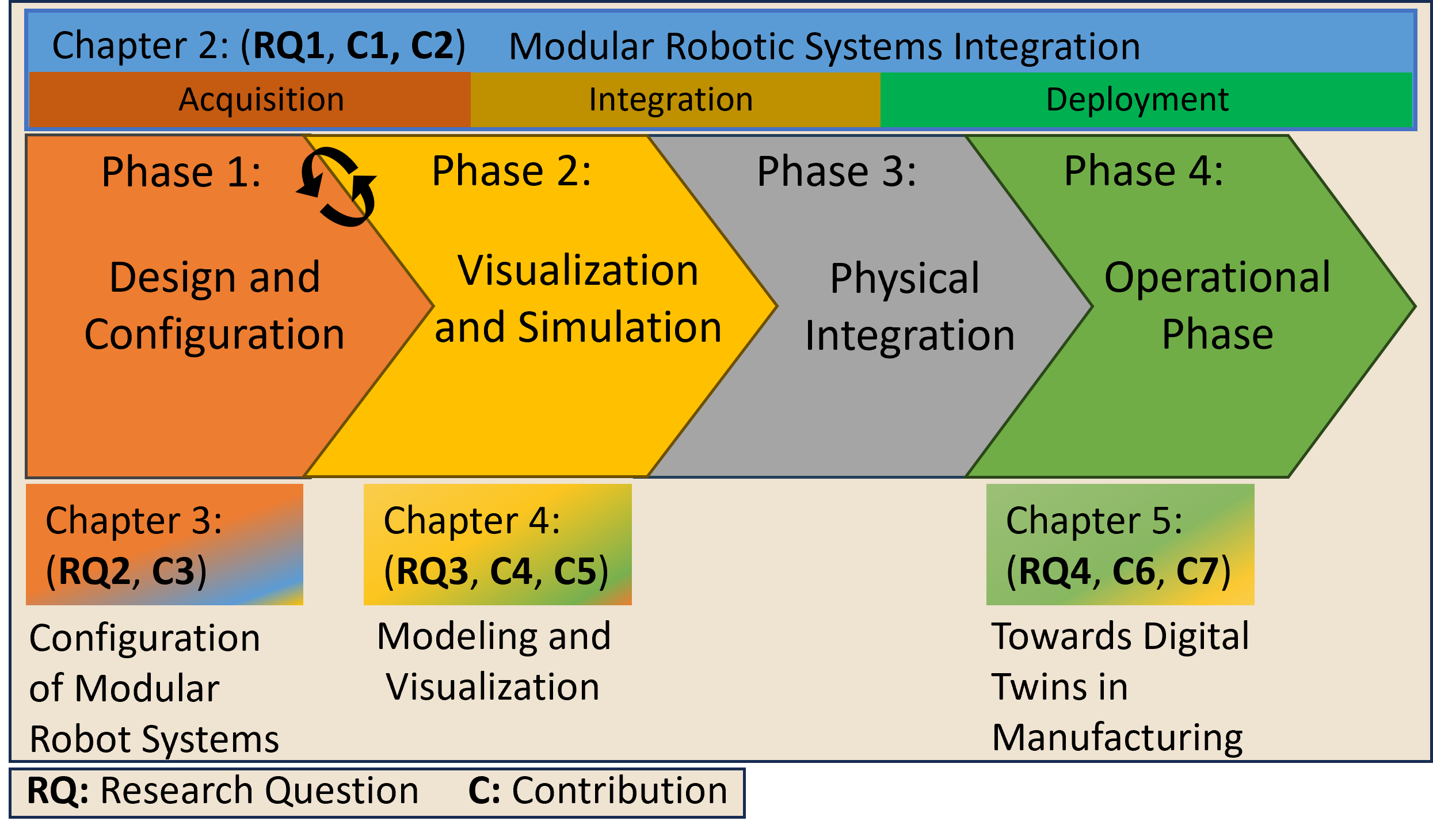}
    \caption{Overview of the technical process of modular robotic systems integration, entailing acquisition (phase~1 and~2), integration (phase~2 and~3), and deployment (phase~3 and~4).
     The chapters in this thesis are presented with their associated Research Questions and contributions. The blend of colors in the chapters shows the colors of the parts of the process that the chapters cover. The arrows between phase~1 and~2 indicate an iterative process between these phases.}
    \label{fig:chapters_overview}
\end{figure}

This PhD thesis is written as a `thesis by publication' consisting of both published and unpublished work conducted during the PhD project.
The following chapters tackle specific Research Questions and present contributions, as illustrated in \cref{fig:chapters_overview}, where:
\begin{description}
    \item[Chapter 2:] describes the process of modular robotic systems integration, the main stakeholders involved, and highlights some of the main challenges found. \textbf{RQ1} is addressed in this chapter.

    \item[Chapter 3:] clarifies the current difficulties of creating a configurator for modular robot systems, and demonstrates an approach using constraint-programming for tackling these challenges. This chapter is related to the acquisition of modular robot systems, where \textbf{RQ2} is addressed.

    \item[Chapter 4:] introduces robot description formats for modeling robots, and details the \acrfull{urdf} and its current shortcomings to achieve interoperability across simulation tools. This chapter is related to visualization and simulation within the integration of modular robot systems, where \textbf{RQ3} is addressed.

    \item[Chapter 5:] presents the concept of \acrshortpl{dt}, where a comprehensive literature review is conducted together with a method to implement \acrshortpl{ds} for modular robot systems. This chapter is related to the deployment of modular robot systems, where \textbf{RQ4} is addressed.
                    
\end{description}

\begin{figure}
    \centering
    \includegraphics[width=0.6\textwidth]{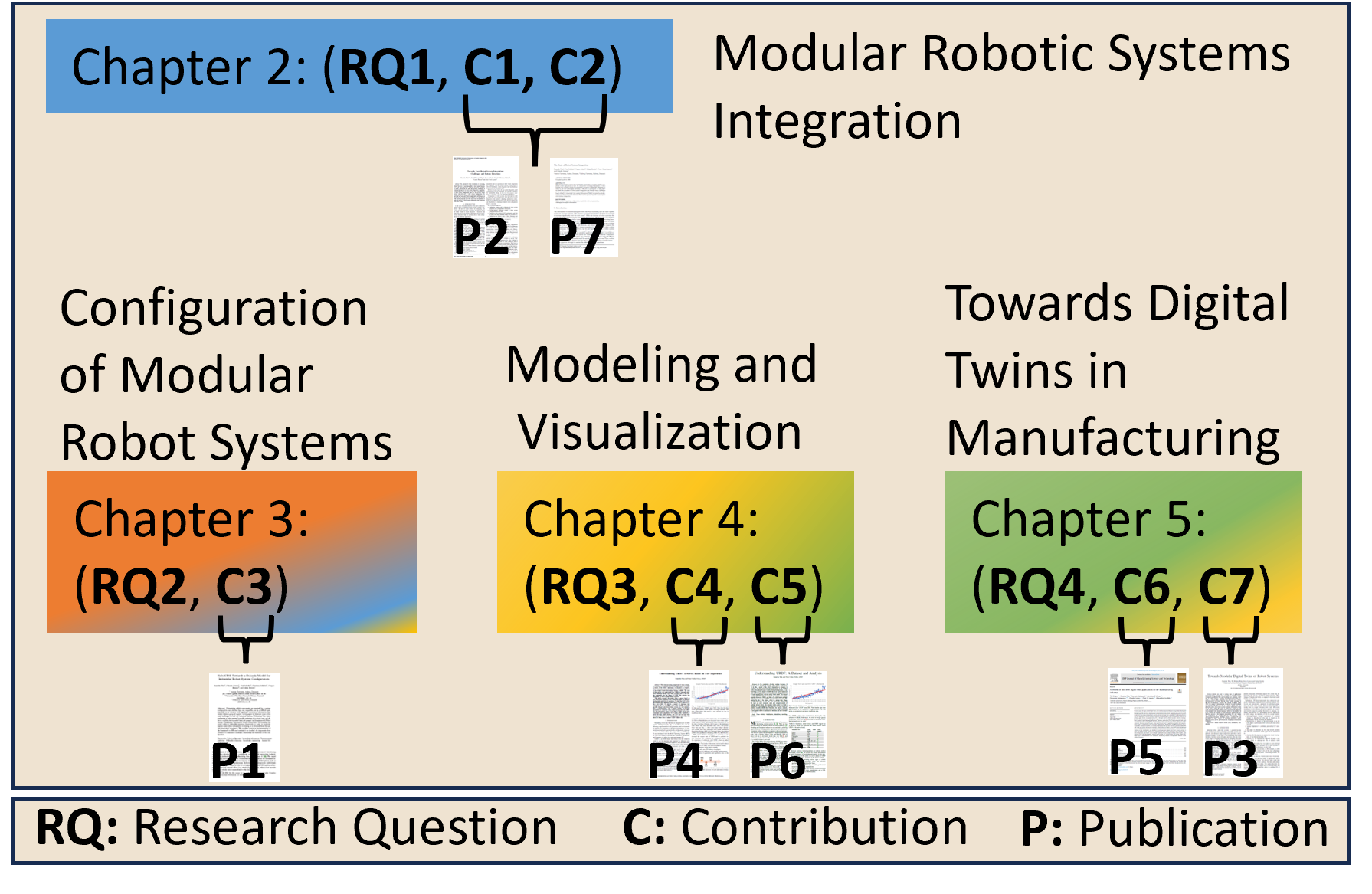}
    \caption{Overview of the chapters, Research Questions, contributions, and the publications related to each.}
    \label{fig:chapters_publications}
\end{figure}

Finally, chapter~6 briefly lists the contributions of the thesis and how they relate to the Research Questions, followed by an assessment of the thesis, and potential future directions.
Furthermore, the publications that each contribution is built upon are shown in \cref{fig:chapters_publications}.

The traditional process carried out by robotic system integrators does not necessarily involve visualization and simulation (phase 2) or maintenance during the system's operational phase (phase 4), in contrast to the illustration in \cref{fig:chapters_overview}.
Furthermore, the conventional process typically includes the programming of the system during the physical integration phase.
The illustration in \cref{fig:chapters_overview} represents our ideal vision of the process of robotic systems integration, where parts of the system are verified through simulations before investing in the hardware, and where the operational phase of the system is also a part of the ``package'' when investing in a robot system.
Furthermore, phases~1 and~2 can be iterated between, as once a design is created, it can be simulated, and the results may lead to a re-design.
The programming of the robot systems would in this case be performed during integration (phase~2 and~3), where simulations of the programmed system are performed, followed by an implementation on the hardware itself.

\chapter{Modular Robotic Systems Integration}

\backgroundsetup{
  scale=1,
  color=black,
  opacity=1,
  angle=90,
  position=current page.north,
  vshift=-250mm,
  hshift=-10mm,
  contents={
    \textcolor{gray}{\rule{5mm}{\paperheight}}
  }
}

To advance the state-of-the-art of robotic systems integration, it is necessary to understand fundamental concepts related to it, such as robot systems and the stakeholders involved.
These are all introduced in this chapter, together with a number of related challenges and potential directions.
This chapter is based on the research output from publications~\cite{Tola2022} and~\cite{Tola&2023a}.

\section{Research Method}
The goal of this chapter is to answer \textbf{RQ1} which is exploratory and defined as: \textit{what are the main barriers of acquiring and integrating modular robot systems, and how can they be addressed?}

As stated in~\cite{NadiaSystemIntegration}, research in robotic systems integration focuses on the hardware and software itself, instead of researching the complete integration process.
This supports the proposition that research in this area is very limited.
Therefore, a combination of results gathered through different research methods that can support each other is the most appropriate.
The research produced in this chapter is a combination of information from empirical studies, discussions with experts (robotic system integrators and research organizations), and research articles.
To address \textbf{RQ1}, we study the process of robotic systems integration, along with the stakeholders and their relationships.
We identify the challenges inherent in this process and present potential methods and directions to address them.

\section{Modular Robot Systems} \label{sec:modular_robot_systems}
        \subsection{Overview}

\begin{figure}
                \centering
                \includegraphics[width=\textwidth]{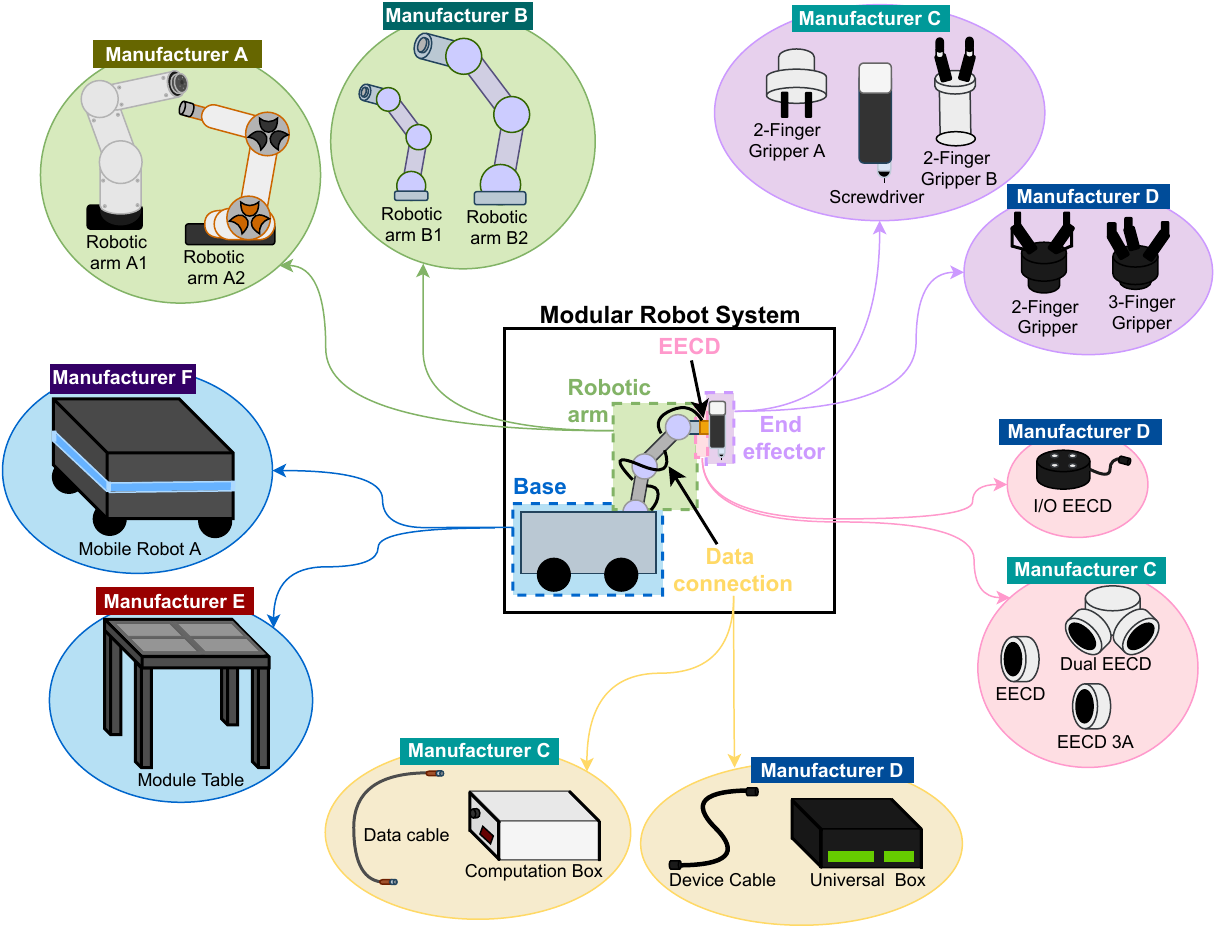}
                \caption{Example of a simple modular robot system in the middle, which can be composed of devices from different manufacturers. The real names of the manufacturers are withheld to avoid harming their reputations, as some of the challenges presented below are based on the lack of documentation by some of them.}
                \label{fig:robot_system}
            \end{figure}

            A robot system is defined by the ISO 8373 standard~\cite{ISO8373} as a system consisting of one or more robots, end-effectors, and other devices used to support the robot in performing a task.
            The main devices that shape a robot system are shown in \cref{fig:robot_system}, and are briefly presented below with more details explained in \cref{ch2:devices}.
            In addition to these devices, equipment such as vision systems, adhesive dispensers, and weld controllers may be part of a robot system if they are vital to fulfill a task.

            \begin{description}
                \item[Robotic arm:] also called a manipulator, is shown in \cref{fig:robot_system} in the green shapes.
                                    This device is the main device in a robot system, used for manipulation, and can be configured to complete different tasks, such as moving objects, polishing, assembling, and welding.

                 \item[End-effector Coupling Device (EECD):] is shown in \cref{fig:robot_system} in the pink shapes.
                                    This device is connected to the tip of the robotic arm, and is used for easily connecting and exchanging different end-effectors onto the robotic arm.
                
                \item[End-effector:] is shown in \cref{fig:robot_system} in the purple shapes.
                                    This device is attached to the tip of the robotic arm through an \acrshort{eecd} and allows the robot to perform the given task.
                                    Different types of end-effectors exist, for example, screwdrivers can be used for assembly and grippers can be used for pick-and-place applications, which are described in more detail below.                                                        

                \item[Data connection:] is shown in \cref{fig:robot_system} in the yellow shapes.
                                    It enables the robotic arm to communicate with the end-effector through a cable or control box.
                                    The control box enables additional features, such as signal processing.
                                    Typically, the \acrshort{eecd}, end-effector, and data connection are all developed by the same manufacturer.
                
                \item[Base:] is shown in \cref{fig:robot_system} in the blue shapes.
                            Robotic arms are typically mounted onto a base, which can be a mobile robot or a stationary module (with or without additional sensors installed).
                            Robotic arms can also be mounted on a wall or on a ceiling.
            \end{description}

            \backgroundsetup{
              scale=1,
              color=black,
              opacity=1,
              angle=90,
              position=current page.north,
              vshift=+250mm,
              hshift=-10mm,
              contents={
                \textcolor{gray}{\rule{5mm}{\paperheight}}
              }
            }

            Such devices can be connected together to create a robot system capable of performing specific tasks.
            ISO 8373~\cite{ISO8373} defines modularity as \textit{``a set of characteristics which allow systems to be separated into discrete modules and recombined''}.
            In this thesis, we refer to such a robot system, comprising commercial off-the-shelf devices, as a \textit{modular robot system}.
            This designation is based on the fact that it consists of modules that can be assembled and ``easily exchanged'', adhering to the definition of \textit{modularity}.
            Modular robot systems enable faster deployment time, resulting in cheaper automation~\cite{RoleOfSystemIntegrators}.
            To create a modular robot system, the devices put together need to be compatible, interoperable, and fulfill the task and requirements of the robot system.

        \subsection{Application Examples}
        This section describes two common applications of robot systems.
        
        \subsubsection{Pick-and-Place}
        Pick-and-place applications, as the name indicates, are applications where the robotic arm moves objects from one place to another.
        Such applications can be targeted at packaging items into boxes, sorting through items on a conveyor, palletizing, or machine tending.
        Palletizing entails placing objects onto a pallet. 
        Machine tending consists of a robotic arm that places an object into a machine, for instance, a CNC machine, and takes out the object after the machine has operated on (tended) the object.
        
        \subsubsection{Screwdriving}
        Automated screwdriving applications consist of a robotic arm, in some cases also a screw-feeder, from which the robotic arm can pick up screws, move them to the intended location on an object, and tighten the screw.
        This can be used for (dis-)assembling products.
        
        \subsection{Robot Devices} \label{ch2:devices}
        This section provides more details on the devices within a modular robot system.
        \backgroundsetup{contents={}}
        \subsubsection{Robotic Arms}

\begin{figure}
        \centering
        \begin{minipage}{.5\textwidth}
          \centering
          \includegraphics[width=0.65\textwidth]{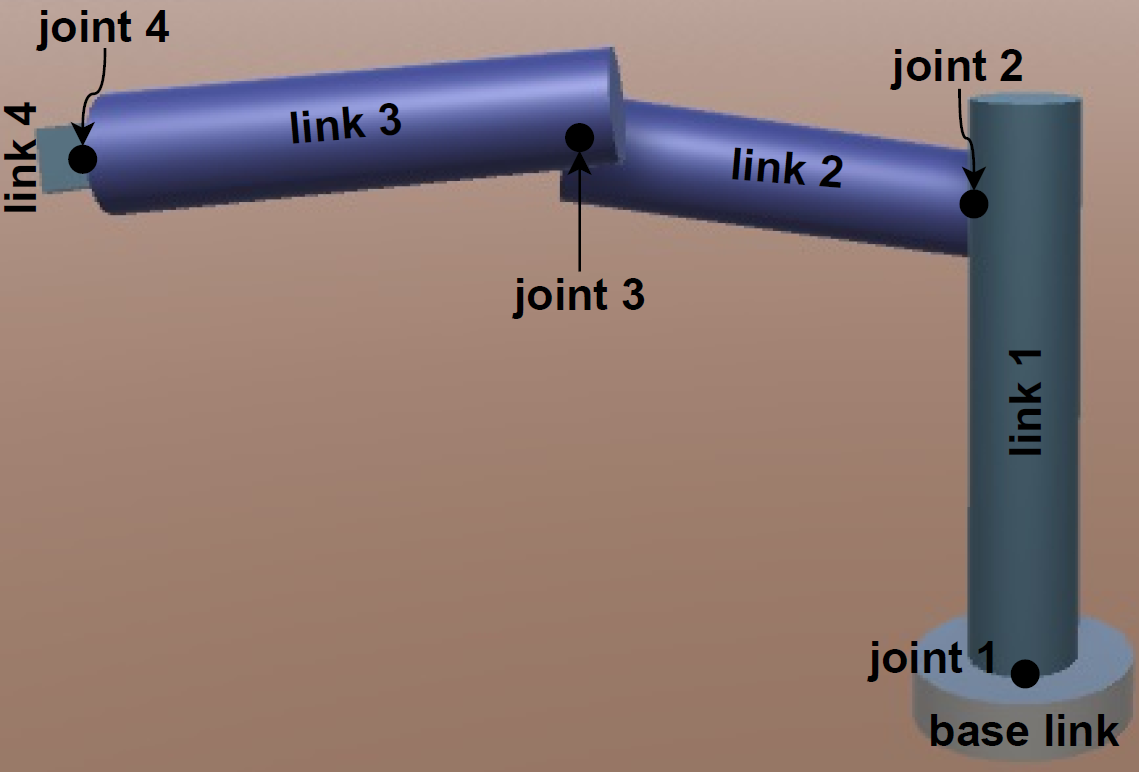}
          \captionof{figure}{Example of robotic arm with annotated joints and links. This figure is recreated from arXiv~\cite{tola2023understanding}.}
          \label{fig:cylindrical_vs_kuka_annotated_all}
        \end{minipage}
        \hfill
        \begin{minipage}{.45\textwidth}
          \centering
          \includegraphics[width=0.65\textwidth]{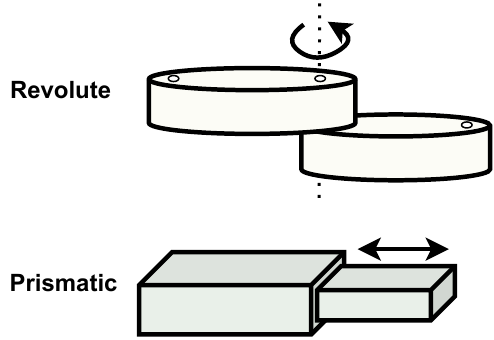}
          \captionof{figure}{Motion illustration of revolute and prismatic joints.}
          \label{fig:joint_types}
        \end{minipage}
        \end{figure}

        The most commonly used type of robotic arms in manufacturing are serial manipulators.
        Such manipulators consist of a series of \textit{links}, typically rigid bodies, attached together and actuated through motors~\cite{ISO8373} at the \textit{joints} (see \cref{fig:cylindrical_vs_kuka_annotated_all}).
        Different joint types exist, with the most common being revolute and prismatic.
        Revolute joints connect two links and rotate about a fixed axis, while prismatic joints, also known as sliding joints, allow a linear motion between the links (see \cref{fig:joint_types}).
        There are two distinct links in a robotic arm that are connected to other devices.
        The base link is the bottom of the robotic arm, which connects the robot onto a base.
        The top link of the robotic arm is the tip, and its mechanical interface is called the robot flange.
        Links and joints are explained in more detail in \cref{chapter:visualization}.
        The mechanical interface of the robot flange is standardized through the ISO 9409-1 standard~\cite{ISO9409}, which is used when attaching \acrshortpl{eecd} or other devices to the tip of the robotic arm.
        A robotic arm typically comes with a controller and a teach pendant to enable programming of the robot. 
        The \textit{payload} of a robotic arm describes the weight that can be carried by the arm,  used in pick-and-place applications.

        Both industrial and collaborative robotic arms are used in manufacturing processes.
        Industrial robots are intended to entirely replace workers and can also operate on heavy-duty manufacturing tasks, while collaborative robots, also called cobots, are developed to work together with or in proximity of humans in a safe manner.

        \subsubsection{End-Effector and Associated Devices}
        The end-effector is mechanically connected to the robotic arm via an \acrshort{eecd} and communicates through the data connection.
        \acrshortpl{eecd} are designed to make the process of changing end-effectors easier; for example, one of the \acrshortpl{eecd} available on the market allows changing the end-effector without the need to screw anything.
        This is crucial in applications, where the robotic arm needs to use multiple end-effectors to execute different parts of the task.

        There are different categories of end-effectors and subcategories of them.
        Grippers can be used to move objects from a start to an end position.
        Depending on the object's size and texture, a specific type of gripper can be chosen.
        As an example, for palletizing hard, rectangular objects, vacuum grippers can be used, while electrical grippers may be used for palletizing objects with softer materials.
        Screwdrivers can be used to assemble or disassemble parts of an object.
        These are chosen based on the required size of the screws, the speed, and the torque.
        Flange adapters are simple mechanical devices that interface between an \acrshort{eecd} and a robotic arm with different ISO flange types.
        These devices are not shown in \cref{fig:robot_system} because they are exclusively used in cases where the flanges are incompatible.

        \subsubsection{Bases}
        Depending on the application requirements and the chosen robotic arm, different types of bases can be used.
        If the arm needs to operate within a larger area than its reach allows, it would be appropriate to mount it on a mobile robot to move throughout the designated area.
        Subsystems that combine robotic arms and mobile robots are called mobile manipulators.
        Such subsystems may reduce integration time, as the two devices are already configured and the programming interfaces are designed for controlling both robots.
        Stationary bases also exist, and can be used in cases where a robotic arm can complete the task within its own reach.
        Such bases may have safety sensors that register if a human is in the vicinity of the (industrial) robot, pausing the robot, until the safety hazard is removed.
      
        \subsubsection{Robot Application Solutions}
        Some robot applications are popular, and therefore complete solutions that can easily be deployed within just a few days have been developed by different manufacturers.
        With such solutions, it is important to check the requirements of a system and compare them against the specifications of the application solution~\cite{BlogLongRobotDeploymentReallyTake}.

\section{Robotic Systems Integration}
This section is mainly based on results from~\cite{Tola&2023a}, and parts of this section are taken directly from the publication which is not submitted yet.
        \subsection{Overview}
        The terms \textit{acquisition}, \textit{integration}, and \textit{deployment} of robot systems are not universally standardized definitions, but are commonly used within the field of robotics and automation.
        In this thesis, \textit{acquisition} is defined as the process of designing, configuring, and visualizing devices.
        \textit{Integration} is defined as the process of visualizing, simulating, and physically integrating the devices.
        It involves physically connecting the hardware and validating the system functionality as a whole.
        \textit{Deployment} is defined as the process of moving the physical system to its intended environment, for instance, to a manufacturing plant, and verifying the operation and safety of the system, in addition to ensuring that it is being optimally used to fulfill the given task.
        Ongoing maintenance of the system during operation is also part of this process~\cite{robotiqBlogDeploymentIntegration}.
        These definitions associate the term \textit{acquisition} with phases~1 and ~2, \textit{integration} with phases~2 and~~3, and the term \textit{deployment} with phases~3 and~4, illustrated in \cref{fig:chapters_overview}.
        The figure provides an overview of the main technical steps involved in robotic systems integration.

        It is worth noting that while the process outlined in \cref{fig:chapters_overview} is often sequentially followed, the process can also be iterative.
        When developers gain new insights from simulation results or after physically integrating the system, they may re-iterate between the designs of certain parts of the system. 
        Another significant point is that not necessarily all companies or all projects follow this process.
        In many cases, the visualization or simulation phase may be entirely omitted depending on the project type and size.
        In the automation world, the visualization and simulation phase may be part of the virtual commissioning process.
        This entails simulating a virtual plant model together with the control software for system verification before physical integration~\cite{virtualCommissioningSurvey,VirtualCommissioningLearningPlatform,IntegratedVirtualCommissioningProcess}.
        Virtual commissioning falls out of the scope of this PhD project; instead, methods for modeling and visualizing robotic devices are focused on.

        An industrial ecosystem is described by the European Cluster Collaboration Platform as an entity containing all of the involved stakeholders in a value-chain~\cite{IndustrialEcosystemEuropeanCluster}.
        These could be companies, research institutions, research service providers, or suppliers.
        Expressing the relationships between the stakeholders in an industrial ecosystem can help identify bottlenecks or weak points in a value-chain.
        As the stakeholders in an ecosystem are typically dependent on each other; if one of them is disrupted, the rest of the ecosystem can be highly affected.
        As robotic systems integration revolves around modular technical systems that are constantly evolving~\cite{EcosystemVsSupplyChains}, it can be viewed as an industrial robotics ecosystem.
        
        European initiatives, such as the RobMoSys project\footnote{\url{https://cordis.europa.eu/project/id/732410}}, have been defined to create methodologies for developing composable models and software for robots systems.
        The idea behind this project is to achieve interoperability across stakeholders and applications within the robotics software ecosystem, illustrating the need for research and developments within robotic systems integration.
        The main focus of RobMoSys was on improving the overall software integration process across stakeholders within the domain of service robots.
        One of the interesting outputs of RobMoSys was an approach for model-driven software development and domain specific languages to solve the related challenges~\cite{StampferMDSD}.
        Similar to such research, this thesis focuses on improving the process of robotic systems integration, by advancing and streamlining the tasks of robotic system integrators.

        \subsection{Main Stakeholders}
        
\begin{figure}
            \centering
            \includegraphics[width=0.8\textwidth]{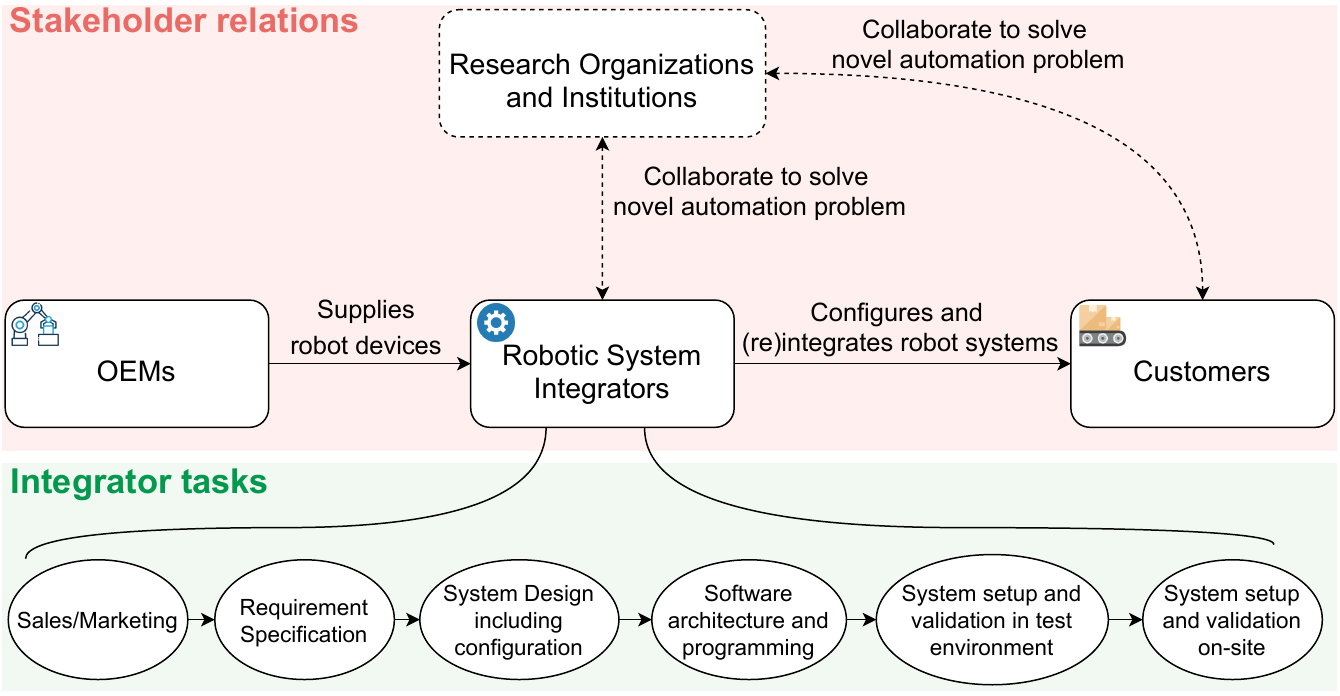}
            \caption{The main stakeholders and their relations when automating a process are shown in the upper rectangle. The dashed lines indicate the collaborations when automating a novel process. Typical integrator tasks are shown in the lower rectangle.}
            \label{fig:stakeholders_process}
        \end{figure}

        The main stakeholders involved in robotic systems integration, which are also part of the industrial robotics ecosystem, are described in this section.
        Their interconnections are shown in \cref{fig:stakeholders_process}, which is similar to the development process described in~\cite{Schlegel2015ModeldrivenSS}.
        Note, that not all stakeholders, such as component suppliers and standardization committees, are included in the figure.
        
        \begin{description}
            \item[\textbf{Robotic system integrators}:] design the complete system to automate the process by taking into account the customer's requirements, the task, and the available devices for building the robot system.
            Robotic system integrators highly rely on their prior experience with automating similar tasks using specific robot devices.
            Most robotic system integrators cover the acquisition, integration, and part of the deployment of robot systems.
            There are cases where the operational phase of the system (phase 4 in \cref{fig:chapters_overview}) is charged as an additional cost~\cite{robotiqBlogDeploymentIntegration,Sannemann&2020}.
            
            The terms robotic system integrators and integrators are used interchangeably throughout this thesis.
        
            \item[\textbf{Customers}:] are the consumers of robot systems who purchase them to automate processes and achieve long-term cost savings through a positive return on investment.
            A customer typically has limited knowledge of the technological aspects of the robot system, but may provide requirements for the task to be automated, such as the speed required to process an object.
            Customers will most commonly contact an integrator to automate a desired process.
            However, some customers have in-house integration~\cite{Sannemann&2020}.
        
            \item[\textbf{Original Equipment Manufacturers (OEMs)}:] are the suppliers of modular devices, such as robotic arms and end-effectors, that can be combined into a robot system.
            As these devices need to be connected together to comprise a robot system, it is necessary that they adhere to standards for interfacing the products, such as the robot flange~\cite{ISO9409} or industrial communication protocols~\cite{IndustrialRobotProtocols}.
            In some cases, integrators can also manufacture their own devices, being both an integrator and \acrshort{oem}.
            The same applies to \acrshortpl{oem}, where they sometimes also offer integration services.

            Both \acrshortpl{oem} and integrators can be distributors of devices, as integrators can in some cases solely sell a device without any integration services.

            \item[\textbf{Research organizations and institutions}:] conduct research on novel automation tasks that are requested by customers or robotic system integrators. 
            Research organizations may act as mediators between research institutions and the industry. 
            Industrial collaborators provide problems that need to be solved, while research institutions offer knowledge on cutting-edge research that the organizations can utilize to solve the industry's problem.
            An example of such an organization is the Danish Technological Institute (DTI)\footnote{\url{https://www.dti.dk/}}.
            The purpose of such an organization is to help improve the Danish industry, which is achieved by developing new technologies or maturing technologies developed from projects at research institutions and adapting them to industrial cases.
            
            In the context of robotic systems integration, research institutions are typically universities that are collaborating on a research project together with a customer, research organization, or integrator.
            As an example, this PhD project was defined under MADE FAST to be carried out at Aarhus University in collaboration with an integrator.

        \end{description}
        In addition to the stakeholders mentioned above, standardization committees are vital for such an industrial robotics ecosystem to function.
        Such committees (e.g.,~\cite{ISOTC299}) may define standards related to the interfacing of devices (e.g.,~\cite{ISO9409}) to ensure interoperability, and standards related to vocabulary (e.g.,~\cite{ISO8373}) to ensure consistency across manufacturers.
        Naturally, the suppliers of the components used by the \acrshortpl{oem} are also part of the ecosystem.

        As new technologies are continuously developed to keep up with the automation requirements of manufacturing lines, the different stakeholders need to collaborate on this.
        We refer to the Technology Readiness Levels (TRLs) defined by the National Aeronautics and Space Administration (NASA) agency, to specify where the different stakeholders are \emph{typically} located on the scale in the development of novel automation tasks:
        \begin{description}
            \item[Research institutions (TRL 1-4):] develop and test novel technology in a laboratory.
            \item[Research organizations (TRL 4-7):] test the technology from research institutions in a more appropriate environment, and evolve it to be ready for industrial use. Research organizations often bridge the gap between research institutions and the industry.
            \item[Robotic system integrators (TRL 7-9):] use the technologies provided by the research organizations, and deploy them in the industry.
        \end{description}
        For more details on this, we refer to~\cite{Tola&2023a}.

        \subsection{The Tasks of a Robotic System Integrator}
        Robotic system integrators are the main stakeholders that are integrating robot systems, and thus it is important to gain fundamental insight in their tasks.
        These tasks were identified through discussions with a Danish robotic system integrator.
        Integrating robot systems into manufacturing lines does not only consist of configuring and programming devices, but also involves sales, requirement specification, testing, and so on, as illustrated in \cref{fig:stakeholders_process}.
        In addition, other tasks may also be carried out, such as CE-marking, safety assurance, documentation, and factory and site acceptance tests.
        The main tasks performed by robotic system integrators are briefly described below:

        \begin{description}
            \item[Sales and marketing:] the integrator will either contact potential customers themselves, or be contacted.
                                        In any case, the integrator visits the customer on-site to document the manufacturing process and potential system requirements, that can be used when estimating the costs of automating the process.
                                        In some cases, especially for large and complex systems, visualizations or simulations will be created for marketing purposes.

            \item[Design and configuration:] this step corresponds to phase 1 in \cref{fig:chapters_overview}.
                                            The system layout is designed, and the specific devices that the robot system composes are chosen.
                                            As the devices are typically developed by different \acrshortpl{oem}, they must have compatible interfaces and be interoperable, setting limitations on which devices can be put together.
                                            It is common that each integrator gains expertise with devices from specific \acrshortpl{oem} and creates partnerships with these \acrshortpl{oem}, therefore prioritizing their devices.
            
            \item[Physical integration:] This step corresponds to phase 3 in \cref{fig:chapters_overview}, which entails physically connecting the                                      devices, programming the robots, and verifying the system as a whole.
                                        The system is first set up at the integrator's site and tested, then moved to the customer's site, integrated, and verified.

            \item[CE-marking and safety assurance:] in Europe, CE-marking means the system needs to fulfill a number of requirements to ensure it follows                                   European health and safety standards\footnote{\url{https://ec.europa.eu/growth/single-market/ce-marking_en}}.
                                            Multiple safety standards exist for industrial robots that need to be satisfied~\cite{ISO10218,ISO102182,ISO15066}, and if the robot collaborates with humans, more safety requirements are imposed on the system.
                                            The integrator must ensure the system is safe and document its safety, which is time consuming, particularly when automating novel processes.   
    
            \item[Other tasks:] In some cases, the integrator will perform virtual commissioning, or simulate the system before physically integrating the hardware, and perform maintenance checks of the system.
                                Additionally, when a new device enters the market, the integrators analyze and experiment with it to obtain knowledge about the capabilities, compatible devices, and the difficulty of deployment.
                                The new knowledge on the device is incorporated and resides within the robotic system integrator organization.
        \end{description}

\begin{contr}{Contribution 1 (C1):}
Characterized the main stakeholders involved in robotic systems integration and detailed the process.
\end{contr}

\section{Challenges} \label{chapter:integration:challenges}
\begin{figure}[!htbp]
    \centering
    \includegraphics[width=\textwidth]{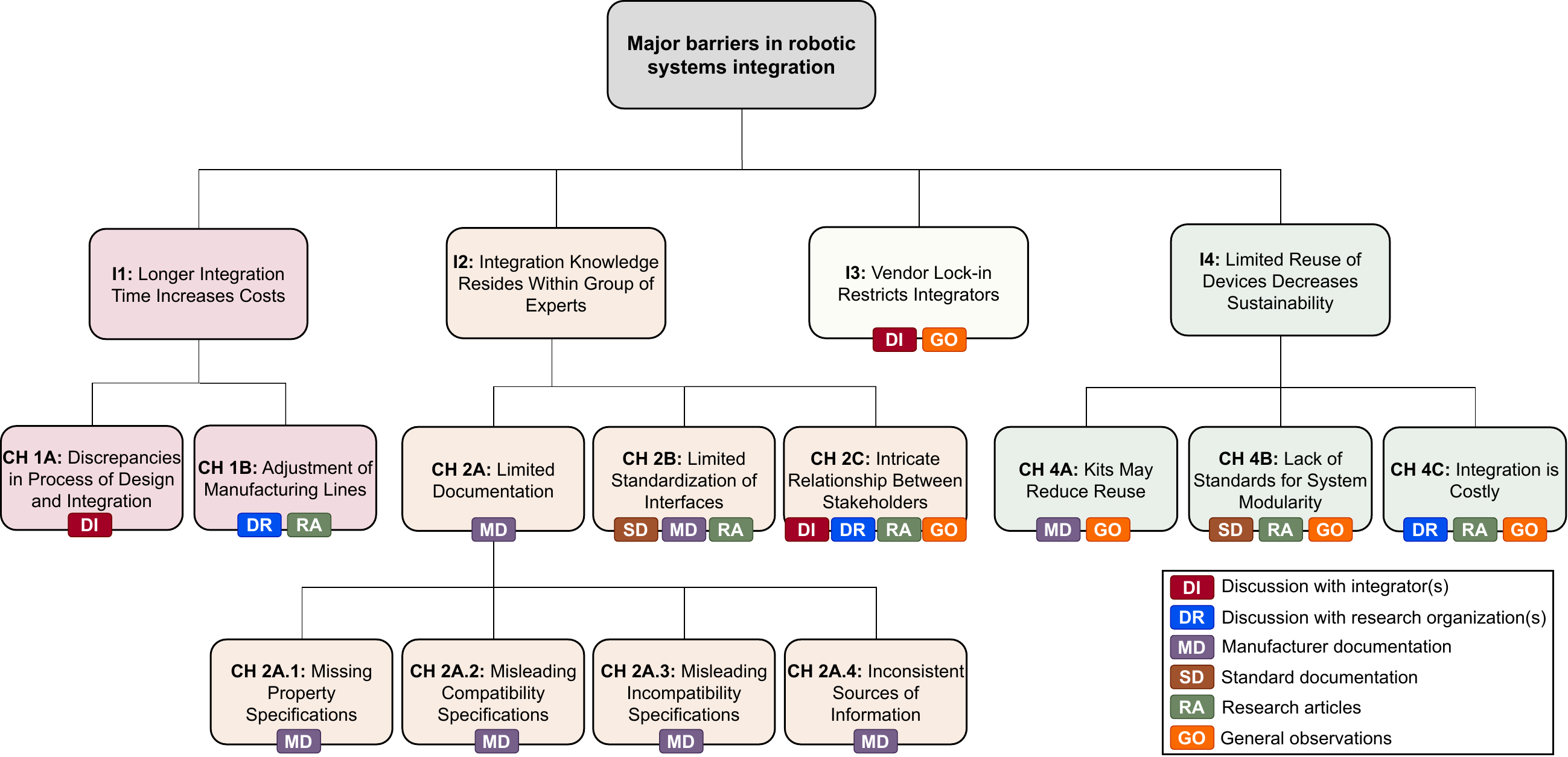}
    \caption{Overview of issues and their underlying challenges. The small boxes under each challenge represent the source of evidence that it is based on. The issues with no evidence are defined by converging related challenges with evidence into a group, and therefore these issues have no direct evidence.}
    \label{fig:fta}
\end{figure}

This section describes some of the major issues in the acquisition and integration of modular robot systems, and the underlying challenges (see \cref{fig:fta}).
The challenges found are based on empirical studies presented in publications~\cite{Tola2021},~\cite{Tola2022}, and~\cite{Tola&2023a}.
Some of which are based on findings during the assembly of robotic cells in the Aarhus University Digital Transformation Lab, during the development of a robot system configurator tool, discussions with experts in the area.
The challenges and examples originate from real industrial manufacturers, but to avoid harming their reputation, their names are anonymized.
A selection of the encountered challenges are presented in this section, the remaining can be found in~\cite{Tola&2023a}.
These challenges are summarized into issues we define as the major barriers in the acquisition and integration of modular robot systems.

\Cref{fig:fta} shows the major issues identified, and the challenges associated with each issue.
The source of evidence that the challenges or issues are based on are shown in boxes under each challenge/issue.
We define six types of sources of evidence, explained below, together with a rating of their credibility on levels~1-4, where~4 is the most credible:
\begin{description}
    \item[Discussion with integrator(s):]
        This involves discussions with two Danish integrators, and as they are directly involved in robotic systems integration, their credibility is rated as the highest, which is \textbf{level four}.

    \item[Discussion with research organization(s):]
        This involves discussions with both a Danish research organization and an Australian one.
        As they are directly involved with robotic systems integration and advancing technologies related to it, their credibility is rated at \textbf{level four}.

    \item[Manufacturer documentation:]
        This involves information provided by \acrshortpl{oem}, and as they are the producers of the devices, their credibility is rated at \textbf{level four}.

    \item[Standard documentation:]
        ISO standards related to robots and robotic systems integration are highly credible, and therefore rated at \textbf{level four}.

    \item[Research articles:]
        This involves information published in research articles related to robotic systems integration.
        As this information comes from researchers and not industrial experts, there may be some missing information, therefore it is rated at \textbf{level three}.

    \item[General observations:] 
        These can be implications made or knowledge gained through experience.
        As this information is based on observations and assumptions, there is a risk that these cannot be generalized, or the assumptions may be incorrect.
        Therefore, the credibility is rated at \textbf{level two}.
\end{description}

    \subsection{I1: Longer Integration Time Increases Costs}
    Current integration times of robot systems are too long, and it is common that the integration efforts are more expensive than the robot devices themselves, making it difficult for especially smaller enterprises to invest in robot systems.
    The cost of robotic systems integration is sometimes between 4-5 times more than the cost of the hardware itself~\cite{Sannemann&2020}.
    This issue is related to two main challenges, \textit{CH~1A} and \textit{CH~1B}, described below.
    
    \subsubsection{CH 1A: Discrepancies in Process of Design and Integration} 
    As the process of robotic systems integration can take place over several months, occasionally the environment (site) in which the system is to be deployed changes in between the design and deployment phase.
    Through discussions with an integrator, examples were provided of a project where an initial design of the system was developed, but required changes after finding that the site had been changed and a pole was blocking the path of a robot. 
    Such changes may force a re-design of the system, depending on their extremity, and can therefore result in longer integration times and an increase in the costs.
    
    \subsubsection{CH 1B: Adjustment of Manufacturing Lines} 
    Many robot systems are integrated and not properly tested after deployment.
    Discussions with a research organization revealed that various integrators neglect the deployment phase, where they simply deliver the robot systems and do not check if the deployed system is functioning and being operated in an optimal manner.
    This leads to a need to adjust the manufacturing lines to ensure the system is functioning properly, which is performed over multiple iterations, increasing the integration time and costs~\cite{Sannemann&2020}. 

    \subsection{I2: Integration Knowledge Resides within Group of Experts}
    A number of customers stated that they are highly dependent on integrators, of which there is a scarce amount~\cite{Sannemann&2020}.
    One of the main issues is that, due to a number of challenges, \textit{CH~2A}, \textit{CH~2B}, and \textit{CH~2C}, described below, knowledge on robotic systems integration resides within a group of experts.
    This makes it difficult for newcomers to enter the market of robotic systems integration, resulting in an insufficient amount of integrators compared to the current demand~\cite{Schaffer&18}.

    \subsubsection{CH 2A: Limited Documentation} 
\begin{figure}
    \centering
    \begin{subfigure}{.5\textwidth}
    \centering
      \includegraphics[width=0.7\linewidth]{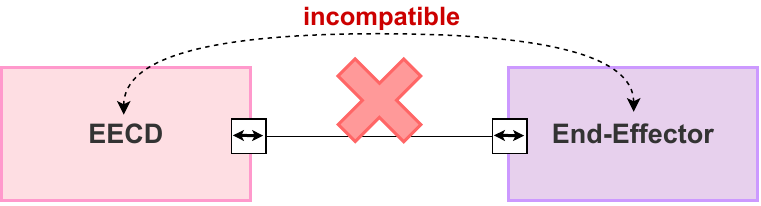}
      \caption{}
      \label{fig:eecd_incompatible} 
    \end{subfigure}%
    \begin{subfigure}{.5\textwidth}
    \centering
      \includegraphics[width=0.7\linewidth]{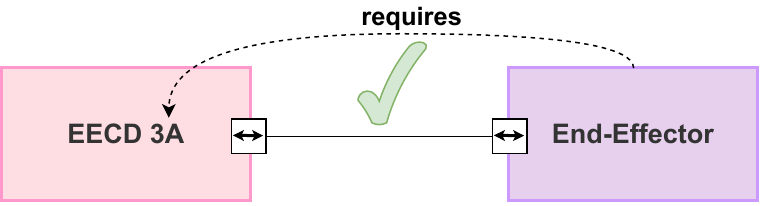}
      \caption{}
      \label{fig:eecd_compatible}
    \end{subfigure}
    \caption{Example of missing property specification (\textit{CH 2A.1}), recreated from~\cite{Tola2021,Tola2022}, included in~\cite{Tola&2023a}. (a) shows a case where the information of incompatibility between the devices was found by performing an empirical test. (b) illustrates a valid combination specified in the data sheet that the end-effector requires the \textit{EECD 3A}. The current of the \textit{EECD} is not specified in the data sheet.}
    \label{fig:eecd_current}
    \end{figure}

\begin{figure}[!htbp]
    \centering
    \begin{subfigure}[b]{\columnwidth}
    \centering
      \includegraphics[width=0.41\linewidth]{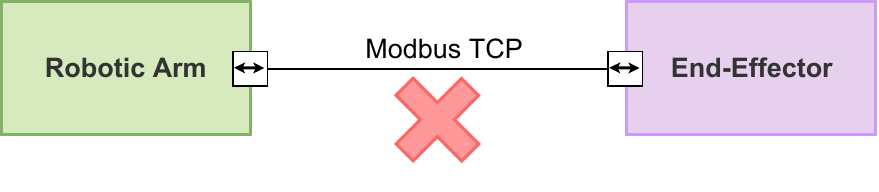}
      \caption{}
      \label{fig:modbus_tcp_incompatible} 
    \end{subfigure}
    \begin{subfigure}[b]{\columnwidth}
    \centering
      \includegraphics[width=0.7\linewidth]{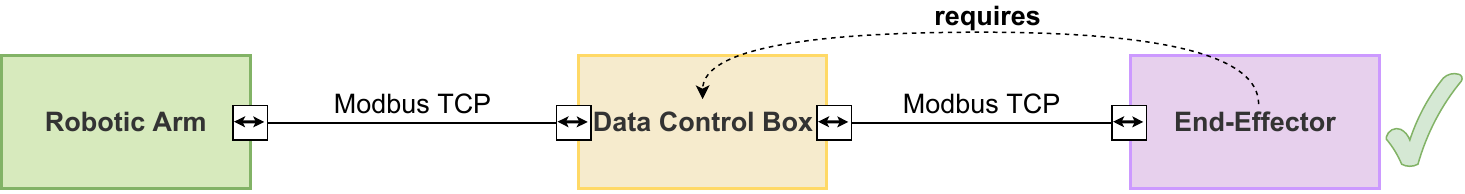}
      \caption{}
      \label{fig:modbus_tcp_compatible}
    \end{subfigure}
    \caption{Example of misleading compatibility specification (\textit{CH 2A.2}), recreated from~\cite{Tola2021,Tola2022}, included in~\cite{Tola&2023a}. In (a), both the robotic arm and end-effector use Modbus TCP as a communication interface, but are specified in the data sheet to be incompatible. Instead, as (b) shows, the end-effector requires a \textit{Data Control Box} for connecting to the robotic arm.}
    \label{fig:modbus_tcp}
    \end{figure}

\begin{figure}
    \centering
    \begin{subfigure}{.5\textwidth}
    \centering
      \includegraphics[width=0.8\linewidth]{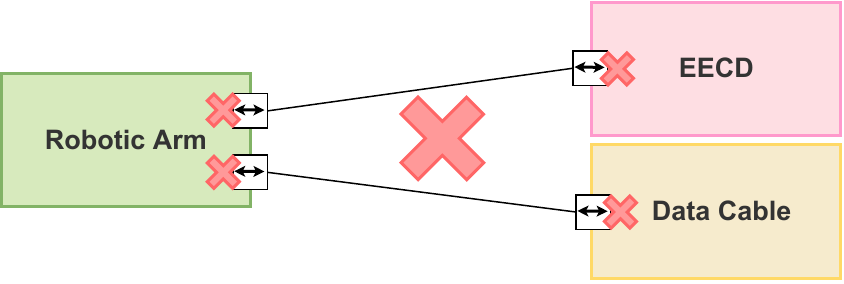}
      \caption{}
      \label{fig:three_incompatible} 
    \end{subfigure}%
    \begin{subfigure}{.5\textwidth}
    \centering
      \includegraphics[width=0.81\linewidth]{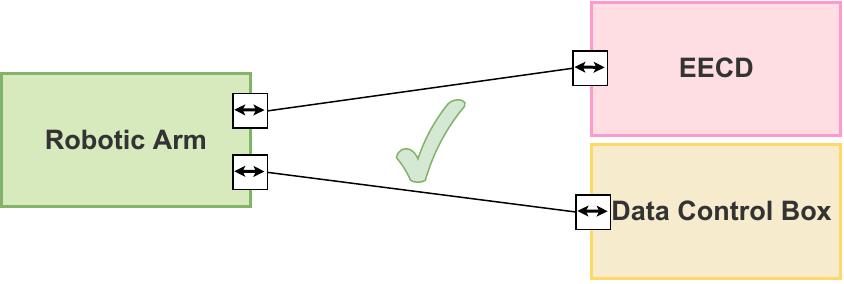}
      \caption{}
      \label{fig:three_compatible1}
    \end{subfigure}
    \caption{Example of misleading incompatibility specification (\textit{CH 2A.3}), recreated from~\cite{Tola2021,Tola2022}, included in~\cite{Tola&2023a}. (a) shows an incompatible configuration, where a data sheet specified that this specific robotic arm must never be connected to the specified \acrshort{eecd}. (b) shows an example where the same robotic arm and \acrshort{eecd} are connected, but through a different data connection, specified to be compatible in the same data sheet.}
    \label{fig:incompatibility_specification}
    \end{figure}

    There is currently no standard for composing data sheets of robotic devices~\cite{Sannemann&2020}, making it difficult to navigate through data sheets from different vendors.
    Furthermore, there may be missing or misleading information~\cite{Schaffer&18}.
    Specific examples related to limited documentation were found in the empirical studies from publication~\cite{Tola2022}, and are summarized below:
    \begin{description}
        \item[CH 2A.1: Missing Property Specifications:]
        The data sheet and user manual of one of the examined \acrshortpl{eecd} did not provide the electrical properties of the device.
        Instead, the naming of the device included an electrical current, leading one to assume this is the electrical property of the device, see \cref{fig:eecd_current}.

        \item[CH 2A.2: Misleading Compatibility Specifications:]
        If two devices have the same interfaces, it can be expected that they are compatible.
        Unfortunately, this is not always the case in specifications of robot devices.
        An example was found where an end-effector and a robotic arm both supported the same industrial communication protocol, Modbus TCP, but when reading further in the data sheet, it was stated that the end-effector required an additional device for processing data in order to be compatible with the robotic arm that supports Modbus TCP, see \cref{fig:modbus_tcp}.

        \item[CH 2A.3: Misleading Incompatibility Specifications:]
        If two devices are specified as incompatible, one would assume that they can never be connected in any manner.
        This assumption was found to be incorrect, as \cref{fig:incompatibility_specification} illustrates, a robotic arm and \acrshort{eecd} were specified as incompatible in the \acrshort{eecd} data sheet, to find further down in the data sheet that these two devices \emph{can} be connected if, and only if, a specific data connection is used.

        \item[CH 2A.4: Inconsistent Sources of Information:]
        Information about the interfaces of a device is expected to be found in the data sheet or user manual by the \acrshort{oem} of the device.
        We found an example, where the robot flange of a robotic arm was not specified in the data sheet by the \acrshort{oem} of the device, but instead by another \acrshort{oem} of end-effectors that can be attached to the given robotic arm.
        
    \end{description}

    \subsubsection{CH 2B: Limited Standardization of Interfaces} 
    Currently, the only mechanical interface in a robot system that is standardized is the robot flange through ISO 9409~\cite{ISO9409}.
    This makes it difficult to determine which modules fit together as an extensive analysis may be required, that is, detailed studies of data sheets and empirical tests of devices~\cite{Sannemann&2020}.
    With regards to the communication protocols, many industrial protocols exist, and each manufacturer chooses their own preferred protocol, complicating the process of determining the connectivity of devices~\cite{Sannemann&2020}.

    Additionally, \acrshortpl{oem} only support interoperability standards to a limited extent, for strategic reasons. 
    Some may potentially do this to make it harder for new competitors to enter the market and to avoid being replaced by cheaper competitors~\cite{Schaffer&18}.
    Specific examples supporting this, include multiple \acrshortpl{oem} not providing the standardized ISO 9409~\cite{ISO9409} robot flange in the documentation of their devices~\cite{Tola2022}.

    \subsubsection{CH 2C: Intricate Relationship Between Stakeholders} 

    As integrators need to keep up with new devices on the market and determine their interfaces and compatibility, this gained knowledge stays within the organization, meaning that new integrators would need to acquire all of this knowledge from scratch.
    This complexity was clear during this PhD project, where a great amount of the compatibility and configuration information about devices needed to be learned.
    Newcomers are dependent on good practice within documentation of devices and standardization of interfaces, which is clearly not the case in this ecosystem.
    This is observed through the intricate relationship between the stakeholders, where the integrators and \acrshortpl{oem} are in some cases competitors as they both perform robotic systems integrations.
    At the same time, \acrshortpl{oem} typically form partnerships with integrators, giving them better prices on devices, making it difficult for newcomers entering the market.
    There are no requirements on the responsibilities of the \acrshortpl{oem}, allowing them to document products and change modules and interfaces as they wish, and the integrators need to ensure that they learn how to use these new devices and keep that knowledge within their organization.
    Additionally, multiple digitalization efforts created by different \acrshortpl{oem} (presented in more detail in \cref{chapter:configurator}) only focus on the devices from the specific \acrshort{oem}, evolving only the business of that \acrshort{oem} instead of collaborating to evolve the complete ecosystem.
    There is nothing wrong with the integrators and \acrshortpl{oem} doing this, but if robotic systems integration needs to be cheaper and less time-consuming, there is a need to re-evaluate these conventional methods within the industrial robotics ecosystem.

    \subsection{I3: Vendor Lock-in Restricts Integrators}

    Vendor lock-in is a major concern in the industrial robotics ecosystem. 
    This occurs when integrators are heavily reliant on a few specific \acrshortpl{oem} for their robotic devices and software, where the integrators invest heavily in training and certification for the \acrshortpl{oem}' devices. 
    Furthermore, integrators often have close relationships, or so-called partnerships, with their preferred \acrshortpl{oem}, which may include discounts, early access to new products, and technical support.
    This can make it difficult and costly to switch to devices from other \acrshortpl{oem}, leading to vendor lock-in.
    This can have a number of negative consequences.
    It can make it difficult for integrators to create the best robot systems for their customers' needs, as they may be limited to using devices and software from few \acrshortpl{oem}.
    Additionally, vendor lock-in can make it difficult for newcomers to enter the industrial robotics market, as they may have difficulty competing with established integrators who have close relationships with major \acrshortpl{oem}.
    
    Recent examples, like Unity's changes to licensing conditions by adding a \textit{runtime fee}\footnote{\url{https://www.theguardian.com/games/2023/sep/12/unity-engine-fees-backlash-response}}, underline the vulnerabilities associated with relying on vendor-specific software. 
    When such changes occur, they can have a domino effect, impacting numerous businesses and raising concerns about future adaptability and costs. 
    In essence, diversification in the industrial robotics ecosystem is not just a matter of reducing risk; it is a means to foster innovation, competition, and resilience within the industry.

    \subsection{I4: Limited Reuse of Devices Decreases Sustainability}
    The reuse of robot devices is currently limited due to a number of challenges, \textit{CH~4A}, \textit{CH~4B}, and \textit{CH~4C}, described below.
    Many robots are designed for re-purposing on new manufacturing lines, and disposing of these devices instead of reusing them is unsustainable.
        
    \subsubsection{CH 4A: Kits May Reduce Reuse}
    Some \acrshortpl{oem} of end-effectors provide \textit{kits} that can be bought for specific series of robots. 
    These kits contain an end-effector, \acrshort{eecd}, data connection, and other necessary equipment for connecting these devices to a robotic arm.
    Kits were found that were created for a specific type of robotic arm but contained the exact same devices (except for screws) as a different kit for a completely different type of robotic arm.
    Non-technical users of such kits may purchase new devices instead of reusing the available ones they already own, as it is not clear that some of these devices can be reused across robotic arms.
    Kits make buying robot devices easier, but they discourage reusing devices because it is easier to buy a new kit than to find compatible devices for a new configuration.
    
    \subsubsection{CH 4B: Lack of Standards for System Modularity}
    There is a general lack of standards for robot systems, as explained in \textit{CH~2B}.
    Without enough standards for system modularity, it is difficult to create plug-and-play solutions that can be composed into modular robot systems, where a module can be exchanged in the re-purposing of a robot system.
    Working towards an improved system modularity within robot systems would increase the interchangeability of devices resulting in faster integrations~\cite{Sannemann&2020}.

    \subsubsection{CH 4C: Integration is Costly}
    As explained above, robotic systems integration is often more expensive than the hardware itself, and reconfiguring an existing system is typically more expensive than configuring it from scratch~\cite{Sannemann&2020}.
    Companies typically opt for the cheaper solution, which due to the current complex integration process, is to configure and integrate a system from scratch instead of reusing devices.
    

    \subsection{Summary and Discussion}
    The consequence of the challenges described above is that customers are highly reliant on integrators, that within their organizations, possess domain knowledge on configuring devices through years of experience with the products.
    A number of companies that participated in the interviews performed by~\cite{Sannemann&2020}, described that one of the main challenges they face is the integration of robots into a manufacturing line.
    Companies are so reliant on integrators that some of them even informed that they could not proceed in the robotic systems integration process due to the scarce amount of integrators.
    One of the companies in the interview claimed that their reliance on integrators was a disadvantage, and therefore opted for in-house integration instead.
    
    The challenges, presented in \textit{CH 2A}, indicate that \acrshortpl{oem} spend too little time on producing proper and consistent documentation for their devices, making the integrators adapt to the scarce documentation by empirically testing devices.
    This means that integrators that have newly entered the market have a great disadvantage with regards to the knowledge on the compatibility of devices compared to an integrator with years of experience.
    
    As the challenges indicate, the stakeholders in the industrial robotics ecosystem are dependent on each other, and there is a need for them to work together to make robotic systems integration more accessible and less expensive.
    This is a must if we want to increase automation and decrease physically tiring labor across the world.

    The main barriers related to robotic systems integration were outlined, and potential directions for tackling them are described below, thus addressing \textbf{RQ1}.
    
\begin{contr}{Contribution 2 (C2):}
Outlined current challenges in acquiring and integrating modular robot systems and presented directions to tackle them.
\end{contr}

\section{Potential Directions}
This section describes some of the potential directions to work towards in order to reduce the effect of the challenges described above.

    \subsection{Increase Interchangeability Through System Modularity}
    To decrease (re-)integration time, and improve re-usability of devices, it is crucial to work towards modular robot systems, where devices can easily be exchanged.
    This can be achieved by developing plug-and-play devices, or by defining and using standardized interfaces for the devices in a robot system.
    Inspiration can be drawn from a successful system, the \acrfull{pc}, which defines standards on interoperability including the responsibilities of \acrshortpl{oem}~\cite{ISO130661}, and describes the usability of the computers by explicitly defining the vocabulary to be used across operating systems~\cite{ISO9241}.
    If robot system modularity is improved in the future, it will help robotic systems integration with all three issues mentioned above.
    However, system modularity is out of the scope of this thesis.
    
    \subsection{Easier Robot System Configuration}
    Manually configuring the devices in a robot system and determining their compatibility can take time, adding to the total integration costs.
    Developing a configurator that can automate this process, similarly to the \acrshort{pc} configurator\footnote{\url{https://pcpartpicker.com/}}, may help customers and integrators easier determine the compatibility of devices, when configuring a robot system.
    This may also lead to an increase in the reuse of devices, if their compatibility can easily be determined in a configuration tool.
    If a configurator is developed containing all the domain knowledge about devices, it will be easier for new integrators to enter the market, and it will also shorten the integration time of robot systems.
    This PhD project addresses this by creating a proof-of-concept configurator, that takes into account some of the documentation challenges described above, and is presented in \cref{chapter:configurator}. 
    \textbf{RQ2} is addressed in this chapter, and can be related to tackling challenges \textit{CH~2A} by creating a configurator that takes these limitations into account, \textit{CH~4A} by enabling reuse of devices through automated configuration, \textit{CH~4C} by enabling a faster acquisition process, and \textit{I3} by providing a tool for finding appropriate devices regardless of \acrshort{oem} restrictions.

    \subsection{Simulation and Digital Twins}
 
    To decrease the integration and deployment time of robot systems, the design can be simulated and verified before investing in the hardware~\cite{Sannemann&2020}.
    Simulation can be used in multiple phases of integration and deployment.
    It can be used in the design phase to experiment with different devices, or before deployment for system verification (as virtual commissioning), or in the context of \acrshortpl{dt} where processes can be simulated and compared to data from the physical system for optimization of the process.

    As simulations can be used across different phases of robotic systems integration, this PhD project focuses on investigating and improving an interoperable format for modeling devices and reusing these models across different simulation environments.
    This is presented in \cref{chapter:visualization}.
    \textbf{RQ3} is addressed in this chapter, and can be related to tackling challenges \textit{CH~1A} by enabling faster simulation setup and therefore faster integration and less time for discrepancies to occur, and \textit{I3} by improving interoperability across software tools.
    
    The concept of \acrshortpl{dt} in the context of unit level machines is explored and a case-study is presented showing how a \acrshort{ds} of a modular robot system can be developed.
    These are introduced in \cref{chapter:digital_twins}.
    \textbf{RQ4} is addressed in this chapter.

    \subsection{Reduce Vendor Lock-in Through Interoperability}
  
    Creating interoperable devices and software for robotics is essential for tackling the negative effects of vendor lock-in and mitigating the associated risks.
    The configurator presented in \cref{chapter:configurator} allows defining rules for various types of devices, thus enabling configuration of a larger database of devices, which may assist in reducing vendor lock-in.
    The modeling format, \acrshort{urdf}, presented in \cref{chapter:visualization}, is an interoperable format that can be imported into various simulation and visualization tools, making users less dependent on a single vendor-specific tool.
    The \acrshort{dt} architecture proposed in \cref{chapter:digital_twins} is based on the concept that modules can be reconfigured and reused across robot systems, avoiding vendor-specific software.
    All of these technologies and methods described in the following chapters provide means to tackle vendor lock-in from both a hardware and software point of view.

\chapter{Configuration of Modular Robot Systems} \label{chapter:configurator} 
\backgroundsetup{
  scale=1,
  color=black,
  opacity=1,
  angle=90,
  position=current page.north,
  vshift=-250mm,
  hshift=-10mm,
  contents={
    \textcolor{gray}{\rule{5mm}{\paperheight}}
  }
}

This PhD thesis aims to reduce the time spent in the process on robotic systems integration by advancing digitalization efforts in this area. 
This chapter introduces the concept of configuration, provides an overview of current industrial digitalization efforts related to acquisition and their limitations, and presents the domain model developed during this PhD project for a robot system configurator that considers device compatibility in spite of missing and inconsistent information.
This chapter is mainly based on the research output from publication~\cite{Tola2021} and partially based on publications~\cite{Tola2022} and~\cite{Tola&2023a}.

\section{Research Method}
The goal of this chapter is to answer \textbf{RQ2} that is defined as: \textit{is it possible to define a configurator of modular robot systems that takes application requirements and compatibility of devices into account?}

\begin{figure} 
    \centering
    \includegraphics[width=0.8\textwidth]{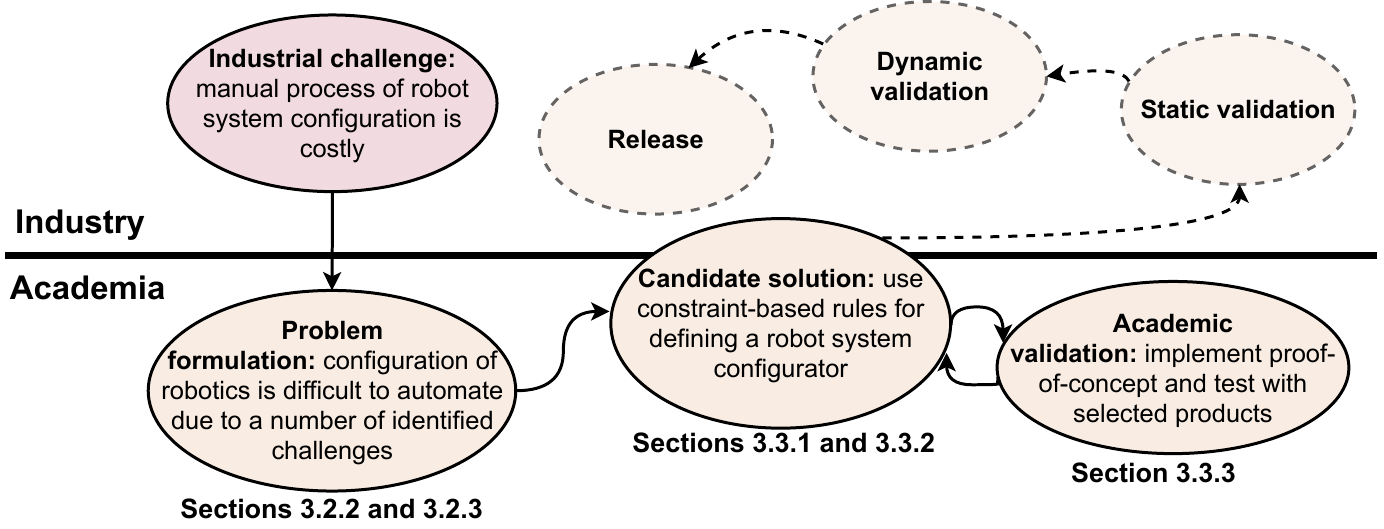}
    \caption{Research method for our configuration approach adapted from the technology transfer model in~\cite{ModelForTechnologyTransferGorschek}. The dashed shapes are steps that have not been carried out in this PhD project. The horizontal line distinguishes between industrial and academic workload. Static validation entails interviews and seminars, while dynamic validation involves pilot projects.}
    \label{fig:robocim_research_method}
\end{figure}

To address \textbf{RQ2}, we present a domain model that tackles some of the documentation and compatibility-related challenges presented above in \cref{chapter:integration:challenges}.
A proof-of-concept of the domain model is developed and validated on a small subset of products from an integrator's (Technicon's) product catalog.
As the goal of this research is to be applied in the industry, we follow the technology transfer model defined by~\cite{ModelForTechnologyTransferGorschek}, as illustrated in \cref{fig:robocim_research_method}, where an industrial challenge is defined and through research, a solution is proposed~\cite{kothari2004research}.
The parts of the transfer model related to academia have been carried out in this PhD project, but the industrial parts of validation and release have not been possible, due to the limited collaboration with Technicon.
The industrial challenge was defined together with Technicon by gaining insight into their configuration process and studying the current industrial methods for robot system configuration.
The results of this research can be used in the development of an industrial robot system configurator.

\section{Background}

In this chapter, we define the following:
\begin{description}
    \item[product/device:] is an object that can be manufactured and potentially composed into a system.

    \item[system:] is a structure composing various devices and has a specific purpose, such as a robot system.

    \item[component:] is an object that can be a product or related to a product. This concept is mainly used in the proposed domain model to distinguish between objects defined in the domain model.
\end{description}

\backgroundsetup{
              scale=1,
              color=black,
              opacity=1,
              angle=90,
              position=current page.north,
              vshift=+250mm,
              hshift=-10mm,
              contents={
                \textcolor{gray}{\rule{5mm}{\paperheight}}
              }
            }

\subsection{Mass Production, Mass Customization, and Configuration}
Mass production was initiated by Henry Ford, the founder of the Ford Motor company, in the early 1900's with the goal of efficient production of large volumes of identical products/systems.
Followed by this, mass customization emerged in the late 1990's which had the same goal as mass production but with the manufacturing of customized products/systems instead of identical ones.
This resulted in technology developments that could support mass customization, and from this, knowledge-based configuration emerged~\cite{KBC_chapter1}.
The main task of configuration, as outlined in~\cite{Sabin&98}, involves selecting combinations of products from a predefined set to collectively meet the task's requirements. 
Additionally, it entails generating a comprehensive list of the products that form both the final system and its structure. 
The term \textit{configuration} can both be used as the task of configuring a system and also as the final solution found for a system composition.

\begin{figure}
    \centering
    \includegraphics[width=\textwidth]{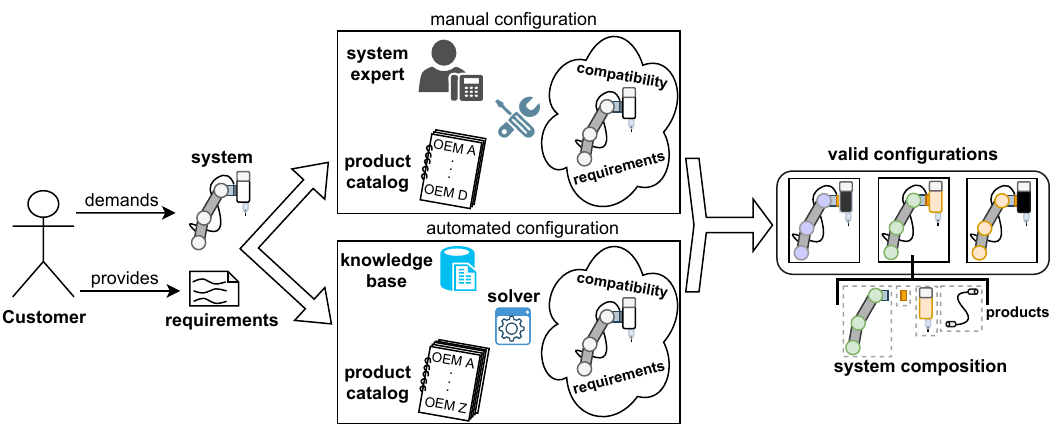}
    \caption{Comparison of main stakeholders and artifacts in manual and automated system configuration processes. The bold text are terms related to configuration.}
    \label{fig:configuration}
\end{figure}

The process of configuration can be carried out manually, as shown in the upper rectangle in \cref{fig:configuration}, or automatically using a solver, as shown in the lower rectangle.
Manual configuration requires a system expert that analyzes the customer's demands and requirements, taking into account which products exist in the product catalog, together with the compatibility properties of the products.
The result is a list of valid configurations that adhere to the given requirements.
Each configuration is a system that composes a number of compatible products.
Automated configuration, on the other hand, only requires a system expert during the development of a knowledge base, and a configuration expert during the development of the configurator software.

\backgroundsetup{contents={}}

\subsubsection{Automating Configurators}
The development of a configurator requires~\cite{KBC_chapter1}:
\begin{description}
    \item[Configuration knowledge base:] developed by knowledge engineers in collaboration with domain or system experts that have knowledge about the system,                                   marketing and sales.
                                        This is the major task of creating a configurator as it involves defining products, their interfaces, constraints on how they can be connected and their compatibility, and the possible structures of the system.
                                        It requires formalizing the domain knowledge about the system and its products.
                                        
    \item[User interface:] developed by software developers and potentially in collaboration with domain experts and end-users.
    
    \item[Software integration:] of the configurator into existing systems. 
\end{description}

The user interface and software integration are not considered in this thesis, as these parts are highly related to industrial use.

Configurators can be created using model-based knowledge representations, which separate problem-solving knowledge (how to determine valid configurations and product relationships) from domain knowledge (specific product properties, relations, and system structure).
The inputs of these configurators are a product catalog and customer requirements, and the outputs are valid configurations of the system.
The problem solving knowledge and domain knowledge are typically represented using constraints~\cite{KBC_chapter6}.

The knowledge in a constraint-based configurator can be formalized using the \acrfull{ocl}~\cite{OCL} that can define system or object properties using logical expressions and invariants. 
\acrshort{ocl} is commonly used together with \acrfull{uml}\footnote{Represents standardized methods for modeling systems and software, see \url{https://www.uml.org/}.} class diagrams, as it allows including additional information about the relationships between the objects.
Two of the commonly used \acrshort{ocl} types are \textit{invariants} and \textit{implications}.
Invariants can be used to define constraints that must always hold.
For example, the age of a person can be defined as at least zero, in the following \acrshort{ocl} listing:
\begin{lstlisting}[numbers=none, label={lst:ocl_invariant}]
context Person inv:
    self.age >= 0
\end{lstlisting}
Implications are used to describe relationships between objects in the form of a logical \textit{if}-\textit{then} statement.
To elaborate, \textit{if} an assumption is true, \textit{then} some conclusion must hold, where the implication is used between the \textit{if} and \textit{then}.
For example, to describe that a married person's partner must be at least eighteen years old, an invariant and implication can be combined in \acrshort{ocl}.
This is shown in the following listing, that creates an invariant defining that if a person has a partner then their partner's age must be greater than eighteen:
\begin{lstlisting}[numbers=none, label={lst:ocl_implication}]
context Person inv:
    (self.partner->notEmpty() implies self.partner.age >= 18)
\end{lstlisting}
The keyword \texttt{context} in \acrshort{ocl} specifies the type of instance of the expression, which in the previous two examples is \texttt{Person}.
The \texttt{self} is an instance of the type \texttt{Person}, and the label \texttt{inv} indicates an invariant constraint.

Although \acrshort{ocl} is well suited for specifying constraints on \acrshort{uml} models, it is not optimized for solving complex general-purpose problems, and therefore must be translated into an executable representation that would then be run by a solver~\cite{KBC_chapter11}.

\subsection{Configurators in the Context of Personal Computers}

An example of a system that can be configured to the user's needs is the \acrshort{pc}, where the set of predetermined products would be the product catalog containing motherboards, Central Processing Units (CPUs), etc., and the produced list would contain configurations of the products that together satisfy the user's needs and compose the \acrshort{pc}.
\acrshortpl{pc} are similar to robot systems in the manner, that they consist of devices that together compose a complete system based on mechanical, electrical, and software interfaces.

Various online configurators for \acrshortpl{pc} exist, such as PC PartPicker\footnote{\url{https://pcpartpicker.com/}}, PC Builder\footnote{\url{https://pcbuilder.net/}}, or the CORSAIR PC Builder\footnote{\url{https://www.corsair.com/eu/en/pc-builder/}}.
All of these configurators only show devices that are compatible with each other.
To give an example, the CPU socket on the motherboard must match the socket of the CPU itself for them to be compatible.
Incorporating this domain knowledge into the configurator simplifies the purchasing of a customized computer, as the user avoids the need to determine the compatibility of the products themselves.

\begin{figure}
\centering
\begin{minipage}{.48\textwidth}
  \centering
  \includegraphics[width=\textwidth]{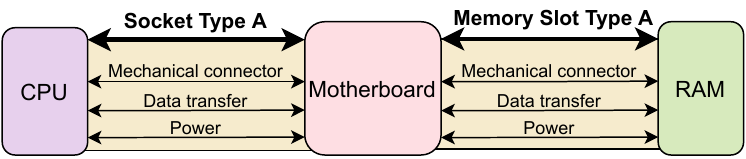}
  \captionof{figure}{Example of standardized connections between \acrshort{pc} devices. The \textit{Socket Type A} connection includes the mechanical connector, data transfer, and power specifications for that specific socket type. The same applies to the \textit{Memory Slot Type A}.}
  \label{fig:pc_config}
\end{minipage}
\hfill
\begin{minipage}{.48\textwidth}
  \centering
  \includegraphics[width=\textwidth]{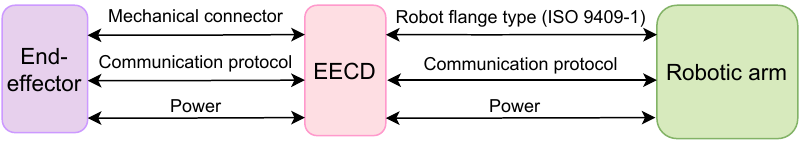}
  \captionof{figure}{Example of interfaces between robot system devices. The only standardized interfaces are the robot flange and the communication protocols that are industrial. No standardized connections that compose these interfaces exist.}
  \label{fig:robotic_arm_config}
\end{minipage}
\end{figure}

Some of the main advantages of devices in \acrshortpl{pc} is that they have standardized connections that composes all mechanical, electrical, and software interfaces into one main connection.
\Cref{fig:pc_config,fig:robotic_arm_config} illustrate the current differences in the interfaces between devices in a \acrshort{pc} and a modular robot system.
There are no standardized connections between the robot devices that comprise mechanical, electrical, and software interfaces altogether, making the task of robot system configuration more complicated than that of \acrshortpl{pc}.

\subsection{Configurators in the Context of Robot Systems}
\emph{Configuration} of robot systems using modularity, as defined by ISO 8373~\cite{ISO8373}, is \textit{``the arrangement of modules to achieve the desired functionality of a robot''}.
It involves composing a robot system of compatible devices that meet the application requirements, and is consistent with the definition of configuration by~\cite{Sabin&98} provided earlier.

Configuring robot systems is not as simple as configuring well-matured systems, such as \acrshortpl{pc}, due to some of the following key points:
\begin{itemize}
    \item the rapid evolution of technologies that are used in robot systems and corresponding devices makes it difficult to keep up with new interface types and compatibility constraints;
    \item the limited efforts on standardization of interfaces, as shown in \cref{fig:pc_config,fig:robotic_arm_config}; and
    \item the limited documentation efforts and their complications as described in \cref{chapter:integration:challenges}.
\end{itemize}

When developing a configurator, the limited standardization of device interfaces would require that a domain expert either defines their own terms for the interfaces or spends immense effort on describing the interfaces in detail.
The limited documentation of devices means that integrators, in some cases, need to perform empirical tests to determine device compatibility.
Once the devices are tested, the knowledge on device compatibility resides within the integrator.
Due to the complexity of gaining this domain knowledge, many integrators only use devices from a limited number of \acrshortpl{oem} in which they form long-term partnerships, limiting the configuration space for customers wanting to buy robot systems.
One of the current challenges of formalizing the domain knowledge of robot system configuration, is that it may require performing empirical tests or collaborating with a robotic system integrator.
Furthermore, the constraints on device compatibility are not straightforward, as the challenges in \textit{CH~2A} illustrate.

\subsection{Industrial Digitalization Efforts within Acquisition} 
Digital applications to assist customers with the acquisition of robot systems have recently been emerging.
Examples of such are the application builders from the robotic arm \acrshortpl{oem} Universal Robots\footnote{\url{https://www.universal-robots.com/builder/} released in 2018.} and ABB\footnote{\url{https://applicationbuilder.robotics.abb.com/en/home} released in 2021.}, where a user can get help to choose appropriate robots for a given application, and view an animation of the application with chosen devices from that specific \acrshort{oem}.
Application builders help customers better understand the potential of automation, and gain insight in the types of devices that a robot system comprises.

Other digitalization efforts in acquisition and integration are the D:PLOY platform\footnote{\url{https://onrobot.com/en/solutions/dploy} released in 2022.} and the Machine Tending Solution Configurator\footnote{\url{https://form.jotform.com/221716496065258} released in 2022.} from the end-effector \acrshortpl{oem} Onrobot and Robotiq, respectively. 
The D:PLOY platform can be used for building, running, monitoring, and re-deploying robot applications.
The Machine Tending Solution Configurator can be used to configure machine tending applications by inputting application requirements.
Note that these applications are all developed by specific \acrshortpl{oem}, thereby aiding customers in configuring only their own devices.

As far as our knowledge extends, the platform \textit{vention.io}\footnote{\url{https://vention.io/}}, is the sole manufacturer-independent platform that facilitates designing and simulating a complete robot system application with 3D visualizations before ordering and deployment of the system.
Inspecting this platform, we found that compatibility and requirements of devices were not automatically accounted for in the design process, instead the option of receiving feedback from a system expert was offered before making a purchase.
This means if a customer chooses to skip the feedback step, they may risk the purchase of an incomplete or incompatible system.
Additionally, only a limited set of \acrshort{oem} products were available on the platform, e.g., robotic arms from one of the leading \acrshortpl{oem}, Kuka, were not supported.
The limitations of such digital platforms are:
\begin{itemize}
    \item manufacturer-specific solutions are produced, limiting the user to devices from that \acrshort{oem}; and
    \item compatibility of devices and the completeness of a system is not considered, allowing the user to purchase a deficient system.
\end{itemize}
These examples are a subset of industrial tools developed to decrease the acquisition time of robot systems, and thus illustrate the significant interest in this area.
The documentation challenges described above, these examples of industrial tools, and the fact that robot system configuration is currently a manual task, all motivate our work on creating a domain model for robot system configuration that can overcome these limitations, and potentially reduce acquisition and integration costs.

\section{RoboCIM: Robot Configurator Information Model} 
This section is mainly based on the research output from publications~\cite{Tola2021} and~\cite{Tola&2023a}, and parts of the text are taken directly from~\cite{Tola&2023a} (that is not submitted yet).
We present our domain model, Robot System Configurator Information Model (RoboCIM), in this section.
For more information, we refer to~\cite{Tola2021}.

\subsection{Related Work}

Recent developments by~\cite{Stampfer18} present a method for composing software services based on their compatibility and application requirements.
Their work primarily focuses on high-level composition of software for service robots and does not delve into constraints and compatibility of the physical devices in a robot system.

A conceptual platform for capturing knowledge, configuring robot systems, and visualizing them is presented in~\cite{Schaffer&18}.
They developed a software mock-up of the framework through an online-based user interface. 
While their work contributes to robot system configuration, it operates at a higher abstraction level and does not address device-specific constraints.

A methodology for developing a robot system configurator is presented in~\cite{SchafferMethodForKnowledgeConfiguration}, with methods for knowledge acquisition and constraint definition.
Although this methodology provides an overview of how a configurator can be developed, it is only presented at a conceptual level with no practical implementation.

A hardware configurator was designed and developed for aiding operators in selecting suitable hardware parts for a given robot application in~\cite{Schou&17}.
They developed an ontology containing information about products and applications. 
However, their approach assumes the availability of complete information regarding device compatibility and connectivity.

Despite these valuable advances in robot system configuration, they all assume the availability of comprehensive and consistent information. 
In contrast, our domain model aims to tackle the challenges posed by incomplete, contradictory, or rapidly changing information, providing insights into the development of configurators under such conditions.

\subsection{Domain Model}

The developed domain model can be used to determine which devices are compatible when customers are purchasing a robot system, or in the re-purposing of a manufacturing line.
This configurator is designed to tackle the challenges in \textit{CH~2A}, presented in \cref{chapter:integration:challenges}.

To our knowledge, no existing configurator or domain model solves these issues, other than our developed domain model, RoboCIM.
The goal of RoboCIM is to formalize a model that can capture uncertainties in information from different sources.
We concentrate on the products within robot systems that exist between the robotic arm and the tool, excluding the base upon which the robotic arm is affixed.
These products possess mechanical, electrical, and data interfaces, and are therefore suitable for use in a proof-of-concept domain model.

\begin{figure}
    \centering
    \includegraphics[width=\textwidth]{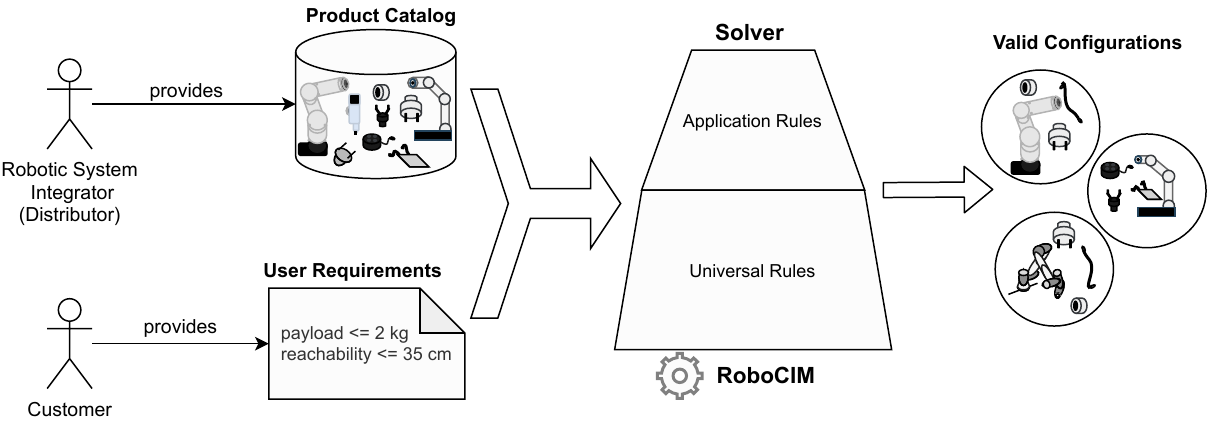}
    \caption{Overview of stakeholders and artifacts involved in robot system configuration, including the RoboCIM domain model, recreated from~\cite{Tola2021} under the CC BY 4.0 license, included in~\cite{Tola&2023a}.}
    \label{fig:robocim_use}
\end{figure}

The stakeholders and artifacts directly involved in robot system configuration are illustrated in \cref{fig:robocim_use}, together with RoboCIM and its two levels of rules, namely \textit{universal} and \textit{application} rules.
The universal rules are general rules for any application in which incomplete or contradictory information is inherent, while the application rules describe domain rules of robot systems and their devices.
RoboCIM uses default reasoning to tackle incomplete information.
This type of reasoning, and how it can be used to tackle incomplete information in the context of robot systems is described below.

\subsubsection{Universal Rules}
\begin{figure}
    \centering
    \includegraphics[width=0.95\textwidth]{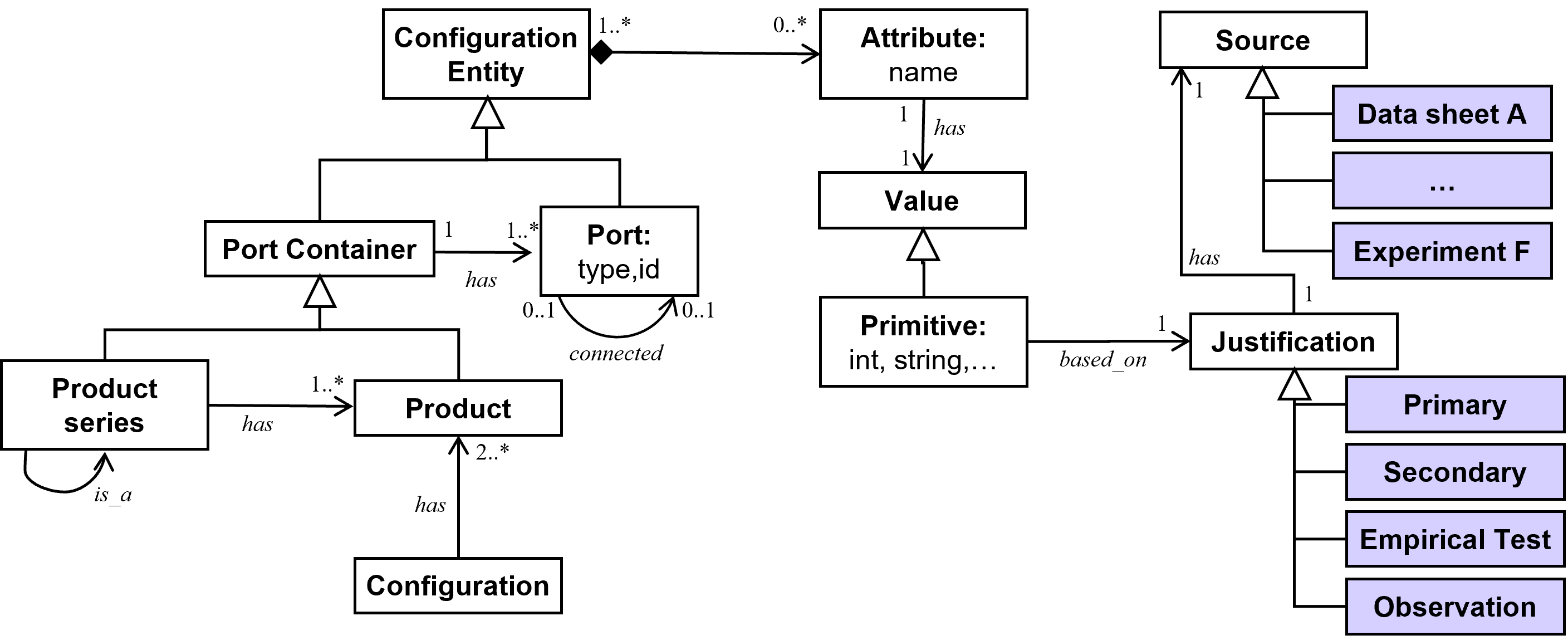}
    \caption{\acrshort{uml} class diagram representing the structure of the universal rules of RoboCIM, recreated from~\cite{Tola2021} nder the CC BY 4.0 license, included in~\cite{Tola&2023a}.}
    \label{fig:robocim_uml}
\end{figure}

We briefly describe how the configuration of devices and their interfaces are modeled for applications where uncertain information is prevalent, based on the developed UML diagram shown in \cref{fig:robocim_uml}.
The universal rules describe multiple concepts of configuration, including products, product series, and ports.
A configuration consists of at least two products, which in the context of robot systems can be devices such as robotic arms or end-effectors.
A product can belong to a product series, which describes common characteristics of a group of similar products.
An example could be to create a product series of robotic arms to describe common structural characteristics, such as possessing a mechanical interface of the type robot flange.
Products and product series are defined as port containers, meaning they possess at least one port.
A port can be used to associate a port container with information through attributes.
These can be in different forms, such as the mechanical, electrical, or software interfaces of a product, or specifications regarding a product series.

\begin{description}
    \item[Configuration:]

                A configuration composes products that connect with each other through compatible ports, that is, have the same port attributes with opposite orientation.
                Each product port can be connected to at most one other port of another product, and each input/output port of a product must be used for a configuration to be valid.
                Examples of constraints, using \acrshort{ocl} are shown in \cref{lst:ports}.
                Constraint~1 describes that a port cannot be connected to itself, and constraint~2 describes that a port must have an orientation, i.e., be an input or output, in order to connect with other ports.

    \item[Compatibility:] One of the main compatibility constraints is that two ports are compatible \textit{if} they have the same port type with different                orientation, \textit{and} there is no  explicit evidence that they are incompatible. 
                Explicit evidence of incompatibility may be defined through additional information, i.e., such as constraint~4 in \cref{lst:compatibility}, which adds information that, \textit{if} two products are directly incompatible \textit{then} their ports are also incompatible.
                Constraint~5 in \cref{lst:compatibility} describes that \textit{if} two products are incompatible, \textit{then} they must not exist in the same configuration.
                This constraint solves the issue in challenge \textit{CH~2A.1}, and can be extended to more than two products to also solve \textit{CH~2A.3}.
                Although \textit{CH~2A.2} is also associated with incompatibility, it cannot be solved with constraint~5, as these two products \textit{can} exist in the same configuration, but \textit{cannot} be directly connected, and is therefore solved using constraint~4.

    \item[Justifications:] To reason about the source of information, we define justifications.
                These are used when defining attribute values. 
                We defined four main justification types, ordered by strength:
                \begin{description}
                    \item[Primary:] can be evidence provided by the \acrshort{oem} of the product in the form of a data sheet or user manual. 
                
                    \item[Empirical:] can be evidence provided by physically experimenting with the products, for example, by attaching them.
                
                    \item[Secondary:] can be evidence provided by another \acrshort{oem}, as in the example of \textit{CH~2A.4}.
                
                    \item[Observation:] can be evidence provided by a domain expert that makes assumptions based on their expertise.
                \end{description}
                Using justifications, users have the flexibility to configure the RoboCIM solver in a way that mandates compatibility to be supported by primary sources, resulting in the provision of strongly justified configurations, as demonstrated in \cref{lst:justification}.
\end{description}
\begin{lstlisting}[numbers=none, label={lst:ports}, caption={Examples of constraints on ports in the universal rules, described using \acrshort{ocl}, modified from~\cite{Tola2021} under the CC BY 4.0 license.}]
--Constraint 1: A port cannot be connected to itself:
context Port inv: self.connected <> self

--Constraint 2: A port must have an orientation:
context Port inv: self.attribute.name = 'input' or self.attribute.name = 'output'

--Constraint 3: A port can connect to another port, if they have opposite orientation and the same interface (represented by attribute value)
context Port inv: 
    self.connected(p2) implies 
        Set{self.attribute.name} union Set{p2.attribute.name} = Set{'input', 'output'}
    and
        self.attribute.name.value = p2.attribute.name.value
\end{lstlisting}

\begin{lstlisting}[numbers=none, label={lst:compatibility}, caption={Examples of compatibility related constraints described using \acrshort{ocl}, modified from~\cite{Tola2021} under the CC BY 4.0 license.}]
--Constraint 4: Two directly incompatible products cannot be connected:
context Configuration inv:
    self.products->forAll(p1,p2 | p1 <> p2 and incompatible_neighbours(p1,p2) implies
        p1.ports->forAll(port1 | p2.ports->forall(port2 | port1.connected <> port2)))

--Constraint 5: Two products (productA and productB), known to be incompatible, must not exist in the same configuration:
context Configuration inv: 
    self.products->excludes('productA') and self.products->excludes('productB')
\end{lstlisting}

\begin{lstlisting}[numbers=none, label={lst:justification}, caption={Example of justification related constraint described using \acrshort{ocl}, modified from~\cite{Tola2021} under the CC BY 4.0 license.}]
--Constraint 6: Configurations using information from primary justifications only:
context Configuration inv:
    self.products->forAll(p1 | p1.attribute.value.justification = primary)
    and
    self.products->forAll(product->forAll(port | port.attribute.value.justification = primary))
\end{lstlisting}

\subsubsection{Application Rules}
These rules build on top of the universal rules by formalizing the domain knowledge within robot systems and ensuring and incorporating application requirements in the configuration.
An example of one of these rules is that a robot system must as a minimum consist of a robotic arm, an end-effector, an \acrshort{eecd}, and a data connection, as illustrated by constraint~7 in \cref{lst:robot_components}.
Another example is to specify required ports of products, e.g., a robotic arm must have a port of the type \textit{robot flange} (see constraint~8 in \cref{lst:robot_components}).
Rules to take user requirements into account are defined in constraint~9 in \cref{lst:robot_components}, where the application chosen restricts the types of end-effectors that can be incorporated in the configuration.

\begin{lstlisting}[numbers=none, label={lst:robot_components}, caption={Application rules on required product and port types in a configuration described using \acrshort{ocl}, modified from~\cite{Tola2021} under the CC BY 4.0 license.}]
--Constraint 7: One product of each type robotic_arm, eecd, end_effector, and data_connection, must exist in the configuration:
context Configuration inv:
    let product_types_attr = self.products->forAll(
                                        p->select(attribute | attribute.name = 'type')) in
        product_types_attr->one(value = 'robotic_arm') and
        product_types_attr->one(value = 'eecd') and
        product_types_attr->one(value = 'end_effector') and
        product_types_attr->one(value = 'data_connection')
    
--Constraint 8: Products of type robotic_arm must have specific port types:
context Product inv:
    self.attributes->exists(attribute | attribute.name = 'type' and attribute.value = 'robotic_arm') implies
    self.ports->exists(port | port.type = 'robotic_arm_flange')

--Constraint 9: User requirement on application
context Configuration inv:
    req_application <> none and self.applications->exists(req_application) implies
    let end_effector_type = self.applications->select(req_application).end_effector_type in
    let end_effector_product = self.products->select(p | p.attributes->exists(attribute | attribute.name = 'type' and attribute.value = 'end_effector')) in 
    end_effector_product.attributes->exists(attribute | attribute.name = 'subtype' and attribute.value = end_effector_type))
\end{lstlisting}

\subsubsection{Default Reasoning}
When developing configurators, it is customary to make assumptions about the general properties and compatibility of products.
However, there are instances where these initial assumptions about a product may prove to be incorrect, as the challenges in \textit{CH~2A} showed.
Creating a program that can reason about these assumptions that are subject to change in the face of new evidence can be complicated, and thus default reasoning can be used.

Default reasoning is typically associated with non-monotonic logic, where new information can lead to the revision of previously drawn conclusions.
This is the opposite of monotonic logic, where once a conclusion is reached, it remains valid regardless of additional information.
An example of using default reasoning can be defining that birds can fly, which is a valid claim until additional information that the specific bird in context is a penguin, then this claim can be overridden with the fact that this specific bird cannot fly~\cite{REITER198081}.

This idea is used in RoboCIM, which rather than treating compatibility between devices as \emph{known facts}, RoboCIM treats compatibility as a \emph{claim} that can be supported by evidence in the context of a rational, logic-based argument.
Similarly, incompatibility is treated as a claim that can be supported by evidence.
Valid configurations are then inferred based on (in)compatibility relations between components.
In RoboCIM, default reasoning can for example be used to claim that two devices with the same interface are compatible, unless additional evidence on these specific devices shows the opposite.
This example aligns with the assumptions that would be made in challenge \textit{CH~2A.2}, where the additional information about the specific devices would override the initial assumption on compatibility.

\subsection{Validation}

The information model, RoboCIM, was validated by developing a prototype and testing it on twenty products from three different \acrshortpl{oem}.
As \acrshort{ocl} is difficult to execute, the programming language \acrfull{asp} was used to express the formalized domain model in an executable manner, with minimal translation efforts from \acrshort{ocl} and \acrshort{uml}~\cite{KBC_chapter6}.
Our choice of \acrshort{asp} is due to the fact that it is known to handle default reasoning in cases where information is missing~\cite{Lifschitz08}, and is capable of providing explanations that justify the proposals~\cite{ExplainableASP}.
The specific choice of \acrshort{asp} is an implementation detail, and is simply used to create an executable representation of the domain model for validation purposes. 
The information model was translated into an \acrshort{asp} program using the class diagram in \cref{fig:robocim_uml} and the \acrshort{ocl} constraints defined above.
The code can be found in the GitHub repository\footnote{\url{https://github.com/Daniella1/robocim_configurator}}, where the \acrshortpl{oem} and products are anonymized to protect their reputation.
The implemented configurator allows choosing the number of composed products in a configuration, and we ran experiments with compositions of four and compositions of five products.
In the case of five products, a flange adapter would be used between the robotic arm and the \acrshort{eecd}.
The results of running the configurator on the products are shown in \cref{tab:eval_results}, which illustrate that the less the amount of valid configurations found, the less time is used to execute the configuration.
The validity of the configurations was manually examined on a smaller subset of products.

\newcolumntype{R}[1]{>{\raggedleft\let\newline\\\arraybackslash\hspace{0pt}}m{#1}}
\begin{table}
\centering
\begin{tabular}{l|R{3cm}|R{3.2cm}|r|r}
\textbf{Application} & \textbf{\#Products in Catalog} & \textbf{\#Products in Configuration} & \textbf{\#Configurations} & \textbf{Time [s]} \\ \hline
any & 20 & 4 & 122 & 0.255 \\ \hline
any & 20 & 5 & 11 & 0.094 \\ \hline
pick-and-place & 20 & 4 & 115 & 0.250\\ \hline
screwdriving & 20 & 4 & 7 & 0.050 \\ \hline
\end{tabular}
\caption{Results of generating valid configurations using \acrshort{asp}, modified from~\cite{Tola2021} under the CC BY 4.0 license.}
\label{tab:eval_results}
\end{table}

\subsection{Summary and Discussion}

RoboCIM captures \emph{evidence sources} as an explicit concept that can be associated with assertions on compatibility. 
This permits a range of new queries about configurations: 
\begin{itemize}
    \item find configurations where the certainty threshold of validity is high, i.e., where all compatibility relationships are based on strong sources of evidence;
    \item infer tentative configurations by reasoning about compatibility \emph{by default} in cases where there is a lack of information, e.g., by default, assume two particular kinds of components are compatible if there is no strong evidence of incompatibility;
    \item identify those components for which compatibility is uncertain and where information gathering would be a valuable exercise, i.e., where the evidence for, or against, a claim of compatibility is lacking.
\end{itemize}
RoboCIM is structured to integrate information about particular products, even if they are not currently included in the product catalog. 
This functionality enables users to gradually expand their knowledge base regarding robotic products and subsequently adding these products to the product catalog.

The limitations of the approach are:
\begin{itemize}
    \item extensibility of the prototype is not properly tested, as it has been tested on a limited product catalog of only 20 products, meaning there may appear other types of challenges where the domain model would need to be adapted. Additionally, sensors and other types of devices have not been tested;
    \item scalability has not been tested, and with the current results of over 100 configurations, it may be difficult for end-users to choose a configuration. Furthermore, adding costs of products and user requirements thereof may reduce the number of configurations;
    \item meeting safety and environmental standards have not been considered in the current version of the configurator, although these are crucial when developing robot systems;
    \item the domain model has only been tested in an academic setting, and not in an industrial setting, therefore improvements can potentially be made through industrial validations, as illustrated in \cref{fig:robocim_research_method};
\end{itemize}

\noindent This chapter addressed \textbf{RQ2} by illustrating how a configurator for modular robot systems \textit{can} be developed, that also considers compatibility of devices and takes a limited number of application requirements into account.

\begin{contr}{Contribution 3 (C3):}
Proposed a formal approach for constraint-based configuration of robot systems to address a subset of the identified challenges.
\end{contr}

\section{Potential Directions}

Potential future work in this area could be:
\begin{description}
    \item[Integration and re-purposing metrics:] properties related to the difficulty of integrating or re-purposing a robot system can be included as a metric in the configurator.
    For example, properties such as ease-of-programming of devices can be calculated.

    \item[Automated CAD assembly:] recent advances in automating the process of assembling CAD models can be used for checking the spatial constraints and compatibility of the devices in a robot system~\cite{AutoMateDatasetForCADAssemblies,LearningRoboticAssemblyFromCAD}.

    \item[Natural language processing:] to reduce the amount of work in the development of the knowledge base, natural language processing may be used to extract relevant information from data sheets of devices~\cite{NLP}.

    \item[Answer Set Programming:] has been shown by previous studies that it can be used for knowledge representation of mid-sized configuration problems containing hundreds of products~\cite{TwentyFiveYearsConfiguration}.
        Thus, it is important to analyze the number of products that would potentially be included in such a product catalog for robot systems, and which type of constraint-programming tool would be the most suitable to use in future developments.

\end{description}

\chapter{Modeling and Visualization} \label{chapter:visualization}

\backgroundsetup{
  scale=1,
  color=black,
  opacity=1,
  angle=90,
  position=current page.north,
  vshift=-250mm,
  hshift=-10mm,
  contents={
    \textcolor{gray}{\rule{5mm}{\paperheight}}
  }
}

After configuring a robot system, it is appropriate to visually illustrate how such a system may perform in a manufacturing line. 
Creating models of the configured robot devices and combining them into a complete robot system for visualization or simulation purposes is essential in the process of robotic systems integration.
The visualizations and simulations can be used during the design and configuration phase and in the operational phase, potentially in the context of \acrshortpl{dt}.
Different types of tools are suitable for visualizing and simulating different phases of robotic systems integration.
Thus, creating robot models that can be reused across different tools is crucial for time saving and reducing the overall integration costs.
Additionally, using and advancing such an interoperable format reduces the negative effects of vendor lock-in by allowing flexibility of the choice of simulation tool.
This chapter briefly introduces robot description formats, and goes into detail with the interoperable and widely used \acrfull{urdf}, outlining the advantages and current limitations of the format.
This chapter is based on the research output from publications~\cite{Tola&2023d} and~\cite{Tola&2023b}.

\section{Research Method}
The goal of this chapter is to answer \textbf{RQ3} that is defined as \textit{what are the advantages and complications of defining an interoperable robot modeling format for visualization and simulation of robots?}

This chapter focuses on \acrshort{urdf}, as it is one of the most commonly used formats for modeling robots and has the potential to fulfill the requirements of \textbf{RQ3}.
Therefore, understanding its usage and limitations is essential.
To address \textbf{RQ3}, which is descriptive, we conduct inferential research, based on a user survey, an analysis of a dataset we manually created, and through empirical data obtained through discussions, online information, and our own experiences.
The goal is to identify challenges and advantages that users experience with \acrshort{urdf}, find commonalities and discrepancies across \acrshort{urdf} files, and support this information with empirical data, thus bringing three types of ``evident knowledge'' from the users' perspectives, from factual data, and through experiences and discussions.
These results are used to highlight the main advantages, challenges, and potential future directions for improving the format.

The survey, with over 500 participants, included demographic and survey questions to ensure \acrshort{urdf} users with diverse backgrounds were represented.
Moreover, survey best practices to reduce bias, as outlined in~\cite{Kitchenham&95}, were adhered to, and threats to validity were assessed and detailed in the publication~\cite{Tola&2023d}.
The survey questions, recruitment material, and anonymous responses are all publicly available on GitHub\footnote{\url{https://github.com/Daniella1/urdf_survey_material}}.

The dataset, with~322 \acrshortpl{urdf}, was sourced from five categories to minimize bias: \acrfull{ros}-related sources, commercialized tools, \acrshortpl{oem}, common tools used by roboticists, and various repositories that users commonly encounter when searching for \acrshortpl{urdf}. 
All these files are publicly available on GitHub\footnote{\url{https://github.com/Daniella1/urdf_files_dataset}}, along with materials for reproducing the results\footnote{\url{https://github.com/Daniella1/urdf_dataset_results_material}} and analyzing other \acrshort{urdf} Bundles\footnote{\url{https://github.com/Daniella1/urdf_analyzer}}.
More details on the construction of the dataset can be found in the publication~\cite{Tola&2023b}.

Empirical data was gathered through online platforms, such as the \acrshort{ros} wiki and \acrshort{ros} Discourse, through discussions with roboticists from the Silicon Valley Robotics community and from a \acrshort{urdf} working group created based on the survey results\footnote{\url{https://discourse.ros.org/t/short-term-wg-about-urdf-description-formats/30843}}, 
These discussions, online user perspectives and information, and our own experiences with the format collectively serve as empirical evidence.


\backgroundsetup{
              scale=1,
              color=black,
              opacity=1,
              angle=90,
              position=current page.north,
              vshift=+250mm,
              hshift=-10mm,
              contents={
                \textcolor{gray}{\rule{5mm}{\paperheight}}
              }
            }

\section{Background}

    \subsection{Modeling, Simulation, and Visualization}
    \subsubsection{Modeling}
    To analyze systems using analytical tools, we require models of these systems.
    A model of a system is defined by J. Van Amerongen as~\cite{dynamicalSystems:Job:2010} \textit{``a simplified description of a system, just complex enough to describe or study the phenomena that are relevant for our problem context''}.
    Thus, both the system characteristics and fidelity of the model should be related to its purpose.
    Creating a model of a system can be a complicated process, depending on the type of model and the system complexity.
    In the context of robot systems, models can be used to describe different characteristics of a device, such as the kinematics, dynamics, or the controller.
    The kinematics and dynamics describe physical attributes of a robot, where kinematics focuses on geometry and motion \textit{without} considering the masses or forces, while they are included in the dynamics~\cite{PeterCorkeRoboticsVisionAndControl}.

    This thesis focuses on kinematic models of robots, where their geometrical positions are defined and mainly used for visualization purposes.
    The kinematics of a robotic arm describe the position and orientation of the links and joints with respect to each other, such as the robot illustrated in \cref{fig:cylindrical_vs_kuka_annotated_all}.
    Creating a kinematic model of a robot requires knowledge about its link size, position, and orientation, and its joint type, position, and orientation.
    Most of these values can be approximated by measuring the physical robot itself, although some values may be provided by the \acrshort{oem}, achieving greater model accuracy.

    \subsubsection{Simulation and Visualization}
    Visualization is mainly related to replicating the geometric aspects and information about a system through rendering, while simulation is typically used for studying the behavior of a system by predicting its state over time.
    Simulation enables digital experimentation of designs and parameters before investing in expensive hardware.
    It is extremely common nowadays to include visualizations in simulation tools, as they help people understand the system being modeled and its behaviors~\cite{SeeingIsBelieving}.

    In the context of robotic systems integration, visualization can be used in the design and configuration phase, where parts of the system layout can be validated using visualizations, and in a potential configurator where the customers can easier understand how such a system may operate.
    Simulation can be used in the design phase for validation of the system behavior and in the operational phase to optimize and maintain the quality of the manufacturing system. 
    The same robot models can potentially be used across different phases of robotic systems integration depending on their fidelity and the purpose of the simulation.
    Additionally, the fidelity of the models can be increased over time, thus reusing parts of the initial model.
    For example, the same models used for visualization in a configurator can potentially also be used for visualization in the context of a \acrshort{dt}; see \cref{chapter:digital_twins} for more information.
    Simulation tools are typically specialized for simulating specific types or aspects of systems, and therefore depending on the purpose, different simulation tools may be used in different phases of robotic systems integration.
    To ensure interoperability and avoid vendor lock-in, we choose to focus on modeling formats that can be reused across robot manufacturer-independent simulation tools.
    
    \backgroundsetup{contents={}}

    \subsection{Robot Description Formats}
    
    Some of the commonly used simulation tools within robotics are Gazebo\footnote{\url{https://gazebosim.org/home}}, CoppeliaSim\footnote{\url{https://www.coppeliarobotics.com/}}, PyBullet\footnote{\url{https://pybullet.org/wordpress/}}, MuJoCo\footnote{\url{https://mujoco.org/}}, Drake\footnote{\url{https://drake.mit.edu/}}, and NVidia Isaac\footnote{\url{https://developer.nvidia.com/isaac-sim}}~\cite{RobotSimulators}, which each have their own native model format, complicating the process of exchanging models between such tools.

    Four common description formats, that are used in robot modeling, are \acrshort{urdf}~\cite{wikiROSURDF}, \acrfull{sdf}~\cite{SDFformat}, \acrfull{usd}~\cite{USDformat}, and \acrfull{mjcf}~\cite{MJCFormat}. 
    \acrshort{urdf} specifies kinematic and (some) dynamic properties of a single robot, while \acrshort{sdf} builds on top of that with friction, physics, and sensors, and by specifying objects surrounding the robot, such as lighting, surface properties, and textures.
    Both of these formats are based on the \acrfull{xml}, and allow modeling a robot with or without 3D meshes.
    A \acrshort{usd} file is either binary or text-based and is used to describe 3D scenes. 
    It can store 3D assets in one file, where schemas allow extending the 3D model with features such as lighting, shading, and physics.
    \acrshort{mjcf} is also an \acrshort{xml}-based format that can describe robot kinematics and dynamics, and the surroundings.
    
    To choose the best format for modeling devices in a robot system, we compare each format with regards to the following criteria, which were inspired from~\cite{IvanouRDF}:
    \begin{description}
        \item[Wide use:] of the format, means more support, more simulation tools that the format can be used in, and potentially active community resulting in a well developed format.

        \item[Composability:] allows describing a separate module, i.e., a robotic device, and enables combining these modules into complex robot systems.
        This allows reuse of devices within different robot systems.

        \item[Well-established file format:] makes it easier to implement tools that can parse the model, if many programming languages support the file format.
        Furthermore, if it is human-readable, then it is easier for users to manipulate.
    \end{description}

    \acrshort{urdf} is supported by more simulation tools than the other formats, see \cref{tab:simulators_model_support}, indicating its high interoperability~\cite{IvanouRDF}.
    \acrshort{urdf} has been around since~2009, compared to the other formats that are more recent, \acrshort{sdf} since~2012, \acrshort{usd} since~2016, and \acrshort{mjcf} since~2021.
    Although \acrshort{urdf} has been around for many years, tools such as MATLAB and Unity have only recently started supporting the format.
    MATLAB supported the importing of \acrshort{urdf} files in their version R2018b, and exporting of \acrshort{urdf} files in their latest version, which is R2023b.
    Unity started supporting the importation of the format in late 2019.
    This shows that there is great interest in continuously using \acrshort{urdf}.
    
     \begin{table}
    \centering
    \begin{tabular}{l|cccc}
    \textbf{Simulation tool} & \textbf{\acrshort{urdf}} & \textbf{\acrshort{sdf}} & \textbf{\acrshort{usd}} & \textbf{\acrshort{mjcf}} \\ \hline
    CoppeliaSim~\cite{CoppeliaSDFURDF}  & \greencheck & \greencheck & \redcross & \redcross  \\ \hline
    Drake~\cite{DrakeFormats} & \greencheck & \greencheck& \redcross & \greencheck \\ \hline
    Gazebo~\cite{SDFsupportUSDMJCF}  &  \greencheck~\cite{SDFsupportURDF} & \greencheck~\cite{SDFformat} & (\greencheck) & (\greencheck) \\ \hline
    MATLAB~\cite{MatlabFormatSupport} & \greencheck  & \greencheck  & \redcross  & \redcross \\ \hline
    MuJoCo~\cite{MJCFormatURDF} &  \greencheck & \redcross & \redcross  & \greencheck \\ \hline
    NVidia Isaac~\cite{IsaacSimFormats} &  \greencheck & \redcross & \greencheck & \greencheck~\cite{IsaacSimFormatsMJCF}\\ \hline
    PyBullet~\cite{PybulletURDFSDFMJCF} & \greencheck & \greencheck & \redcross & \greencheck \\ \hline
    RoboDK~\cite{RoboDKDocumentation} &  \redcross & \redcross & \redcross & \redcross \\ \hline
    Robotics Toolbox for Python~\cite{RTBFormats} & \greencheck & \redcross & \redcross & \redcross \\ \hline
    RViz~\cite{wikiRViz} & \greencheck & \redcross & \redcross & \redcross \\ \hline
    Unity~\cite{UnityRoboticsHub}  &  \greencheck~\cite{UnityURDF} & \redcross & \redcross & \redcross \\ \hline
    Webots~\cite{WebotsFormatOther}  & (\greencheck)~\cite{WebotsFormat} & \redcross & \redcross & \redcross  \\ \hline
    \textbf{Total} \hspace{4.3cm} 12 &  11 &  5 & 2 &  5 \\ \hline 
    \end{tabular}
    \caption{Simulators and the formats they support (as of 10/2023), adapted from~\cite{Tola&2023b}. The parentheses indicate that tools exist for converting the format into a supported one. The references on the left provide information that all of the checkmarks and crosses in the same row are derived from, except for the checkmarks with references right beside them.} 
    \label{tab:simulators_model_support}
    \end{table}

    Of these four formats, \acrshort{urdf} is the most basic, modeling the robot by itself.
    All of the four formats allow composition of devices.
    However, the abstraction level of \acrshort{urdf} models is extremely suitable, as it allows simple composition of each device in a robot system, similar to the system composition in \cref{fig:configuration}.
    The simplicity of \acrshort{urdf} modeling a single robotic device means that its level of fidelity is appropriate for visualizations and simple simulations across different phases of robotic systems integration. 
    \acrshort{sdf} contains more details about the robots and their surroundings, which may not be necessary in all situations and can potentially complicate composing device models into a system.
    However, in some cases, such as the operational phase, this may prove beneficial.

    \acrshort{urdf}, \acrshort{sdf}, and \acrshort{mjcf} are all based on \acrshort{xml}, meaning their file format is well-established, making it easier to import, export, and manipulate these formats in different programming languages as most provide libraries supporting \acrshort{xml}.
    Furthermore, \acrshort{xml} is human-readable, meaning humans can with limited effort read the file, understand it, and manipulate relevant parameters when needed.

    \begin{table}
    \centering
    \begin{tabular}{l|cccc}
    \textbf{Requirements} & \textbf{\acrshort{urdf}} & \textbf{\acrshort{sdf}} & \textbf{\acrshort{usd}} & \textbf{\acrshort{mjcf}} \\ \hline
    Wide use    &  \greencheck    &  \redcross   &   \redcross   &  \redcross   \\ \hline
    Composability &  \greencheck    &  \greencheck   &   \greencheck   &  \greencheck   \\ \hline
    Well-established file format &  \greencheck    &  \greencheck   &   \redcross    &  \greencheck  \\ \hline
    \end{tabular}
    \caption{Comparison of robot description formats, partially based on~\cite{IvanouRDF} and \cref{tab:simulators_model_support}.} 
    \label{tab:format_comparisons}
    \end{table}

    As \cref{tab:format_comparisons} illustrates, \acrshort{urdf} is both composable, widely used, and is based on the well-established file format, \acrshort{xml}.
    Additionally, as one of the constraints of the PhD project was to use Unity for visualizations, it is necessary that the chosen format can be used with this.
    Therefore, \acrshort{urdf} is chosen to be investigated in this thesis.
    Both \acrshort{urdf} and \acrshort{sdf} are open-source, and tools for converting \acrshort{urdf} to \acrshort{sdf} exist.
    Thus, \acrshort{urdf} files can be used across different phases of robotic systems integration, and if further information about the robots is needed, the \acrshort{urdf} models can be converted into \acrshort{sdf} and extended.

\section{Unified Robot Description Format (URDF)}
This section introduces \acrshort{urdf} in more detail, and is based on the research output from publications~\cite{Tola&2023d} and~\cite{Tola&2023b} and empirical data.
We present its main advantages and limitations that need to be addressed to be able to use this format in robotic systems integration.

    \subsection{Overview}
    \acrshort{urdf} was introduced in 2009 by the developers of \acrshort{ros} to be a self-contained description format with all the necessary modeling parameters in a single file~\cite{Quigley&2015}. 
    Although the format was initially introduced in \acrshort{ros}, its ease-of-support through mature \acrshort{xml} parsers and its standalone characteristics have allowed it to become universally used outside of the \acrshort{ros} ecosystem, as highlighted in \cref{tab:simulators_model_support}.

    A \acrshort{urdf} file is based on \acrshort{xml} and has the extension \textit{.urdf}.
    It can describe the kinematic structure, dynamic parameters, visual representation, and collision geometries of a robot.
    The file itself is standalone but may refer to external 3D geometries representing links of the robot.
    To create a \acrshort{urdf} file, the minimal requirements are the name of the robot and a link.
    An example of a partial \acrshort{urdf} file is shown in \cref{lst:urdf2dof}, which represents a robot with two joints, and three links visualized using boxes and cylinders, depicted in \cref{fig:2dof_robot_urdf}.
    
    \definecolor{baselinkcolor}{rgb}{0.96,0.88,0.76}
    \definecolor{link1color}{rgb}{0.93,0.93,0.95}
    \definecolor{link2color}{rgb}{0.86,0.93,1}
    \definecolor{joint1color}{rgb}{0.96,0.93,0.89}
    \definecolor{joint2color}{rgb}{0.8,0.85,0.97}
    \definecolor{others}{rgb}{0.76,0.74,0.82}
    \newcommand{\coloropacity}{!65}%
    
    \newcommand{\Hilight}[1]{\makebox[0pt][l]{\color{#1}\rule[-4pt]{\columnwidth}{9pt}}} 
    \lstset{style=xmlStyle,escapechar=|}
    
    \begin{minipage}{.4\textwidth}
        \centering
        \includegraphics[width=0.65\textwidth]{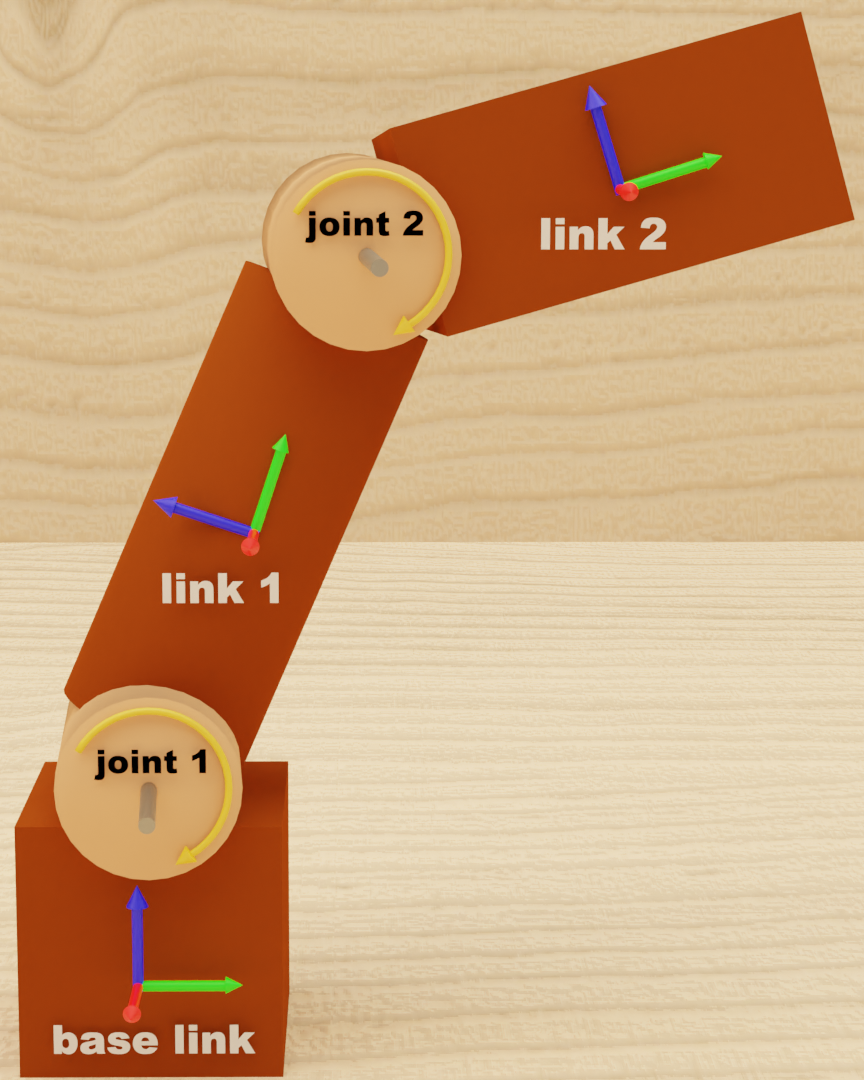}
        \captionof{figure}{Visualization of robot defined in Listing \ref{lst:urdf2dof}. This figure is included in~\cite{Tola&2023b}.}
        \label{fig:2dof_robot_urdf}
    \end{minipage}
    \hfill
    \begin{minipage}{.4\textwidth}
\begin{lstlisting}[caption={URDF file contents of a planar robot, included in~\cite{Tola&2023b}.},label={lst:urdf2dof},style=xmlStyle]
|\Hilight{others\coloropacity}|<?xml version="1.0" encoding="utf-8"?>
|\Hilight{others\coloropacity}|<robot name="2 DOF planar robot">
|\Hilight{baselinkcolor\coloropacity}| <link name="base link">
|\Hilight{baselinkcolor\coloropacity}\hspace{0.2cm}|  <visual>
|\Hilight{baselinkcolor\coloropacity}\hspace{0.4cm}|   <origin xyz="0 0 0.25"/>
|\Hilight{baselinkcolor\coloropacity}\hspace{0.4cm}|   <geometry>
|\Hilight{baselinkcolor\coloropacity}\hspace{0.65cm}|    <box size="0.5 0.5 0.5"/>
|\Hilight{baselinkcolor\coloropacity}\hspace{0.4cm}|   </geometry>
|\Hilight{baselinkcolor\coloropacity}\hspace{0.2cm}|  </visual>
|\Hilight{baselinkcolor\coloropacity}| </link>
|\Hilight{baselinkcolor\coloropacity}| ...
|\Hilight{joint1color\coloropacity}| <joint name="joint 1" type="continuous">
|\Hilight{joint1color\coloropacity}\hspace{0.2cm}|  <parent link="base link" />
|\Hilight{joint1color\coloropacity}\hspace{0.2cm}|  <child link="link 1" />
|\Hilight{joint1color\coloropacity}\hspace{0.2cm}|  <axis xyz="0 1 0" />
|\Hilight{joint1color\coloropacity}\hspace{0.2cm}|  <origin xyz="0 0 0.5"/>
|\Hilight{joint1color\coloropacity}| </joint>
 ...
|\Hilight{others\coloropacity}|</robot>
\end{lstlisting}
    \end{minipage}
    
    \subsubsection{Links}
    Links are rigid bodies or parts that are connected through joints\footnote{\url{wiki.ros.org/urdf/XML/link}}.
    They can be described using inertial properties which depict the link's mass, the position of the center of mass, and the moments and products of inertia.
    Links are also described with visual, and collision properties which are detailed below.
    \acrshort{urdf} links can only be represented using rigid bodies and not deformable ones, thus limiting its use for soft robotics.

    \subsubsection{Joints}
    Joints are connectors between two links: a parent and child link\footnote{\url{wiki.ros.org/urdf/XML/joint}}.
    The parent link is closer to the base of the robot, and the child is closer to the tip.
    \acrshort{urdf} supports joints of types revolute, continuous, prismatic, fixed, floating and planar.
    Continuous joints are similar to revolute joints but without any motion limits.
    Joint properties that can be specified are the type (kinematics), dynamics, and safety limits.

    \subsubsection{Visual and Collision Geometries}
    The links of a robot are represented using geometric objects, called meshes, and can be used for visualization or collision purposes.
    There are different \acrfull{cad} file formats that can be used for representing meshes, and in \acrshort{urdf} the most common ones are STL, COLLADA, and OBJ.
    Each file format has its own advantages and limitations, and should be chosen depending on its application.
    STL has the file extension \textit{.stl}, and represents 3D surface geometries using only triangles with no texture or color information.
    COLLADA has the file extension \textit{.dae}, and supports both color and texture information, thus it is commonly used for visualization purposes.
    OBJ has the file extension \textit{.obj}, and supports free-form curves, meaning it has a higher level of detail compared to the other two formats, and also supports color and texture properties in a separate \textit{.mtl} file.

    Collision geometries are used for detecting collisions during a simulation.
    Depending on the level of fidelity for the collision detection, the geometries may be simple cylinders, or they may be 3D meshes.
    It is common to use both visual and collision meshes in \acrshort{urdf} Bundles, where COLLADA and STL can be used respectively.

    \subsubsection{URDF Bundle}
\begin{figure}
        \centering
        \includegraphics[width=0.6\columnwidth]{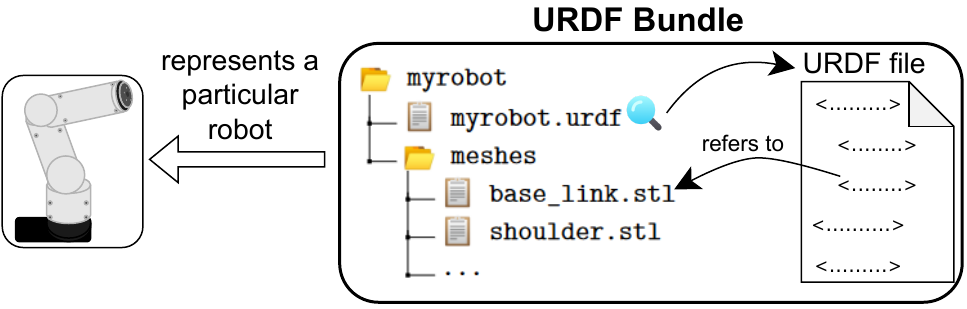}
        \caption{Example of a \acrshort{urdf} Bundle, included in~\cite{Tola&2023b}. It is common for mesh files to be located in a separate folder.}
        \label{fig:urdf_bundle}
    \end{figure}

    As a \acrshort{urdf} file is standalone, but can refer to external mesh files for visualization and collision, we define such a collection of files as a \acrshort{urdf} Bundle, see \cref{fig:urdf_bundle}.
    The example shows that the \acrshort{urdf} Bundle contains a \acrshort{urdf} file \textit{myrobot.urdf}, and a number of mesh files that represent the links of the robot.

    \subsubsection{Xacro}
    Generating \acrshort{urdf} Bundles is commonly done using the \textit{xacro} preprocessor from \acrshort{ros}, that is a macro language for \acrshort{xml}\footnote{\url{wiki.ros.org/xacro}}.
    Xacro allows using tags to configure the \acrshort{urdf} file with parameters through operations, such as variable subsitution and math calculations, that help reduce redundancy and increase maintainability of the models~\cite{Albergo&2022}.
    
    \subsection{Understanding URDF} \label{sec:understanding_urdf}
    While creating \acrshort{urdf} Bundles, we encountered several difficulties due to limited documentation and other issues with the format.
    Thus, to better understand the advantages, challenges, and future perspectives of \acrshort{urdf}, we conducted a user survey, created and analyzed a dataset, and used empirical data to identify these key points.
    Based on these three methods, this section provides answers to a number of questions, together with key points indicated by \kp{\#}, where \# specifies the key point number.

    \subsubsection{Who Uses URDF and How?}    
    Based on~\cite{Tola&2023d}, \acrshort{urdf} is used by roboticists with different levels of experience, with the majority having~1-5 years of experience.
    The survey participants (out of 497 responses) reported using \acrshort{urdf} both in academia (82\%) and the industry (52\%), and (out of 472 responses) in the domains of manufacturing (46\%), transportation (24\%), and agriculture (14\%).

    ~\cite{Tola&2023d} indicates (out of 472 responses) that the most common types of robots modeled with \acrshort{urdf} are: robotic arm (81\%), mobile robot (64\%), end-effector (39\%), and dual arm robot (22\%), with the least modeled type being the delta robot (4\%).
    In~\cite{Tola&2023b}, the most common robot types were: robotic arm (60\%), humanoid robot (18\%), end-effector (7\%), and mobile robot (6\%).
    The main manufacturers used by the participants in~\cite{Tola&2023d} were (out of 472 responses): Universal Robots (54\%), Kuka (37\%), Franka Emika (26\%), and Robotiq (24\%).
    In~\cite{Tola&2023b}, the dataset showed different figures, with Fanuc (13\%), ABB (10\%), NASA (9\%), and Kuka (8\%) being the most common.
    It is worth noting that the numbers from~\cite{Tola&2023d} and~\cite{Tola&2023b} cannot be directly compared, as the survey responses allowed for multiple choices, while the dataset results are based on the fraction of robots within the complete dataset.
    Summarizing the robot types, the applications, and the main manufacturers, it is clear that:
    \begin{center}
        \kp{1} \textcolor{kpblue}{\acrshort{urdf} is widely used for modeling devices commonly found in modular robot systems.}
    \end{center}
    Interestingly, 16\% of the participants (out of 472 responses) in~\cite{Tola&2023d} reported using \acrshort{urdf} for modeling custom robots, indicating:
    \begin{center}
        \kp{2} \textcolor{kpblue}{an advantage of \acrshort{urdf} is ``easily'' creating interoperable models of custom robots.}
    \end{center}

    The majority of participants in P4 (out of 472 responses) use \acrshort{urdf} with 3D geometric visualizations (93\%), essentially as \acrshort{urdf} Bundles. Despite the considerable experience of many \acrshort{urdf} users with the format, there are differing opinions on whether \acrshort{urdf} is primarily used for visualization or for simulation purposes as well\footnote{\url{https://discourse.ros.org/t/urdf-improvements/30520/15}}, showing a:
    \begin{center}
        \kp{3} \textcolor{kpblue}{lack of transparency in the format.}
    \end{center}

    \subsubsection{How Are URDF Bundles Obtained?} \label{sec:urdf_obtained}
    As mentioned above, \acrshort{urdf} Bundles can be generated using xacro.
    In~\cite{Tola&2023d} (out of 361 responses),~90\% of the participants had used xacro to generate \acrshort{urdf} Bundles, while in~\cite{Tola&2023b}~95\% of the \acrshort{urdf} Bundles were generated using xacro.
    However, despite its common usage, xacro has been found to present several challenges~\cite{Albergo&2022}.

\begin{figure}
        \centering
        \includegraphics[width=\textwidth]{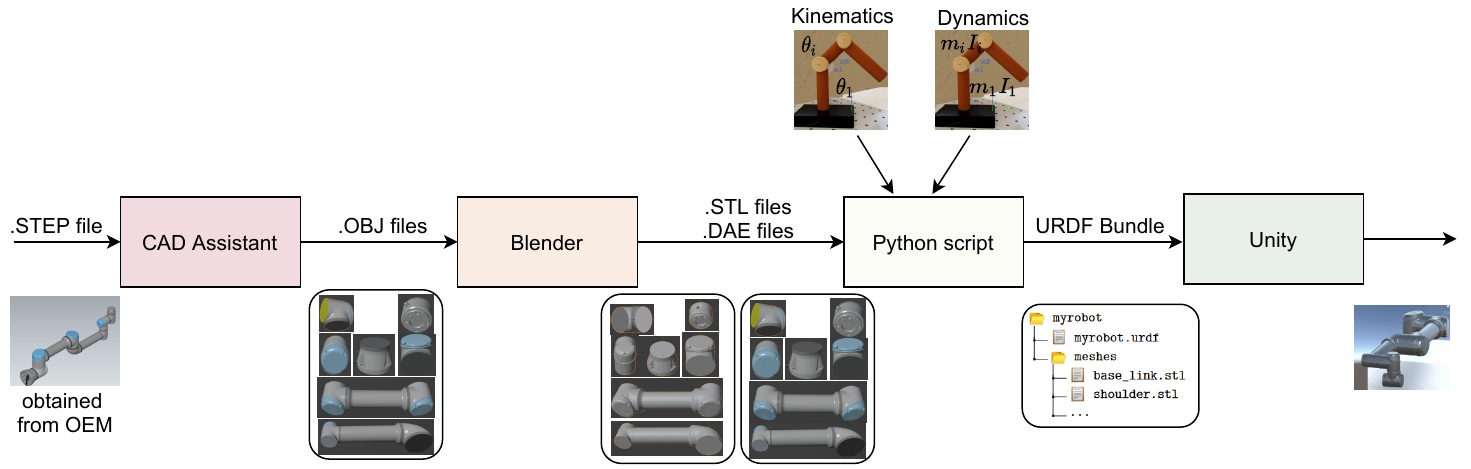}
        \caption{Illustration of how we created a \acrshort{urdf} Bundle from scratch.}
        \label{fig:urdf_steps}
    \end{figure}

    As shown in \cref{fig:urdf_steps}, four different programs (all available for personal use at no cost) were employed for the manual creation of a \acrshort{urdf} Bundle. CAD Assistant\footnote{\url{https://www.opencascade.com/products/cad-assistant/}} was used to separate the links, Blender\footnote{\url{https://www.blender.org/}} to modify the link origin and export to the desired \acrshort{cad} file formats, Python to generate the \acrshort{urdf} file and combine it with the meshes, resulting in the creation of the \acrshort{urdf} Bundle, and Unity\footnote{\url{https://unity.com/}} was used for visualizing the \acrshort{urdf}.
    The majority of the survey participants (out of 441 responses) in~\cite{Tola&2023d} knew xacro (82\%) and the SolidWorks exporter (57\%) which is a \acrshort{cad} tool, and the majority stated to have developed (out of 439 responses) \acrshort{urdf} Bundles by hand (67\%), using \acrshort{cad} tools (61\%), from \acrshort{ros} packages (59\%), and from the website of the \acrshort{oem} (47\%).

    Participants in~\cite{Tola&2023d} (out of 278 responses) described the most difficult parts of creating or modifying \acrshort{urdf} Bundles were related to:
    \begin{description}
        \item[Poses and meshes (21\%):] were difficult to define, especially:
        \begin{itemize}
            \item ensuring the accuracy of the link frames, 
            \item including colors in 3D meshes,
            \item ensuring correct file paths to 3D meshes, and
            \item matching the visual and collision meshes.
        \end{itemize}
        Additionally, some wished that the \acrshortpl{oem} provide separate 3D meshes for each link with correct frames.

        \item[Lack of tooling (19\%):] and difficult to debug, with requests for:
        \begin{itemize}
            \item intellisense and linters,
            \item a single tool to develop \acrshortpl{urdf}, debug, and preview the visualization, preferably with live editing, and
            \item a tool for validating the \acrshort{urdf} file or \acrshort{xml}. 
        \end{itemize}

         \item[Adjusting parameters (9\%):] to ensure the accuracy of the kinematics and dynamics was challenging, and the lack of support for parallel linkages, closed-chain systems, and deformable bodies limited the usage of the format.
    \end{description}

    Creating a \acrshort{urdf} Bundle clearly requires a great amount of effort, where different parts must be combined in an accurate manner, and various tools are used, illustrates the:
    \begin{center}
        \kp{4} \textcolor{kpblue}{need for a single platform that tackles the described challenges to simplify the workflow of creating and modifying \acrshortpl{urdf}.}
    \end{center}

    As \acrshort{urdf} Bundles can be obtained through online sources, like the ones in~\cite{Tola&2023b}, we analyzed some of these Bundles that were multiply defined across sources, see \cref{fig:multiply_defined_robot} and \cref{tab:duplicates_diff_robots}.
    The interesting results here are that~18\% of these \acrshortpl{urdf} had different forward kinematics, which is used to calculate the position of the tip of the robot from given joint positions, and is therefore highly related to the functionality of the \acrshort{urdf}.
    This issue coincides with responses from the survey in~\cite{Tola&2023d}, where 26\% (out of 440 responses) stated they have had issues with incorrect forward kinematics of \acrshort{urdf} Bundles.
    This shows that:
    \begin{center}
        \kp{5} \textcolor{kpblue}{the lack of standardization across \acrshort{urdf} Bundles results in inconsistencies of the models and reduces their accuracy.}
    \end{center}

    \begin{minipage}{\textwidth}
      \begin{minipage}[b]{0.49\textwidth}
        \centering
        \includegraphics[width=0.75\textwidth]{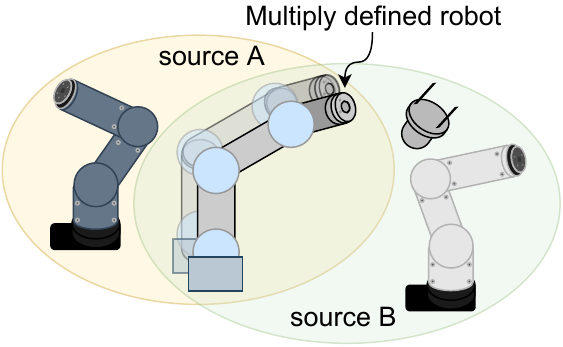}
        \captionof{figure}{Illustration of a multiply defined robot (middle) with two different models from two different sources of the dataset~\cite{Tola&2023b}.}
        \label{fig:multiply_defined_robot}
      \end{minipage}
      \hfill
      \begin{minipage}[b]{0.49\textwidth}
        \centering
        \begin{tabular}{l|c}
            \textbf{Feature discrepancies} & \textbf{\#Robots} \\ \hline
            number of joints & 9 \\ 
            number of links & 9 \\ 
            CAD file format & 6 \\ 
            forward kinematics & 11 \\
            any & 12 \\ \hline 
            \end{tabular}
          \captionof{table}{Feature differences across the, in total~60, multiply defined robots. \emph{any} indicates at least one difference was found. This table is modified from~\cite{Tola&2023b}.}
          \label{tab:duplicates_diff_robots}
        \end{minipage}
      \end{minipage}

    \subsubsection{What are the Challenges and Limitations of the Surroundings of URDF?}

    As stated above, there is a desire and need for tools to help improve the workflow of \acrshort{urdf}.
    A limited number of \acrshort{urdf}-related tools exist, and one of them is the \texttt{check\_urdf}\footnote{\url{http://wiki.ros.org/urdf\#Verification}}, developed by \acrshort{ros}, for validating \acrshort{urdf} files.
    We cross-checked the functionality of the tool with the \acrshort{ros} documentation\footnote{\url{http://wiki.ros.org/urdf/XML/robot}} on \acrshort{urdf}, and found the documentation states a joint is required in a \acrshort{urdf} file.
    However, when defining a \acrshort{urdf} file without any joints, and using \texttt{check\_urdf\footnote{Version 3.1.1: \url{https://anaconda.org/conda-forge/urdfdom}}} to validate it, we found that it successfully parsed the file, and did not report any warnings or errors on missing joint definitions, showing inconsistencies between the tool and the documentation.

    Furthermore, \acrfull{xsd} can be used to define the structure and constraints of \acrshort{xml} files, and such a file is defined for \acrshort{urdf}\footnote{\url{https://github.com/ros/urdfdom/blob/master/xsd/urdf.xsd}}.
    However, the \acrshort{xsd} has not been updated in the last ten years, although \acrshort{urdf} has, meaning these efforts to standardize the \acrshort{xml} are weak.
    Moreover, one of the developers of a \acrshort{urdf} parser reported finding many inconsistencies across the \acrshort{ros} documentation on \acrshort{urdf} and the original \acrshort{ros} parser.
    These examples indicate that:
    \begin{center}
        \kp{6} \textcolor{kpblue}{there are inconsistencies between the \acrshort{ros}-constructed documentation and \acrshort{ros}-developed tools associated with \acrshort{urdf}.}
    \end{center}

    By analyzing \acrshort{urdf} parsers created by different people and organizations, we found major discrepancies in their implementations, where \cref{tab:urdf_tool_parsing_results} shows that none of the parsers can successfully parse the same amount of \acrshort{urdf} files.
    These inconsistencies across the parsers may relate to the fact that~44\% of the survey participants (440 responses) experienced difficulties importing \acrshortpl{urdf} into simulation tools, indicating that:
    \begin{center}
        \kp{7} \textcolor{kpblue}{there are inconsistencies across the implementations of \acrshort{urdf} tools, which may be caused by an inherent lack of standardization and transparency, affecting the interoperability of the format.}
    \end{center}
    
    \begin{table}
    \centering
    \begin{tabular}{r|c|c|c|c|c|c}
    \textbf{Parsers}  & \textit{yourdfpy} & \textit{urdfpy} & \textit{pybullet} & \textit{robotics-toolbox} & \textit{MATLAB} & \textit{ROS parser}\\ \hline
    \textbf{\#URDFs}   & 220 & 7 & 252 &  306 & 309 & 311 \\ \hline
    \end{tabular}
    \caption{Results from running the~322 \acrshort{urdf} files in~\cite{Tola&2023b} through the listed parsers. The versions of the parsers can be found in~\cite{Tola&2023b}.}
    \label{tab:urdf_tool_parsing_results}
    \end{table}

    Over the years \acrshort{urdf} has been extended with mutations and tags by \acrshort{ros} and other tools.
    \acrshort{ros} extended the joints in \acrshort{urdf} to support mimic tags\footnote{\url{http://wiki.ros.org/urdf/XML/joint}} in the \acrshort{ros} Groovy distribution which was released in 2012.
    Not all simulation tools support this change, e.g., one of the limitations of the MATLAB parser is no support for mimic tags in joints\footnote{\url{https://se.mathworks.com/help/releases/R2023b/robotics/ref/importrobot.html}}.
    In another tool, urdfpy\footnote{\url{https://github.com/mmatl/urdfpy}}, it was found through experiments that the tool does not support path references to meshes where \textit{``package''} was in the path.
    For a user to learn this, they would need to check the complete documentation or experiment with each tool they use, which complicates the usability of \acrshort{urdf}.
    In another example, Drake describes some of the \acrshort{urdf} features it does not support, and presents its own extensions to \acrshort{urdf}\footnote{\url{https://drake.mit.edu/doxygen_cxx/group__multibody__parsing.html\#multibody_parsing_drake_extensions}}.
    Survey participants in~\cite{Tola&2023d} (out of 461 responses, 35\%) wished for a versioned standard, i.e., \acrshort{urdf} v1.0, \acrshort{urdf} v1.1, etc.
    This is also suggested by one of the original developers of \acrshort{urdf}\footnote{\url{http://sachinchitta.github.io/urdf2/}}.
    These examples and desires for improvement signify the:
    \begin{center}
        \kp{8} \textcolor{kpblue}{need for a versioned \acrshort{urdf} standard, where parsers and simulation tools can specify which version of \acrshort{urdf} they support, to increase the interoperability of the format.}
    \end{center}

    Most of the \acrshort{urdf} Bundles found in the dataset in~\cite{Tola&2023b} and used by the participants in~\cite{Tola&2023d} are models of generic robots from \acrshortpl{oem}.
    An analysis of the number of \acrshort{urdf} Bundles and information on the format provided by \acrshortpl{oem} is shown in \cref{tab:manufacturer_provide_urdf_and_search}.
    As illustrated, only~4/27 \acrshortpl{oem} provide information about \acrshort{urdf} on their website.
    Furthermore, kinematic and dynamic parameters of the robots are not always provided by the \acrshortpl{oem}, making it difficult to create accurate models, if these values must be determined by the \acrshort{urdf} creator.
    Additionally, as mentioned above in \cref{sec:urdf_obtained}, the \acrshortpl{oem} provide 3D meshes of the complete robot, i.e., with all the links attached together, but what is needed in \acrshort{urdf} is 3D meshes of the separate links with their origins accurately defined in the positions of the joints.
    Summarizing this information can imply that:
    \begin{center}
        \kp{9} \textcolor{kpblue}{as \acrshortpl{urdf} of robots by \acrshortpl{oem} are commonly used, it is reasonable for the \acrshortpl{oem} to provide more support by supplying kinematic and dynamic information about their robots, together with 3D meshes of separate links of each robot.}
    \end{center}

    \begin{table}[!htbp]
    \centering
    \begin{tabular}{lr|ccc} 
    \multirow{2}{*}{\textbf{Provides URDF}} & \multirow{2}{*}{\textit{total}} &
    \multicolumn{3}{c}{\textbf{`URDF' in search}}  \\ 
    \cline{3-5}
    && \textbf{not found} & \textbf{found} & \textbf{no search bar} \\ \hline
    \textbf{yes} & 14 & 7 & 4 & 3 \\ 
    \textbf{no} & 13 & 9 & 0 & 4 \\ \hline
    \end{tabular}
    \caption{Number of \acrshortpl{oem} that provide \acrshort{urdf} Bundles versus how many \acrshortpl{oem} link to these \acrshort{urdf} Bundles directly from their website. As the table shows only 4/16 \acrshortpl{oem} provide information about \acrshort{urdf} on their websites. This table is modified from~\cite{Tola&2023b}.}
    \label{tab:manufacturer_provide_urdf_and_search}
    \end{table}

    \subsubsection{What are the Future Perspectives of URDF?}
    The majority (53.5\%) of the survey participants in~\cite{Tola&2023d} (out of 477 responses) believe that \acrshort{urdf} will be used in the future, with the following arguments:
    \begin{itemize}
        \item \acrshort{urdf} is a widely accepted and used format and many tools are built around it,
        \item it is the de-facto standard in \acrshort{ros}, and
        \item it is a good option for a format, as it allows interoperability, is simple to use, and open-source.
    \end{itemize}
    The remaining~47\% were either not sure about its future use (34.5\%) or did not believe it will be used in the future (12\%), with the following arguments:
    \begin{itemize}
        \item A new format will take over, for example, \acrshort{sdf} tackles many of the challenges that \acrshort{urdf} has,
        \item the format is too limited in terms of support for mechanisms with parallel linkages or closed-chain systems, and
        \item it has a steep learning curve.
    \end{itemize}
    These arguments provide the basis that:
    \begin{center}
        \kp{10} \textcolor{kpblue}{\acrshort{urdf}'s advantage is being open-source, interoperable, and widely used.}
    \end{center}


    \subsection{Summary and Discussion}

    This section provided an overview of \acrshort{urdf} and the main findings of the research conducted related to the format.
    The main advantages and complications of the interoperable modeling format were presented, thus addressing \textbf{RQ3}.
    It was shown that \acrshort{urdf} is widely used for modeling devices within a modular robot system, and are also related to robots in the manufacturing domain.
    The universal use of \acrshort{urdf} may be due to the level of fidelity that it represents models at, which is suitable for many applications.
    Although \acrshort{urdf} is highly interoperable compared to other formats (as shown in \cref{tab:simulators_model_support}), it has fundamental issues that must be addressed to maximize its benefits, and to be used across different phases of robotic systems integration.
    The information provided above indicates that \acrshort{urdf}'s advantage is that it is a simple method for representing a robot, but its surroundings negatively affect its interoperability and usability, due to:
    \begin{description}
        \item[Lack of standardization:] there is a clear lack of standardization within the format, where:
        \begin{itemize}
            \item \acrshort{urdf} parsers have entirely different parsing results, meaning importing \acrshortpl{urdf} into different simulators may produce different behaviors;
            \item there is no versioning of \acrshort{urdf} making it difficult for simulation tools to specify which parts of the format are supported; and
            \item inconsistencies between \acrshort{ros}-based tools and \acrshort{ros}-based documentation were found, complicating the process of both developing parsers and \acrshortpl{urdf}.
        \end{itemize}
    
        \item[Lack of documentation:] in various areas of \acrshort{urdf}:
        \begin{itemize}
            \item missing transparency on the usages of the format (even experts were confused about this);
            \item the documentation can sometimes be complicated or inadequate, making it difficult to create and use \acrshortpl{urdf}; and
            \item \acrshortpl{oem} provide insufficient amounts of information about their robots, for example, there is a general lack of kinematics and dynamics parameter values, and limited provision of 3D meshes of links that can be combined within a \acrshort{urdf}.
        \end{itemize}
    
        \item[Cumbersome workflow:] the workflow of both creating and using \acrshort{urdf} is challenging:
        \begin{itemize}
            \item there is a lack of tools for creating, debugging, validating, and previewing \acrshort{urdf} Bundle;
            \item multiple programs are needed to create a single \acrshort{urdf} Bundle, meaning the user needs expertise with each; and
            \item ensuring the accuracy of a model can be difficult.
        \end{itemize}
    \end{description}

\begin{contr}{Contribution 4 (C4):}
Identified the usage, current advantages, and challenges of \acrshort{urdf} through a user survey.
\end{contr}
            
\begin{contr}{Contribution 5 (C5):}
Created a dataset of \acrshort{urdf} files and highlighted significant discrepancies in the format and associated tools, and provided common guidelines on developing \acrshortpl{urdf}.
\end{contr}

\section{Potential Directions}

Potential improvements to the format can be made to increase:
\begin{description}
    \item[Interoperability:] one of the main purposes of \acrshort{urdf} is to have an interoperable format that can be used across             various simulation tools.
            This has resulted in its use across different organizations and tools, which have in some cases created their own extensions to \acrshort{urdf}, limiting the interoperability of the format.
            These interoperability issues can be compared with the \acrshort{uml} interoperability issues~\cite{UMLinteroperability}, where various versions and extensions are supported in different manners depending on the tools used.
            As interoperability is one of the main benefits of \acrshort{urdf}, it is essential to tackle the related challenges, by for example drawing inspiration from similar interoperable formats, such as \acrshort{uml} and the \acrfull{fmi}.
            Additionally, improving documentation and standardization would be difficult but extremely beneficial.

            To increase the use of the \acrshort{xsd}, it should be updated, and \acrshort{urdf} files should be evaluated against it.
            Tools that can help with this already exist, such as the \texttt{xsd2vdm} tool\footnote{\url{https://github.com/INTO-CPS-Association/FMI-VDM-Model}} that can convert an \acrshort{xsd} to a formal programming language called Vienna Development Method (VDM), and then convert the associated \acrshort{xml}-based file to VDM, to be evaluated against the formal representation of the \acrshort{xsd}.

    \item[Usability:] guidelines for how simulators can implement support for the format can be defined to provide consistent naming of         functions and features across simulation tools. For inspiration, usability guidelines~\cite{ISO130661} for \acrshortpl{pc}           can be studied.
            Developing a platform for creating, debugging, validating, and visualizing \acrshortpl{urdf} would make working with the format less painful.

    \item[Accuracy:] creating accurate \acrshort{urdf} models without information provided by \acrshortpl{oem} may be challenging, thus there is a need to use alternative methods or tools for this. For example, if an organization owns the physical robot to be modeled, they can use tools similar to AURT~\cite{Madsen2022} to determine the dynamics of the robot.
            Furthermore, creating an online database where \acrshort{urdf} Bundles can be shared and maintained open-source, and where developers can provide comments and improvements of the models, may result in more accurate models which are publicly available.

\end{description}

Visualizations of a simple robot system were made using \acrshortpl{urdf} in Unity, see \cref{chapter:digital_twins} for more information.
However, these visualizations have not been combined with the developed configurator, and should therefore be carried out in a potential extension of this thesis.

\chapter{Towards Digital Twins in Manufacturing} \label{chapter:digital_twins}

\backgroundsetup{
  scale=1,
  color=black,
  opacity=1,
  angle=90,
  position=current page.north,
  vshift=-250mm,
  hshift=-10mm,
  contents={
    \textcolor{gray}{\rule{5mm}{\paperheight}}
  }
}

Simulating robot systems before and after deployment can both increase the quality of the automated process and reduce integration and deployment time~\cite{Sannemann&2020}.
Recently, the concept of \acrfullpl{dt} has been gaining interest both in academia and the industry.
\acrshortpl{dt} can be used to optimize processes by updating simulation models through data traces from the physical system~\cite{Fuller20}, potentially reducing the cost of system deployment and maintenance in the long run.
This chapter introduces the concept of \acrshortpl{dt} and its associated components for unit level manufacturing processes, followed by a proposed approach for the development of modular \acrshortpl{dt} of robot systems.
This chapter is based on the research output from publications~\cite{Tola&22b} and~\cite{Bottjer2023}.

\section{Research Method}
The goal of this chapter is to address \textbf{RQ4} that is defined as: \textit{is it possible to produce \acrfullpl{dt} of modular robot systems?}

As this RQ is exploratory and applied, the goal is to gain insight into the fundamentals of \acrshortpl{dt} of robot systems, and determine how they can be developed.
To achieve this, a systematic literature review, based on~96 publications, was conducted on unit level manufacturing \acrshort{dt} applications, where generic reference models, implementation methods, key challenges, and opportunities were identified.
The review followed the deductive approach, which is beneficial when reviewing topics with an abundant number of publications~\cite{snyder2019}, see~\cite{Bottjer2023} for more information.
Furthermore, an architecture for developing modular \acrshortpl{dt} of robot systems was proposed and a prototype solution was developed.
\textbf{RQ4} is addressed by combining both fundamental knowledge derived from a literature review, and practical experience through prototyping \acrfullpl{ds}.
The results of this research can be used to guide other researchers or practitioners working in the field of \acrshortpl{dt} of robot systems to develop them.

\section{Digital Twins in Manufacturing}

\subsection{Overview}

The concept of \acrshortpl{dt} was initially introduced in space robotics in 1970, then conceptualized by Michael Grieves in 2002, and have in recent years been gaining an increasing amount of interest~\cite{NextGenerationDTRobots}.
A \acrshort{dt}, in the context of manufacturing, is defined by ISO 23247~\cite{ISO23247}, as \textit{``a fit for purpose digital representation of an observable manufacturing element with synchronization between the element and its digital representation''}.
To elaborate on this, the digital representation is most commonly used to improve the overall performance of the manufacturing process through monitoring and optimization, and is the main benefit of \acrshortpl{dt}.

\begin{figure}
        \centering
        \begin{minipage}{.63\textwidth}
          \centering
          \includegraphics[width=\textwidth]{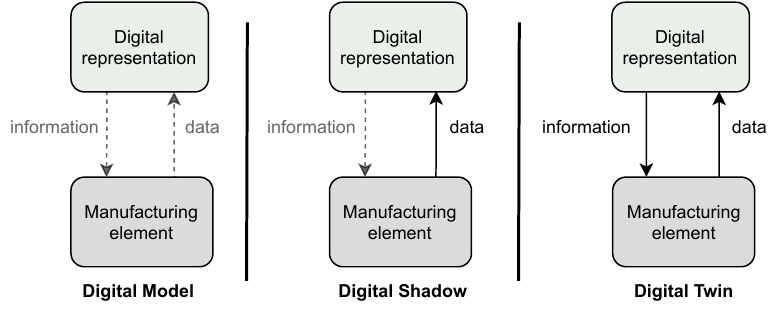}
        \captionof{figure}{Three levels of physical and digital integration related to \acrshortpl{dt}, as defined by~\cite{KRITZINGER20181016,Fuller20}. The dashed lines indicate that the flow of data/information is manual.}
        \label{fig:dt}
        \end{minipage}
        \hfill
        \begin{minipage}{.34\textwidth}
          \centering
          \includegraphics[width=\textwidth]{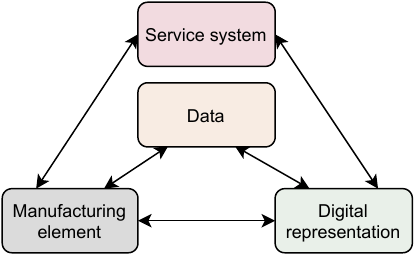}
        \captionof{figure}{Example of 5-dimensional \acrshort{dt}.}
        \label{fig:dt5d}
        \end{minipage}
        \end{figure}

There are a number of misconceptions and various understandings of what a \acrshort{dt} is.
To avoid confusion,~\cite{KRITZINGER20181016} and~\cite{Fuller20} have defined three levels of physical and digital integration, as illustrated in \cref{fig:dt}.
The simplest integration level is the \acrfull{dm} which consists of a digital representation and a manufacturing element, where data and information are manually exchanged.
This is followed by the more complex \acrshort{ds} which has an automatic data flow from the physical manufacturing element to the digital representation, and the most complex, which is the \acrshort{dt} with fully automatic data and information flow.

\backgroundsetup{
              scale=1,
              color=black,
              opacity=1,
              angle=90,
              position=current page.north,
              vshift=+250mm,
              hshift=-10mm,
              contents={
                \textcolor{gray}{\rule{5mm}{\paperheight}}
              }
            }

\subsection{Digital Twin Components of Machines at the Unit Level}
In the hierarchical levels of manufacturing, a unit process is a production unit that is capable of performing an activity to manufacture a product, and machines that can do this belong to the unit level~\cite{NAP4827}.
Robot systems are categorized as production units, thus they belong to the subset of manufacturing machines at the unit level.
This level, within manufacturing, typically represents a single machine that can carry out a specific task in the manufacturing of a product.
This could, for instance, be a \acrfull{cnc} machine for milling an object, or an additive manufacturing machine to 3D-print part of an object.
Given that modular robot systems can be used in various applications, such as assembly, palletizing, and welding, it is reasonable to investigate \acrshortpl{dt} of machines at the unit level in general, as there may be similarities in development methods and challenges. 
This section provides a brief overview of such \acrshort{dt} components, and is based on~\cite{Bottjer2023}, where more information can be found.

\subsubsection{Generic Reference Models}
A generic reference model is defined as a conceptual model that can assist in the development of a system, which is based on state-of-the-art knowledge and best practices~\cite{genereicReferenceModel}.
The two most common generic reference models of such \acrshortpl{dt} are the 3-dimensional (3D) and 5-dimensional (5D) models, with the 3D model being the most popular.
The 3D reference model for \acrshortpl{dt} was the first reference model introduced and corresponds to the \acrshort{dt} shown in \cref{fig:dt}.
The 3D reference model offers a great amount of freedom for interpretation, as it is mainly focused on conceptual aspects, which is one of the reasons for the emergence of \acrshortpl{dm} and \acrshortpl{ds}.
Thus, a 5D reference model was designed with less possibility for varying interpretations that builds on top of the 3D model with services and a central data storage, see \cref{fig:dt5d}.
Although the 3D and 5D reference models were the most widely used in our survey, other models were found, such as hierarchical and life-cycle models.
It is assumed that 3D and 5D models, due to their generality, can be applied to a wider range of applications compared to other reference models.

We found that the current challenges of such reference models are related to a lack of standardization and limited interoperability efforts, making it difficult to combine different types of services and models into a \acrshort{dt}.

\subsubsection{Services}
\backgroundsetup{contents={}}

Services, in the context of \acrshortpl{dt}, refer to functionalities or capabilities that help with the creation, management, monitoring, or interaction with the \acrshort{dt}.
Two categories of services that were commonly used in \acrshortpl{dt} are: \textit{visualization}, and \textit{monitoring} presented below.

\textit{Visualization} within \acrshortpl{dt} can be accomplished with dashboards showing graphs of sensor data, or with 3D visualizations either for interacting with the manufacturing element remotely or as a mean to quickly identify the root cause of a problem.
Dashboards can provide the health status of a machine, its energy consumption, and production efficiency, allowing experts to better understand and operate a system~\cite{Tong20201113}.
3D visualizations offer operators and technicians a more intuitive and immersive experience, for instance, during reparation.
Additionally, 3D visualizations and simulations can be used during the design of a system for determining optimal layouts, before physically integrating the hardware, to improve the process and reduce waste.
In such a case, the optimizations would be based on a \acrshort{dm}.
We found that visualizations were developed using 3D simulation engines and web-based platforms, with the former being more prevalent.
Of the 3D simulation engines, Unity was the most commonly used, and with different purposes, such as 3D visualization and workstation safety assessment.
Web-based applications were mainly used for visualizing user interfaces using JavaScript libraries, such as WebGL or BabylonJS.

\textit{Monitoring} entails tracking data and checking if the system is functioning as intended with no anomalies.
By monitoring manufacturing elements, root causes of failures can be detected, reducing their occurrence in the future and thereby ensuring product quality.
This can be employed in robot systems with safety or pharmaceutical requirements, where monitoring the system during operation ensures compliance with these regulations.
Monitoring is often used in combination with other features, such as optimization or anomaly detection.
For instance,~\cite{Redelinghuys20201383} developed an approach using \acrshortpl{dt} to detect anomalies of pneumatic grippers in robot applications, by identifying pneumatic cylinder leakages and bearing failures.

\subsubsection{Engineering Models}
Models and data are foundational in the development of \acrshortpl{dt}, and depending on the purpose of the \acrshort{dt}, specific model types can be used.
The most commonly used model types found were geometric and physical models.

\textit{Geometric models} represent the shape, appearance, and geometry of objects, which in the context of \acrshortpl{dt}, are primarily used to 3D model the manufacturing element in visualization or simulation environments.
For instance, they have also been used in Virtual Reality in interactive applications, where the user could be trained to work with the system through the visualization, or to assess safety and ergonomics in collaborative settings.

\textit{Physics-based models} represent the physical phenomena of a manufacturing element, and are often created using first-principles models or data-driven models.
Domain knowledge is required to describe the physical phenomena of a system using a first-principles approach, where equations and assumptions on the system are defined.
We found that first-principles models were commonly used in robot-aided manufacturing, where kinematic and dynamic models of the robots were created.
Kinematic models can be combined with geometric models and used for visualizations and simulations, where motion constraints can be applied.
Unlike other systems, such as \acrshort{cnc} machines, most robotic arms allow for extracting joint data directly from the robot.
This can be used for updating the \acrshort{dt}, making robot systems which are application-agnostic well suited for the development of \acrshortpl{dt}.

Data-driven models capture the physical phenomena of the model without requiring expert knowledge, instead, great amounts of data from the system are needed.
These models were mainly used for health prediction of the manufacturing element.
Their advantage is the knowledge requirements about the system are minimal, while their disadvantage is that they are specific to the application on which they have been trained on, and can therefore not be generalized to other situations outside of their training data range.

Information about the manufacturing elements' dynamics and behavior, and the ability to record data from the element, are essential for developing physics-based and data-driven models, respectively.
However, the information on manufacturing elements provided by \acrshortpl{oem} is often lacking, making it difficult for users to create accurate physics-based models.
Additionally, current software interfacing capabilities for extracting data from the elements are limited, and therefore such interfaces must be developed independently, when used for creating data-driven models.
These issues are often encountered with \acrshort{cnc} machines, and indicate that there is a need for standardizing software interfaces of manufacturing elements and facilitate a method for model exchange between the \acrshortpl{oem} and users~\cite{Werner20191743}.

Commercial software tools used for modeling and simulating \acrshortpl{dt} were: Modelica for robot motion, Tecnomatix Plant Simulation for test sequencing, and \acrshort{ros} and \acrshort{urdf} were used for creating \acrshortpl{dt} of robot systems for kinematic modeling and motion planning.

\subsubsection{Data and Communication}
Data are one of the essential components of a \acrshort{dt}, used to create data-driven models, to update the digital representation of a manufacturing element so that it reflects the physical system, and in various \acrshort{dt} services.
There are cases where physical data cannot be directly measured, and in such cases fusing information from various sensors and models can provide virtual data that is useful for monitoring and quality assessment.

The information and data exchanged between the manufacturing element and the digital representation are often carried out through communication protocols.
These can be standardized protocols, such as the OPC Unified Architecture (OPC-UA), or the Message Queuing Telemetry Transport (MQTT).
OPC-UA is a platform-independent protocol, specifically designed for industrial automation with focus on interoperability, and is often used with complex data structures.
MQTT is a messaging-protocol, more lightweight than OPC-UA, designed for low bandwidth and low power consumption applications such as Internet of Things.
Furthermore, messaging libraries can be used, such as ZeroMQ, which is lightweight, high-performing, and flexible, as it allows creating custom communication patterns, making it suitable for various \acrshort{dt} applications.
MATLAB, often used in research environments, and Python, were used for automating data-related tasks and customizing the \acrshort{dt} functionality.

\subsubsection{Closed and Open Systems}
In some cases, it can be challenging to obtain data from the manufacturing element, where additional sensors, that do not originate from the \acrshort{oem}, must be deployed onto the system to collect data.
Such systems may rely on vendor-specific software tools for extracting data or interacting with the control unit, and are called closed-systems.
One of the major benefits of \acrshortpl{dt} is automatic update of the manufacturing element.
However, this requires the ability to update control parameters of the system and send control commands, which is often very restricted for closed systems.
While vendor-specific software offers advantages like ease of use and enhanced functionality, it may pose limitations when integrating with hardware or software from other \acrshortpl{oem}, making it difficult to use in the context of \acrshortpl{dt}.

Open systems, on the other hand, are based on open communication protocols, and accessible control units and sensors.
Such systems are commonly built in laboratory environments for demonstrating \acrshortpl{dt}.
Systems can have different levels of openness, for instance, most robotic arms provide direct access to their sensor output data, such as joint positions, while updating the control parameters in a robot controller is typically not possible.
Instead, providing control commands on where the robot should move through external, custom-made protocols, is often supported, allowing \acrshortpl{dt} to redirect a robot in the case of an anomaly, but not reconfiguring the control system itself.

\subsection{Summary and Discussion}
\acrshortpl{dt} have been developed mainly at the \acrshort{ds} level due to a number of challenges, with the most problematic being the closedness of the manufacturing elements.
Closed systems are difficult to extend to \acrshortpl{dt}, and we believe one of the issues related to this are that if \acrshortpl{oem} allow users to reconfigure control parameters on their machines, there may follow warranty issues or other reputation-related issues that are difficult to handle.
Furthermore, the current architectures of the closed systems would need to be changed and this would be a costly matter.
Most of the ``\acrshortpl{dt}'' found in the literature review were uni-directional, and we believe that this may be related to the issues with closed systems, making it complicated and time-consuming to develop custom protocols, and challenging to ensure the model accuracy.

With the current level of openness of robot systems, it is possible to create simple \acrshortpl{dt} that can update the path of a robot, but not reconfigure its control parameters if they have diverged.
Comparing robot systems with \acrshort{cnc} machines, robot systems have a much higher level of openness and are therefore, with the current technologies, more suitable for the development of \acrshortpl{dt}.
Additionally, we found that \acrshortpl{dt} within the domain of robotics were commonly used with collaborative and industrial robots, and in assembly applications. 

Future directions to improve the tools and processes for creating \acrshortpl{dt} include more standardization to improve interoperability and composition of the components in a \acrshort{dt}.
Creating such tools may dramatically improve the efficiency of \acrshort{dt} development by reusing modules.

\begin{contr}{Contribution 6 (C6):}
Classified the key components of \acrshort{dt} approaches in manufacturing based on unit level processes, identified critical barriers that complicate the development of such \acrshortpl{dt}, and highlighted future opportunities.
\end{contr}

\section{Modular Digital Twins}
As described above, the literature survey we conducted on \acrshortpl{dt} characterized that robot systems are suitable for this technology.
One of the findings of the survey, was the need for a modular approach of creating \acrshortpl{dt} where modules are interoperable and can be reused.
To address this, a modular architecture for developing \acrshortpl{dt} was proposed in~\cite{Tola&22b} with a prototype implementation of two \acrshortpl{ds}.
This section presents the main findings of the publication, for more information we refer to~\cite{Tola&22b}.

\subsection{Architecture and Modules}
Developing models and services that can be reused with minor reconfigurations is possible for robot systems as they are general-purpose and composed of multiple devices.
The process of developing \acrshortpl{dt} of robot systems can be simplified by making use of reusable models and services, which allows the deployment of the \acrshort{dt} simultaneously with the physical robot system.
Creating \acrshortpl{dt} for robot systems can optimize and enhance the robot system's performance by, for instance, monitoring devices for faults or wear.
However, it is important to note that such a \acrshort{dt} only considers improving the task of the robot system itself, and does not consider optimization across other parts of the manufacturing line.

\begin{figure}
    \centering
    \includegraphics[width=\textwidth]{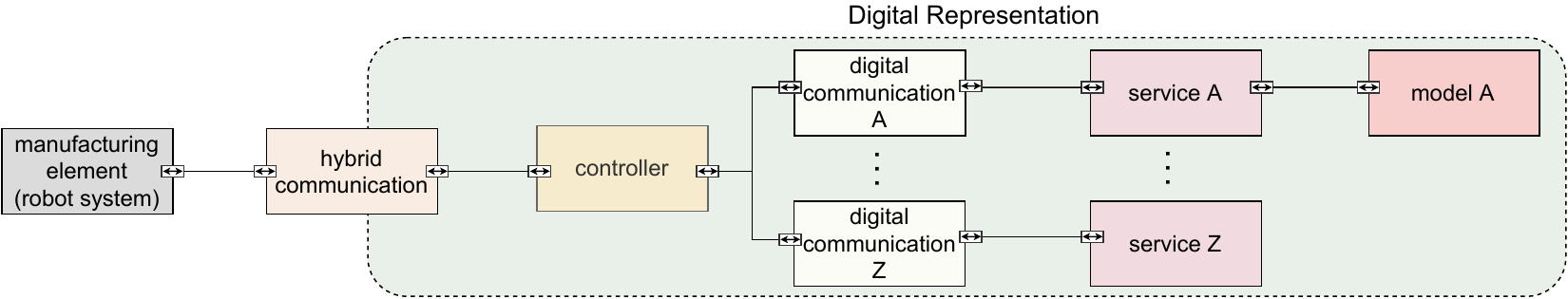}
    \caption{Proposed modular architecture for developing \acrshortpl{dt} of robot systems, redrawn with inspiration from~\cite{Tola&22b} © 2023 IEEE.}
    \label{fig:modular_dt_architecture}
\end{figure}

The proposed architecture is depicted in \cref{fig:modular_dt_architecture}, and each module is described below.
The architecture is inspired by the 5D generic reference model for \acrshortpl{dt}, see \cref{fig:dt5d}, by including service and data modules.
The core concept of this architecture revolves around developing modules with configurable parameters, fostering interoperability. 
This approach allows the same modules to be repurposed and reconfigured to suit the parameters of a new robot system. 
The modules presented in \cref{fig:modular_dt_architecture} are described below:

\begin{description}
    \item[Manufacturing element:] characterizes the physical robot system consisting of the robotic arm, attached \acrshort{eecd}, and end-effector. 

    \item[Hybrid communication:] includes both a physical cable or attachment for interfacing the digital representation with the manufacturing element and the digital communication between them. This may also represent wireless communication.

    \item[Controller:] ensures relevant data exchange between the manufacturing element and the \acrshort{dt} services, also acting as a mediator between the two.

    \item[Digital communication:] sends relevant data from the controller to subscribed services. 

    \item[Services:] provide the \acrshort{dt} improvements to the robot system, such as optimization, monitoring, and visualization. 
                    Most services require models to provide enhancements. However, some may not, such as dashboards for visualizing sensor data, which are more commonly used in \acrshortpl{ds}.

    \item[Models:] are representations of the manufacturing element used together with a service to provide \acrshort{dt} enhancements to the system. 
                
\end{description}

The architecture assumes that the controller, digital communication, services, and models all operate on the same hardware, while the hybrid communication and manufacturing element each have dedicated hardware.

\subsection{Proof-of-Concept}
\begin{figure}
    \centering
    \includegraphics[width=\textwidth]{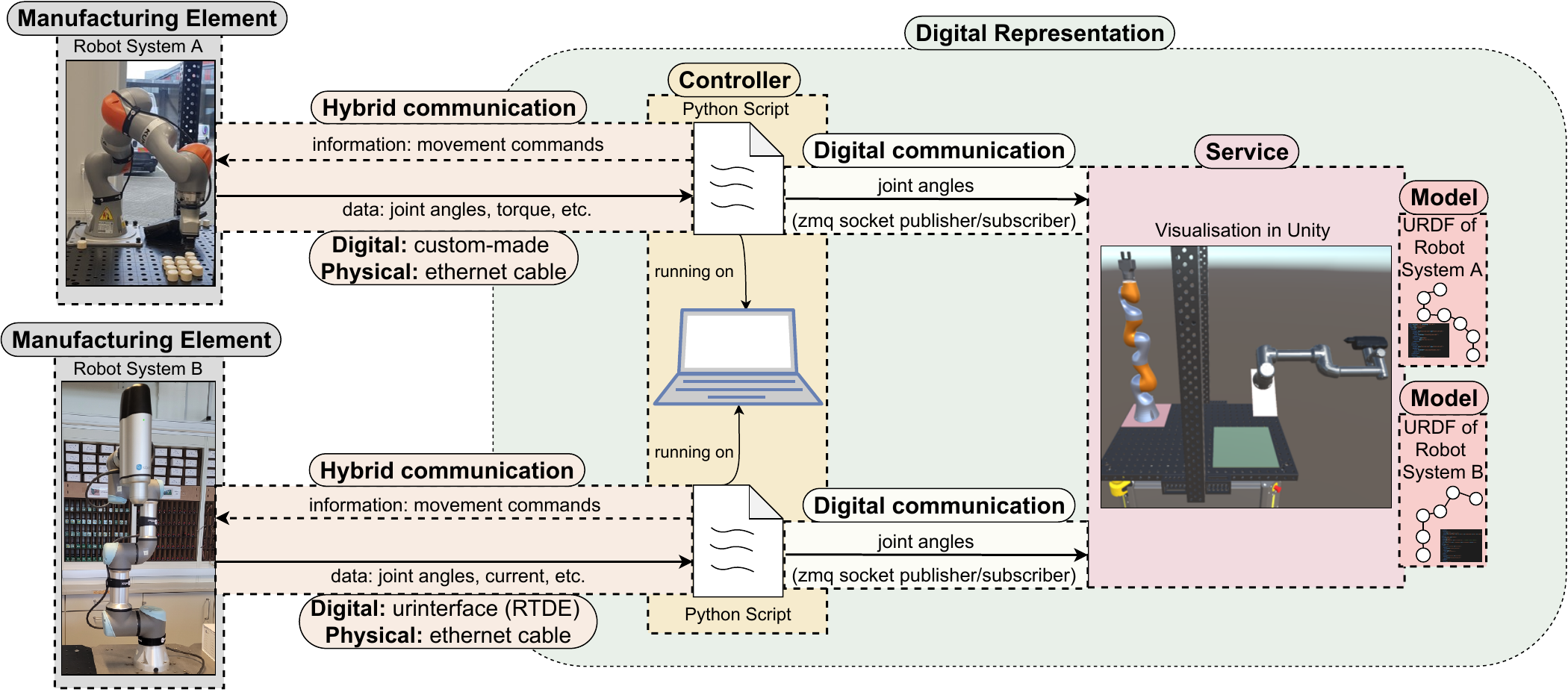}
    \caption{Proof-of-concept implementation of the modular \acrshort{dt} architecture for robot systems, modified from~\cite{Tola&22b} © 2023 IEEE. Note, that the communication is automatic from the manufacturing elements to the digital representations, but manual in the opposite direction.}
    \label{fig:modular_dt}
\end{figure}

A proof-of-concept based on a \acrshort{ds} of the proposed architecture is illustrated in \cref{fig:modular_dt}.
The specific systems and technologies employed in developing the \acrshortpl{ds} are shown. 
In this context, two different robot systems were utilized, and a single visualization service was created to integrate these two systems.
The robot systems consisted of robotic arms from two different manufacturers, and two different types of end-effectors: a gripper and a screwdriver.
The data measured from the manufacturing element were the joint angles of the robotic arms, based on built-in sensors within the robotic arms.
These data were used in the visualization service, illustrating the movement of the robotic arms in the 3D game-engine Unity, simultaneously with the movement of the physical arms\footnote{Similar to another \acrshort{ds} we developed, shown in \url{https://www.youtube.com/watch?v=1vt7-qrFvZc}.}.

The hybrid communication between the robot systems and the controller was developed separately because each robotic arm supports different communication protocols for direct manipulation.
This challenge is related to the lack of standardization discussed in challenge \textit{CH~2B} in \cref{chapter:integration:challenges}.
Consequently, a custom communication solution was created for one of the robot systems, while an open-source library was utilized for the other.
Note, that it is not possible with such robotic arms to configure the control loop and parameters, due to their controllers being closed.
Therefore, developing a \acrshort{dt} of such a system could involve creating an external control loop which can only command the position of the robot, but not adapt its dynamics.

The controllers of the \acrshortpl{ds} were implemented on a single \acrshort{pc} using Python to perform the required tasks.
A controller was created for each robot system, with the main tasks executed in the following order, where tasks two and three are continuously run during operation:
\begin{enumerate}
    \item Sends predefined movement commands to the robot systems to initiate the process.
    \item Receives measurement data from the robot.
    \item Publishes relevant measurement data through a port using the ZeroMQ library\footnote{\url{https://zeromq.org/languages/python/}}.
\end{enumerate}

The digital communication was developed using ZeroMQ sockets for publishing and subscribing to data.
The sockets were implemented in Python for the controller, and C\# for the services, where Unity was used.
The Unity application was set up to subscribe to the published joint angles of each robotic arm.
Changes in the robotic arms' positions were visualized using the data.
\acrshort{urdf} was used to model the robots, by representing their kinematics and 3D geometries.
The complete robot system was composed using \acrshort{urdf} models of the robotic arms and end-effectors, and \acrshort{cad} models of the \acrshortpl{eecd}.

\subsection{Summary and Discussion}

The goal of this proof-of-concept is to demonstrate the creation of \acrshortpl{ds} for modular robot systems, with a primary focus on visualization services. 
In the future, this concept has the potential to be expanded to include visualization of virtual data or other parameters, such as joint currents, or simulating collision-free trajectories. 
The current implementation required some initial implementation efforts, but by developing modules in a modular and reconfigurable manner, they can potentially be reused for \acrshortpl{dt} in other robot systems.

One of the current challenges in \acrshort{dt} development is the closed nature of the controllers for robotic arms, limiting control to just the trajectory of the robot. 
We believe \acrshort{dt} developments could greatly benefit from device \acrshortpl{oem} providing plug-and-play hybrid communication modules, and interoperable models, such as \acrshort{urdf}, for their devices. 
Safety concerns in control situations and synchronization between the digital representation and the manufacturing element have not been addressed in this architecture.

This chapter addresses \textbf{RQ4} by demonstrating how a \acrshort{ds} \textit{can} be created for modular robot systems, and presenting the challenges of developing \acrshortpl{dt} of robot systems with full command of the control parameters.

\begin{contr}{Contribution 7 (C7):}
Demonstrated an approach for creating modular \acrshortpl{ds} of robot systems.
\end{contr}

\section{Potential Directions}
In general, there are many advances to be made to ease the process of developing \acrshortpl{dt} of modular robot systems.
Potential future work in this area could include the following:

\begin{description}
    \item[Modeling and simulation:] Concepts such as co-simulation and the \acrfull{fmi} can be explored for coupling models of different aspects/parts of a \acrshort{dt} together. 
    Black-box models of devices can be created using \acrshort{fmi}, referred to as Functional Mock-Up Units (FMUs).
    If device \acrshortpl{oem} provide FMUs of their controllers and physical systems, there is significant potential for reusing these models and integrating them into co-simulations, reducing modeling time and effort.
    We have worked on tools for easily generating FMUs, using well-known languages like Python and C\#~\cite{Legaard2021}, and tools for visualizing FMUs in applications such as Unity\footnote{\url{https://github.com/INTO-CPS-Association/unifmu_examples/tree/master/examples/UnityFMUTemplate}}.
    
    Furthermore, achieving model accuracy requires calibrating model parameters.
    Tools like AURT, that we have developed, can be utilized to calibrate the dynamic parameters of a robotic arm through simple trajectory experiments and optimization~\cite{Madsen2022}.

    \item[DT platforms:] The development of \acrshort{dt} platforms should aim to consolidate all the necessary software for \acrshort{dt} creation into a single, user-friendly environment, facilitating ease of use. 
    Addressing synchronization challenges among different systems and providing guides on how to tackle these issues can enhance the usability of such platforms.

\end{description}

\chapter{Concluding Remarks} \label{chapter:conclusion}

\backgroundsetup{
  scale=1,
  color=black,
  opacity=1,
  angle=90,
  position=current page.north,
  vshift=-250mm,
  hshift=-10mm,
  contents={
    \textcolor{gray}{\rule{5mm}{\paperheight}}
  }
}

\noindent This chapter evaluates the research conducted in this thesis, by assessing the contributions, and through assessment of the methods used to address the Research Questions.
Afterwards, concluding remarks are provided, and potential future directions are outlined.

\section{Assessment of Contributions} \label{sec:assessment_contributions}
The contributions and criteria for the self-assessment of them are presented in this section.

\subsection{Research Contributions}
The research contributions are initially presented in \cref{sec:intro_contributions}.
\begin{itemize}
    \item[\textbf{C1:}] Characterized the main stakeholders involved in robotic systems integration and detailed the process.

    \item[\textbf{C2:}] Outlined current challenges in acquiring and integrating modular robot systems and presented directions to tackle them.

    \item[\textbf{C3:}] Proposed a formal approach for constraint-based configuration of robot systems to address a subset of the identified challenges. 

    \item[\textbf{C4:}] Identified the usage, current advantages, and challenges of \acrshort{urdf} through a user survey.

    \item[\textbf{C5:}] Created a dataset of \acrshort{urdf} files and highlighted significant discrepancies in the format and associated tools, and provided common guidelines on developing \acrshortpl{urdf}.

    \item[\textbf{C6:}] Classified the key components of \acrshort{dt} approaches in manufacturing based on unit level processes, identified critical barriers that complicate the development of such \acrshortpl{dt}, and highlighted future opportunities.

    \item[\textbf{C7:}] Demonstrated an approach for creating modular \acrshortpl{ds} of robot systems.
\end{itemize}

\subsection{Assessment Criteria}
The assessment criteria and the reasoning behind them are presented in \cref{sec:intro_assessment}.
The criteria are defined based on the main hypothesis of the project to aid in the assessment of proving or disproving the hypothesis.
Each ranking criterion is evaluated on a scale from~1-4, where~1 is the lowest and~4 is the highest ranking.
The guidelines for ranking the \textit{acquisition}, \textit{integration}, and \textit{deployment} of robot systems are all described in the same paragraph as they can be ranked in a similar manner.
They are described using \textit{[X]}, where \textit{X} indicates each specific ranking criterion.

\backgroundsetup{
              scale=1,
              color=black,
              opacity=1,
              angle=90,
              position=current page.north,
              vshift=+250mm,
              hshift=-10mm,
              contents={
                \textcolor{gray}{\rule{5mm}{\paperheight}}
              }
            }

\begin{description}
    \item[\textit{Advancement of [Acquiring, Integrating, Deploying] Robot Systems:}] The process of [acquiring, integrating, deploying] robot systems can be advanced in various dimensions, leading to easier and faster [acquisition, integration, deployment] times.
    These dimensions can be anything from discovering fundamental knowledge about the process to the development of tools for assisting in parts of the process. Furthermore, advancing this area across different types of robot applications is considered.
    The following guidelines are used for evaluation:
        \begin{description}
            \item[Level 1] The contribution does not provide any significant advancement in [acquiring, integrating, deploying] robot systems.

            \item[Level 2] The contribution offers limited value, enabling slight improvements in the [acquisition, integration, deployment] of robot systems.

            \item[Level 3] The contribution significantly advances the [acquisition, integration, deployment] of robot systems either through methods or knowledge to build upon, and can be transferred to various applications.

            \item[Level 4] The contribution is disruptive in the [acquisition, integration, deployment] of robot systems, making the process less labor-intensive, and is applicable to various robot applications.
        \end{description}

    \item[\textit{Reducing Vendor-Lock-in:}] Vendor-specific tools exist for visualization and simulation of robot systems, making it difficult for users to migrate all their models to other tools when needed. Furthermore, it is common that stakeholders become dependent on vendor-specific robot devices when integrating robot systems due to partnerships with \acrshortpl{oem}. This criterion ranks a contribution by assessing its reduction of vendor lock-in within any of these two aspects, using the following guidelines:
        \begin{description}
            \item[Level 1] The contribution offers limited solutions to address vendor lock-in and may not effectively mitigate this issue.

            \item[Level 2] The contribution provides partial solutions to reduce vendor lock-in but may not fully eliminate the problem.

            \item[Level 3] The contribution significantly reduces vendor lock-in, offering practical solutions for the industry to build upon.

            \item[Level 4] The contribution effectively eliminates vendor lock-in, ensuring maximum flexibility and independence for industry stakeholders.
        \end{description}
\end{description}

\subsection{Assessment According to Ranking Criteria}

\begin{figure}[htb]
	\centering
	\usetikzlibrary{shapes}
\usetikzlibrary{positioning,fit,calc}

\ifdefined\cref\else\def\cref#1{[C1]}\fi
\newcommand{\D}{7} 
\newcommand{\U}{4} 

\newdimen\R 
\R=3.5cm
\newdimen\L 
\L=4.3cm

\newcommand{\A}{360/\D} 

\def\makeweb{
    \tikzset{spiderwebpath/.style={opacity=0.4}}
    \tikzset{measurepath/.style={ultra thick,opacity=0.8,rounded corners=1pt}}
    \tikzset{measurepath2/.style={ultra thick,opacity=1,rounded corners=1pt}}
    \tikzset{measurepath3/.style={ultra thick,opacity=0.6,rounded corners=1pt}}
    \tikzset{mcol1/.style={color={rgb:red,197;green,90;blue,17}}}
    \tikzset{mcol2/.style={color={rgb:red,191;green,144;blue,0}}}
    \tikzset{mcol3/.style={color={rgb:red,100;green,200;blue,50}}}
    \tikzset{mcol4/.style={color={rgb:red,102;green,44;blue,145}}}
    \tikzset{mcol5/.style={color=olive}}
    \tikzset{mcol6/.style={color=darkgray}}

    \path (0:0cm) coordinate (O); 
    
    \foreach \X in {1,...,\D}{
    \draw[spiderwebpath] (\X*\A:0) -- (\X*\A:\R);
    }
    
    \foreach \Y in {0,...,\U}{
    \foreach \X in {1,...,\D}{
      \path (\X*\A:\Y*\R/\U) coordinate (D\X-\Y);
      \fill[spiderwebpath] (D\X-\Y) circle (4pt);
    }
    
    \draw [spiderwebpath] (0:\Y*\R/\U) \foreach \X in {1,...,\D}{
        -- (\X*\A:\Y*\R/\U)
    } -- cycle;
    }
       \fill[white] (0,0) circle (4.03pt);
     \fill[spiderwebpath] (0,0) circle (4pt);
    
    \path (1*\A:\L) node (L1) {\labelformat{C1}};
    \path (2*\A:\L) node (L2) {\labelformat{C2}};
    \path (3*\A:\L) node (L3) {\labelformat{C3}};
    \path (4*\A:\L) node (L4) {\labelformat{C4}};
    \path (5*\A:\L) node (L5) {\labelformat{C5}};
    \path (6*\A:\L) node (L6) {\labelformat{C6}};
    \path (7*\A:\L) node (L7) {\labelformat{C7}};	
}

\centering

\def\labelformat#1{\large #1}
\def\subfixspiderscale{0.5}

\def\makeContributionCompareWeb{

\begin{subfigure}[t]{0.3\textwidth}
\centering
\begin{tikzpicture}[scale=\subfixspiderscale, transform shape]
\makeweb
      \fill [mcol1,measurepath]
    (D1-2) -- 
    (D2-3) --
    (D3-4) --
    (D4-2) --
    (D5-2) --
    (D6-1) --
    (D7-1) --
    cycle;
    \fill [mcol6,measurepath2]
     (D1-1) -- 
     (D2-1) -- 
     (D3-1) -- 
     (D4-1) -- 
     (D5-1) -- 
     (D6-1) -- 
     (D7-1) -- 
    cycle;

\end{tikzpicture}
\caption{\scriptsize Advancement of Acquiring Robot Systems}
\label{fig:acquiring}
\end{subfigure}\quad%
\begin{subfigure}[t]{0.3\textwidth}
\centering
\begin{tikzpicture}[scale=\subfixspiderscale, transform shape]
\makeweb
  \fill [mcol2,measurepath]
    (D1-2) -- 
    (D2-2) --
    (D3-2) --
    (D4-3) --
    (D5-3) --
    (D6-1) --
    (D7-1) --
    cycle;
    \fill [mcol6,measurepath2]
     (D1-1) -- 
     (D2-1) -- 
     (D3-1) -- 
     (D4-1) -- 
     (D5-1) -- 
     (D6-1) -- 
     (D7-1) -- 
    cycle;

\end{tikzpicture}
\caption{\scriptsize Advancement of Integrating Robot Systems}
\label{fig:ctr_integrating}
\end{subfigure}\quad%
\begin{subfigure}[t]{0.3\textwidth}
\centering
\begin{tikzpicture}[scale=\subfixspiderscale, transform shape]
\makeweb
  \fill [mcol3,measurepath]
    (D1-2) -- 
    (D2-1) --
    (D3-1) --
    (D4-2) --
    (D5-2) --
    (D6-3) --
    (D7-3) --
    cycle;
    \fill [mcol6,measurepath2]
     (D1-1) -- 
     (D2-1) -- 
     (D3-1) -- 
     (D4-1) -- 
     (D5-1) -- 
     (D6-1) -- 
     (D7-1) -- 
    cycle;

\end{tikzpicture}
\caption{\scriptsize Advancement of Deploying Robot Systems}
\label{fig:deploying}
\end{subfigure}\quad\\%
\begin{subfigure}[t]{0.3\textwidth}
\centering
\begin{tikzpicture}[scale=\subfixspiderscale, transform shape]
\makeweb
  \fill [mcol4,measurepath]
    (D1-2) -- 
    (D2-2) --
    (D3-3) --
    (D4-3) --
    (D5-3) --
    (D6-1) --
    (D7-2) --
    cycle;
    \fill [mcol6,measurepath2]
     (D1-1) -- 
     (D2-1) -- 
     (D3-1) -- 
     (D4-1) -- 
     (D5-1) -- 
     (D6-1) -- 
     (D7-1) -- 
    cycle;

\end{tikzpicture}
\caption{\scriptsize Reducing Vendor Lock-in}
\label{fig:vendor_lockin}
\end{subfigure}\quad%
\begin{subfigure}[t]{0.3\textwidth}
\centering
\begin{tikzpicture}[scale=\subfixspiderscale, transform shape]
\makeweb
    \fill [mcol1,measurepath3]
    (D1-2) -- 
    (D2-3) --
    (D3-4) --
    (D4-2) --
    (D5-2) --
    (D6-1) --
    (D7-1) --
    cycle;
    \fill [mcol2,measurepath3]
    (D1-2) -- 
    (D2-2) --
    (D3-2) --
    (D4-3) --
    (D5-3) --
    (D6-1) --
    (D7-1) --
    cycle;
    \fill [mcol3,measurepath3]
    (D1-2) -- 
    (D2-1) --
    (D3-1) --
    (D4-2) --
    (D5-2) --
    (D6-3) --
    (D7-3) --
    cycle;
    \fill [mcol4,measurepath3]
    (D1-2) -- 
    (D2-2) --
    (D3-3) --
    (D4-3) --
    (D5-3) --
    (D6-1) --
    (D7-2) --
    cycle;
    \fill [mcol6,measurepath2]
    (D1-1) -- 
    (D2-1) -- 
    (D3-1) -- 
    (D4-1) -- 
    (D5-1) -- 
    (D6-1) -- 
    (D7-1) -- 
    cycle;

\end{tikzpicture}
\caption{\scriptsize Combined Overview}
\label{fig:combined_overview}
\end{subfigure}%
}


	\makeContributionCompareWeb
	\caption{Evaluation of contributions based on ranking criteria.}
	\label{fig:contributions_ranking}
\end{figure}
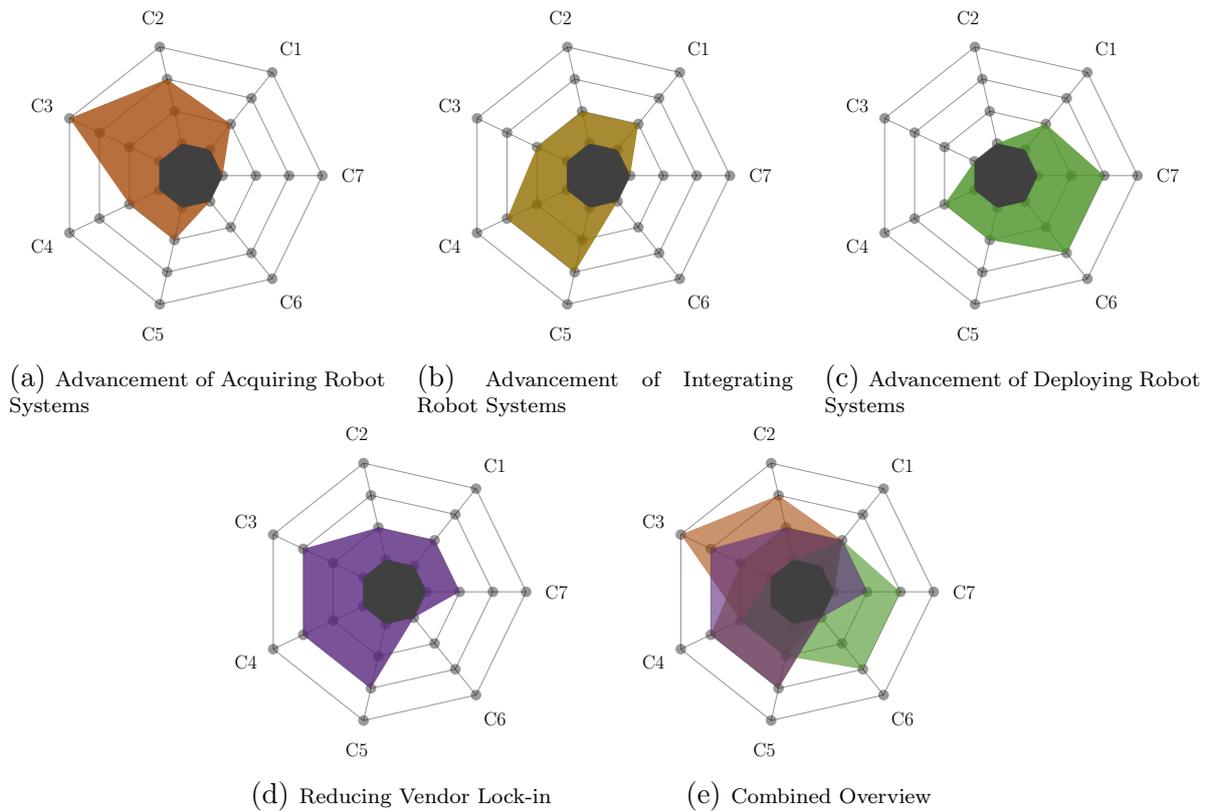

\Cref{fig:contributions_ranking} shows the ranking of the contributions based on the different ranking criteria introduced above, as judged by the author, with \cref{fig:combined_overview} showing a combined overview.
The reason behind the rankings of each contribution are:
\begin{description}
    \item[\textit{Advancement of Acquiring Robot Systems:}] this criterion focuses on the process of acquiring robot systems, see \cref{fig:acquiring}.
    Contributions that are not related to acquisition are ranked as \textbf{level one}, these are \textbf{C6} and \textbf{C7}.
    The contributions related to \acrshort{urdf} can be used in the context of visualization and simulation of the system during the acquisition process, and therefore \textbf{C4} and \textbf{C5} are ranked at \textbf{level two}.
    Furthermore, \textbf{C1} provides fundamental knowledge on the complete process of acquisition, integration, and deployment of robot systems, together with the stakeholders involved. 
    Thus, this fundamental knowledge can be used to build upon when developing tools or analyzing areas for improvement, and therefore this contribution is ranked at \textbf{level two}.
    \textbf{C2} identifies critical challenges related to configuring robot systems, and lays the foundation for \textbf{C3}.
    Therefore, \textbf{C2} is ranked at \textbf{level three}.
    \textbf{C3} provides a novel proof-of-concept configurator that tackles a number of configuration challenges and issues that have not been solved before, and can reduce the need for experts to review a configuration before sales.
    Therefore, it is ranked at \textbf{level four}.

    \item[\textit{Advancement of Integrating Robot Systems:}]
    this criterion focuses on the process of integrating robot systems, see \cref{fig:ctr_integrating}.
    \textbf{C1} advances integration by providing information about the stakeholders involved in the process, and \textbf{C2} identifies barriers related to the process.
    Therefore, these two contributions are ranked at \textbf{level two}.
    \textbf{C3} can be used to validate the compatibility of devices within a robot system to ensure valid simulations, and is ranked at \textbf{level two}.
    \textbf{C4} and \textbf{C5} both provide new knowledge about \acrshort{urdf}, which can be used in the visualization or simulation of a robot system.
    \acrshort{urdf}'s capabilities and drawbacks are identified, showing the limitations of using the format in the context of robot system integration.
    These two contributions are ranked at \textbf{level three}.
    \textbf{C6} and \textbf{C7} only provide contributions to the deployment of robot systems, and are therefore ranked at \textbf{level one}.
    
\backgroundsetup{contents={}}

    \item[\textit{Advancement of Deploying Robot Systems:}] this criterion focuses on the process of deploying robot systems, see \cref{fig:deploying}.
    \textbf{C4} and \textbf{C5} are related to advancing the interoperable robot modeling format \acrshort{urdf}, which can be used in the context of \acrshortpl{dt}, and are therefore ranked at \textbf{level two}.
    \textbf{C1} provides fundamental knowledge on the process of deploying robots and the stakeholders involved, which can be built upon, and is therefore ranked at \textbf{level two}.
    \textbf{C2} outlines the challenges related to acquisition and integration, and is not directly related to deployment, therefore ranked at \textbf{level one}.
    \textbf{C3} can potentially be used in online reconfiguration, if the objects that a robot system manipulates vary in size, shape, and texture, requiring the system to adapt its end-effector.
    This has not been tested in this thesis and would require some future work, therefore ranking this contribution at \textbf{level one}.
    \textbf{C6} provides fundamental information on \acrshortpl{dt} of unit level machines, which can be applied to robot systems for various applications.
    \acrshortpl{dt} can be used to optimize processes, and as this information is broad on concepts, applications, and technologies that can be used for developing \acrshortpl{dt}, this contribution is ranked at \textbf{level three}.
    \textbf{C7} illustrates an approach to develop modular \acrshortpl{ds} of robot systems, where two robot systems were used in the proof-of-concept. 
    Thus, this contribution is ranked at \textbf{level three}.

    \item[\textit{Reducing Vendor-Lock-in:}] this criterion focuses on how the contributions help reduce the negative effects of vendor lock-in, see \cref{fig:vendor_lockin}.
    \textbf{C1} and \textbf{C2} both provide fundamental knowledge on robotic systems integration through identifying stakeholder relations and challenges in this process. 
    This knowledge can be built upon by analyzing the contributed information and determining how vendor lock-in can be reduced.
    Therefore, these contributions are ranked at \textbf{level two}.
    \textbf{C3} allows integrators to reduce their reliability on the knowledge gained through years of experience working with a limited number of \acrshortpl{oem}.
    Furthermore, it allows easier configuration of a greater number of devices from multiple \acrshortpl{oem}, and therefore it is ranked at \textbf{level three}.
    \textbf{C4} and \textbf{C5} provide information on how the interoperable modeling format, \acrshort{urdf}, is used, its current limitations, and advantages. 
    This information can be built upon to further improve \acrshort{urdf}, and thus reduce the effects of vendor lock-in.
    Therefore, these contributions are ranked at \textbf{level three}.
    \textbf{C6} does not provide any beneficial information for reducing vendor lock-in, and is therefore ranked at \textbf{level one}.
    \textbf{C7} shows a modular method to create \acrshortpl{dt} of robot systems, using \acrshort{urdf}, Python, and Unity, which together reduce the dependency on vendor-specific simulation tools, and is therefore ranked at \textbf{level two}.
    
\end{description}

\subsection{Limitations of the Contributions}
\backgroundsetup{contents={}}

\textbf{C1} and \textbf{C2} both provide contributions across the assessment criteria.
However, they only consider a limited number of applications, pick-and-place and screwdriving, and therefore their overall output can not necessarily be generalized to all applications.
Furthermore, we have not considered the percentage of the different types of applications used in industry, compared with the applications considered in this thesis.
Therefore, to determine the impact of this contribution, it is necessary to provide specific details on this.

\textbf{C3} demonstrated a method to create a configurator that addresses the challenges of lacking and misleading documentation.
Although its contribution is ranked at the highest level, it can be improved to consider more applications and more devices.

\textbf{C4} and \textbf{C5} both present novel information about \acrshort{urdf} that can be used to build upon in the development of tools for improving the format.
However, taking this format into the context of integration of robot systems, it is important to define more detailed requirements for the capabilities and limitations of such an interoperable format, and outline the current drawbacks of \acrshort{urdf} with regards to the requirements.

\textbf{C6} provides fundamental knowledge on \acrshortpl{dt} of machines at the unit level.
As this knowledge is general for all machines at the unit level, an improvement to this contribution would be creating a comprehensive analysis with focus on \acrshortpl{dt} for modular robot systems.
\textbf{C7} demonstrates an approach for developing \acrshortpl{ds} of modular robot systems.
However, it only focuses on one service, which is visualization, and does not describe in detail how the \acrshortpl{ds} can be extended to \acrshortpl{dt}.

\section{Assessment of Research Questions}
\begin{figure}
    \centering
    \includegraphics[width=0.95\textwidth]{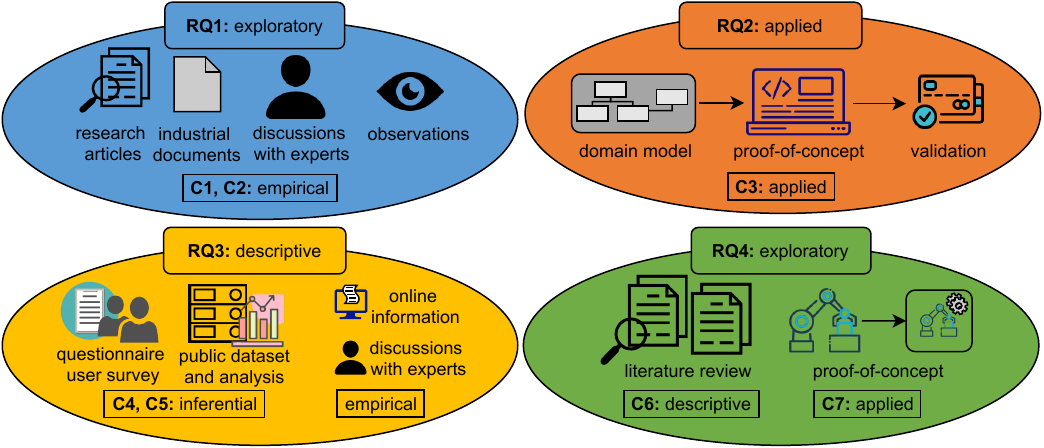}
    \caption{Overview of Research Questions, their types, the information and methods used to answer these questions, together with their related contributions.}
    \label{fig:research_objectives}
\end{figure}

In this section the addressing of the Research Questions (RQs) is assessed, where a preferable method to address these questions is described for each, followed by how the questions were addressed in this thesis and what their limitations are.
\Cref{fig:research_objectives} shows an overview of the RQ types, the methods used to address them, the contributions related to each RQ, and the research method the contribution is based on.
Before delving into the specific RQs, a brief description of the benefits and limitations of each type of RQ is presented.

\subsection{Advantages and Limitations of Research Question Types}

\textit{Exploratory research} is used to gain new insights into an area through exploration.
Thus, it does not provide final conclusions on how to solve a specific problem.
It is typically used when addressing new issues where research on the topic has been limited.
The main advantages of using \textit{exploratory} research for addressing \textbf{RQ1} and \textbf{RQ4} are that it can help provide a foundation that future research can build upon. 
This can be done by generating new ideas and hypotheses, and by testing the feasibility of new research ideas.
The limitations of exploratory research are that it is often difficult to generalize as the research may be based on a limited number of samples, and the results may be subjective~\cite{ExploratoryResearch}.

\textit{Applied research} is used for finding a solution to a specific problem.
In the case of \textbf{RQ2}, it is used as a follow up on the results of researching \textbf{RQ1} to address the problems related to configuration.
The main advantage of using \textit{applied} research for addressing \textbf{RQ2} is that it demonstrates a method for solving the problem of configuration.
The drawbacks of this type of research are that it cannot be generalized to other domains, i.e., it is limited to the specific problem at hand~\cite{AppliedResearch}.

\textit{Descriptive research} is used to explain the current state of a research area by presenting its characteristics.
It is typically used to highlight problems through systematic data collection, which is suitable for \textbf{RQ3}.
Furthermore, this research method often leads to new hypotheses that can be explored.
The advantages of using \textit{descriptive} research for addressing \textbf{RQ3} are that it can provide a detailed snapshot of the topic at hand, and can be used to identify trends and patterns.
The limitations of this type of research are the findings may not be generalizable depending on the data collected, and the cause and effect relationships are not established~\cite{DescriptiveResearch}.

\subsection{Assessment of Methods Used to Address Research Questions}
This section presents the RQs, followed by the preferable method for addressing them, the actual methods used to address them, and an assessment of these.
The assessment is judged by the author, and therefore there is a risk of subjective bias that the reader should be aware of.
The fulfillment of the Research Questions is ranked between levels~1-4, where~1 indicates a very low level of fulfillment, and~4 indicates everything is addressed in a proper and detailed manner.

\subsubsection{RQ1: \textit{What are the main barriers of acquiring and integrating modular robot systems, and how can they be addressed?}}
    \begin{underlinedescription}
        \item[Preferable method to address RQ:] 
                This RQ should be addressed using experts from industry and producing proof-of-concepts.
                Industrial experts from robotic system integrators should be involved to verify the identified challenges and rate the complexity level of each challenge.
                It is desirable that experts come from different robotic system integrators, and with expertise in different applications to determine if the barriers discovered are generalizable to various manufacturing applications.
                To determine the applicability of the proposed methods, prototypes could be created to prove if the methods properly address the challenges found.
                Additionally, in some cases the methods to address the issues may require adapting or creating standards for the process of robotic systems integration. 
                In such a case, assessing such proposed methods should be performed through systematic interviews with industrial experts and standard committee members.
                Furthermore, achieving applied experience with this RQ would be beneficial, but as the process of robotic systems integration is lengthy, it may be difficult, unless a smaller project is available and an integrator is willing to include the researcher in the process.

        \item[Applied method to address RQ:] 
        This RQ is addressed through inductive methods, where information from specific cases has been used to create generalizations.
        The information is gathered, based on:
                \begin{itemize}
                    \item various research publications;
                    \item industrial documents, such as data sheets from \acrshortpl{oem}, brochures, user manuals, or standards;
                    \item discussions with two Danish robotic system integrators;
                    \item discussions with two research organizations that work with advancing robots for manufacturing, one Danish and one Australian; and
                    \item general observations made using knowledge gained through experience.
                \end{itemize}

        \item[Assessment:] 
            This RQ is well addressed by gathering knowledge through different, suitable methods by discussing with relevant stakeholders and providing both research-based and industry-based knowledge.
            However, the knowledge is limited to only a few stakeholders and documents, and therefore the generalizability of the found information is uncertain.
            Furthermore, a lack of feedback and iterating over the defined challenges with integrators may suggest that not all challenges have been identified.
            The addressing of this RQ is ranked at level~3.
    \end{underlinedescription}

\subsubsection{RQ2:  \textit{Is it possible to define a configurator of modular robot systems that takes application requirements and compatibility of devices into account?}}

    \begin{underlinedescription}
        \item[Preferable method to address RQ:] 
                This RQ should be addressed through creating and testing a proof-of-concept on various datasets of products and applications.
                The functionality of the proof-of-concept should be validated by testing different applications and requirements.
                The configurator should be extensible so that if new challenges are discovered, it can be extended to tackle these challenges in a proper manner.
                This can be tested by adding new functionalities, for example, different types of requirements that can improve the configuration for a given application.
                Furthermore, the applicability and limitations of the configurator across manufacturing domains should be determined.

        \item[Applied method to address RQ:] 
                This RQ is addressed through deductive methods, where a hypothesis that a configurator can be made to solve the identified challenges is posed, and a proof-of-concept based on a domain model is developed.
                The proof-of-concept was tested on a limited number of devices,~20, and from~3 different \acrshortpl{oem}, which all played a role in the challenges found.                
                The functionality of the proof-of-concept was verified on two different applications.

        \item[Assessment:]
                This RQ is well addressed, as a domain model was developed that tackles the identified compatibility challenges and a proof-of-concept was created to illustrate the functionality of the configurator.
                The weakness of this assessment is the minimal testing of the configurator, which does not capture its limitations and applicability to various manufacturing domains.
                Furthermore, the scalability and extensibility of the method is not analyzed, and therefore if new challenges are discovered, the complexity of extending the domain model to these is uncertain.
                Moreover, including experts in the validation of the configurator would have been beneficial, as potential issues or improvements could have been identified.
                The addressing of this RQ is ranked at level~3.
                
    \end{underlinedescription}
  
\subsubsection{RQ3: \textit{What are the advantages and complications of defining an interoperable robot modeling format for visualization and simulation of robots?}}

    \begin{underlinedescription}
        \item[Preferable method to address RQ:]
        This RQ should be addressed by investigating existing formats that can be used for visualizing and simulating robots, and comparing their capabilities and limitations, how widespread their use is, and how relevant they are for modeling robots across different parts of the robotic systems integration process.
        Defining requirements for the capabilities of the format to be used across the different phases of robotic systems integration is essential, as these requirements can be used to choose a format.
        Once a format is chosen, it should be intensely investigated, by identifying who uses it, how it is used, what its advantages and current limitations are, and how can it be improved.
        This can be identified through different quantitative and qualitative methods, such as surveys, interviews, and data analysis.
        Furthermore, identifying how such a format can be used across different phases of robotic systems integration is beneficial.

        \item[Applied method to address RQ:] 
        This RQ is addressed by investigating four different formats for modeling robots, where their main usage is compared, along with which simulation tools they are supported by.
        The \acrshort{urdf} format was chosen, as it was the most suitable for modeling robot devices at a simple level that can be used across different phases of robotic systems integration.
        The choice of \acrshort{urdf} was based on a limited number of requirements, and restrictions of this PhD project that Unity must be used.
        It was further investigated through a user survey, creating a dataset and analyzing it, and using empirical data by discussing with experts and from public information.

        \item[Assessment:] 
        This RQ is well addressed, as an interoperable modeling format is found through brief comparisons, and investigated using quantitative and empirical methods.
        The limitation of this approach is that the comparison of the current formats was at a high abstraction level with limited requirements on the format's capabilities, and therefore not sufficiently detailed.
        Requirements should be defined based on the most commonly used robotic devices in a robot systems, where the format's capabilities are analyzed with regards to these.
        Furthermore, mapping the use of this format to identify which simulation tools and in which phases of robotic systems integration it is most suitable to use was not performed.
        The addressing of this RQ is ranked at level~3.
    \end{underlinedescription}

\subsubsection{RQ4: \textit{Is it possible to produce \acrfullpl{dt} of modular robot systems?}}
    \begin{underlinedescription}
        \item[Preferable method to address RQ:]
        As this RQ is a combination of exploratory and applied research, there are multiple techniques that can be used in addressing it.
        Examples could be gathering new information through interviews with \acrshort{dt} experts and robotic system integrators on how they believe \acrshortpl{dt} can be used and produced for modular robot systems.
        Literature reviews on the area of \acrshortpl{dt}, specifically related to robot systems, can be conducted to understand the state-of-the-art in this area.
        Performing a feasibility study on developing a proof-of-concept of modular robot systems should be carried out, where the capabilities and limitations of such a \acrshort{dt} are investigated.
    
        \item[Applied method to address RQ:] 
        This RQ was addressed by conducting a comprehensive literature review on machines at the unit level as robot systems are application-agnostic and can be used in various applications.
        A proof-of-concept based on a case study of two \acrshortpl{ds} was developed, illustrating the feasibility of developing a \acrshort{ds}, and defining some of the limitations.

        \item[Assessment:] 
        This RQ is addressed in an acceptable manner, with some weaknesses.
        Its strengths are research results from a literature review and the creation of a feasibility study of a \acrshort{ds}.
        However, the capabilities and limitations of such a \acrshort{ds} were not clearly outlined, and the challenges of extending it to a \acrshort{dt} were not detailed.
        For instance, the synchronization techniques were not delved into.
        Furthermore, suggestions for methods and technologies to develop \acrshortpl{dt} of modular robot systems were not investigated in detail.
        Therefore, the addressing of this RQ is ranked at level~2.
    \end{underlinedescription}

\section{Conclusion}
This section provides the concluding remarks, by providing answers on testing the hypothesis, and the impact of this thesis.

\subsection{Testing the Hypothesis}
We recall the hypothesis:
\begin{hypothesis}{Hypothesis}
    Solving relevant \textbf{barriers} related to the \textcolor{acqcolor}{\textbf{acquisition}} and \textcolor{intcolor}{\textbf{integration}} of modular robot systems in the industry advances the \textbf{\textcolor{depcolor}{deployment}} of such systems.
\end{hypothesis}

\begin{figure}
    \centering
    \includegraphics[width=0.65\textwidth]{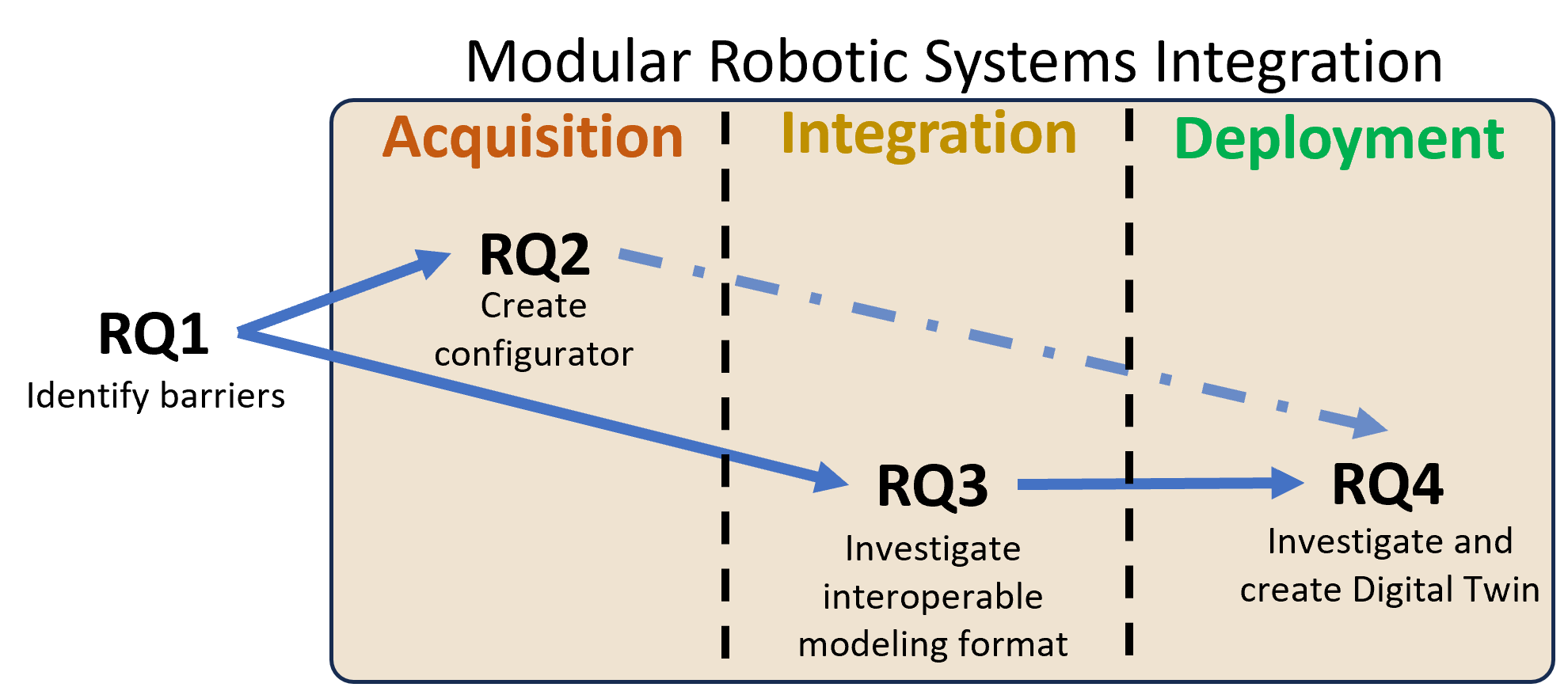}
    \caption{The found connections between the outputs of the Research Questions and how these can be used in addressing other RQs. The dashed line indicates this connection was initially assumed, but could not be proven.}
    \label{fig:rqs_relation_conclusion}
\end{figure}

This hypothesis is addressed through four Research Questions as shown in \cref{fig:rqs_relation_conclusion}, where \textbf{RQ1} focuses on identifying relevant barriers.
\textbf{RQ2} outputs a configurator that tackles specific challenges found in \textbf{RQ1}, related to lack of documentation, restrictions of vendor lock-in, and limited device reusability.
\textbf{RQ3} investigates the interoperable modeling format, \acrshort{urdf}, identifying its advantages and limitations, and tackles challenges related to restrictions of vendor lock-in found in \textbf{RQ1}.
\textbf{RQ4} investigates the concept of \acrshortpl{dt} for modular robot systems by creating a proof-of-concept and identifying the advantages and limitations.
\textbf{RQ2} solves barriers related to the acquisition of modular robot systems, while \textbf{RQ3} solves barriers related to integration of such systems.
\textbf{RQ4} investigates technologies for improving maintenance and operation of modular robot systems, advancing the deployment aspect of robotic systems integration.

The hypothesis assumes that the outputs from \textbf{RQ2} and \textbf{RQ3} can be used to advance \textbf{RQ4}.
As indicated in \cref{fig:rqs_relation_conclusion}, this thesis illustrates how the outputs from \textbf{RQ1} can be used to advance the acquisition and integration of modular robot systems, that is, these can be used in addressing \textbf{RQ2} and \textbf{RQ3}.
This thesis shows that the output from \textbf{RQ3}, which advances \acrshort{urdf}, can be used in addressing \textbf{RQ4}, which uses \acrshort{urdf}.
However, the relation between \textbf{RQ2} and \textbf{RQ4} could not be proven, as the developed configurator was not directly used in \textbf{RQ4}.
We found that configuration could be used in addressing \textbf{RQ4}, in environments where the \acrshort{dt} would need to self-adapt and change its devices.
This could be in the handling of objects, where the sizes, shapes, and textures differ, and the end-effector needs to be adapted to fit the properties of the object.
In such a case, a configurator could be used to choose an appropriate and compatible end-effector.
However, there are multiple aspects in this thesis that are neglected with regards to this:
\begin{itemize}
    \item Our configurator does not take into account the object size, shape, or texture, and would therefore need to be extended to do this.
    \item We have not considered the mechanical aspects, whether cables need to be changed, and how the end-effector would be substituted.
    \item Neither have we considered the programming of the substituted end-effector.
\end{itemize}

With these results we \textit{cannot} confirm the hypothesis that solving relevant barriers related to acquisition and integration advances the deployment of modular robot systems.
We have shown results that indicate confirmation of the hypothesis, however, we have not conducted enough research or provided tangible results that prove this.

\subsection{Impact of This Thesis}
The goal of this thesis is to advance areas within modular robotic systems integration by enabling digitalization.
This is achieved through comprehensive investigations and developments of proof-of-concepts.

We advance the acquisition of robot systems by creating a domain model for configuration and implementing a proof-of-concept configurator.
To our knowledge, no other configurator that addresses device compatibility of robot systems exists, as described in \cref{chapter:configurator}, and therefore our configurator presents a novel approach for automating the configuration of modular robot systems.
The advancement is accomplished by enabling a faster configuration process with limited need for expert interference, thus reducing overall costs.
However, this has not been tested in industry, which is a limitation in the impact of our configurator.

We advance the integration of robot systems by investigating the interoperable modeling format, \acrshort{urdf}, that can be used across various parts of robotic systems integration.
We identify the main capabilities of this format and its limitations, indicating areas for improvement.
The impact of this research has already been established through interest from other roboticists on \acrshort{ros} Discourse\footnote{\url{https://discourse.ros.org/t/urdf-improvements/30520}}, from Silicon Valley Robotics where the author was invited to present the findings of~\cite{Tola&2023d}.
Furthermore, the interest in advancing \acrshort{urdf} is apparent in the LinkedIn post\footnote{\url{https://www.linkedin.com/posts/petercorke_robotics-urdf-survey-activity-7005453864472649728-dNxb?utm_source=share&utm_medium=member_desktop}} for recruiting participants in~\cite{Tola&2023d} which received over~40.000 impressions and more than~400 reactions.
The results from~\cite{Tola&2023d} and~\cite{Tola&2023b} were discussed with a roboticist from Fraunhofer IPA\footnote{\url{https://www.ipa.fraunhofer.de/en.html}}, who has built upon our findings in their work on \acrshort{urdf}.

We advance the deployment of robot systems by investigating \acrshortpl{dt} and demonstrating how a proof-of-concept modular \acrshort{ds} of a robot system can be created.
We identify the potential capabilities of using \acrshortpl{dt} of modular robot systems, and the current limitations in developing them.
This enables future research to build upon our findings and extend the implementation to \acrshortpl{dt} of modular robot systems.

With this thesis we illustrate different ways that digitalization can be used in advancing the process of modular robotic systems integration.
However, we have not applied our research output to the industry directly, and is yet to be implemented in future work.

\section{Ethical Considerations}

Ensuring the ethics of our work is crucial as we consider the potential consequences of our research outputs.
Take, for instance, companies focused on developing helper robots for various aspects of our daily lives.
While these robots undoubtedly enhance our quality of life by assisting with numerous tasks, it is equally important to contemplate their potential negative impacts and devise strategies to mitigate any adverse effects associated with introducing new technologies to the world.
Imagine a robot that can bring us coffee and/or print and bring documents to us.
While this might seem convenient, it could lead to a reduction in our motivation to get out of bed or leave our desks, which can potentially lead to other problems, such as obesity. 
Therefore, it is crucial to contemplate how such potential societal consequences can be minimized. 

Our creations have a profound impact on the world, and we must therefore evaluate the ethical and moral implications of everything we create.
Often, our innovations address immediate problems but neglect to account for their broader, long-term effects.
Consider the development of cars, which at the time of creation were extremely convenient.
However, over the years we have gained more knowledge, and have understood the negative impacts of using petrol-based cars and how they affect the climate.
This has seemingly lead cities to create car-free Sundays, or ban them from their central city districts\footnote{\url{https://www.nytimes.com/2023/10/12/world/europe/stockholm-bans-gas-diesel-vehicles.html}}.
It is crucial to weigh the trade-offs between the benefits and drawbacks of new technologies and make decisions that benefit everyone.

While we move forward with developing new types of robots and technologies to tackle various challenges, we must not overlook the societal aspects.
Where are the laws and regulations regarding robots, and how are they evolving alongside technological advancements? We should prioritize finding ways to gradually integrate these solutions into society while ensuring their responsible and ethical implementation.

This thesis contributes to automation within the field of manufacturing, which has various jobs that can be dangerous and result in injuries.
Furthermore, fewer people are interested in taking on such types of jobs.
Although automating these processes helps us now, we do not know the potential effects of automating manufacturing lines in~30 years.
For instance, where do the worn out robot devices go? Are they recycled? 
How would these advancements affect society? Would they lead the world towards an even higher throw-away culture?
These are questions that must be considered when developing such systems. 
As far as I know, there are research projects looking into recycling robot devices.
However, I am not aware of any projects that look at the societal impact of more efficient automation of manufacturing lines.
Nevertheless, I believe it is suitable to create robots that can overtake or assist with some of these manufacturing tasks, but I do hope that the societal consequences will be considered and methods to mitigate the negative impacts will be investigated.

\section{Potential Directions}
Although this thesis tackles many different areas of robotic systems integration, it is a complex process with many different opportunities.
Therefore, there are still various areas that can be explored:

\begin{description}
    \item[Configuration:]
        The implementation of the configurator was not iterated upon with industry, and may therefore lack essential improvements, some of which are:
        \begin{itemize}
            \item In the case of handling objects, such as pick-and-place applications, taking into account the object size, shapes, and textures can be implemented.
            \item Extending the configurator to additional applications, such as welding, and the addition of more application requirements.
            \item Analyzing the usability of such a configurator for users with limited technical knowledge of robotic applications.
            \item Designing the configurator in a manner that is extensible, as different types of challenges of missing documentation may be found in the future.
            \item Investigating whether \acrshort{asp} is the best technology for implementing this configurator is important with regards to efficiency and scalability.
            \item Developing a user interface and integrating the configurator into a software platform is essential for industrial use.
        \end{itemize}

    \item[Unified Robot Description Format:]
            More detailed guidelines for this format can be developed, by creating comprehensive tutorials explaining how to use the format, when to use the different features of it, and provide an overview of the tools that support it.
            Furthermore, guidelines for how tools implement support for the format should be developed to improve the usability.
            Inspirations from \acrshort{fmi} can be drawn\footnote{\url{https://fmi-standard.org/tools/}}, which is also an interoperable format, for instance, a list of \acrshort{urdf} parsers, exporters, and simulators can be compiled.
            This thesis conducted research on \acrshort{urdf} in a descriptive manner.
            In the future it would make sense to research the format in an analytical manner to discover the root causes of its inherent issues.

    \item[Digital Twins:]
            The limitations of our work in this area are that only a \acrshort{ds} was implemented, and therefore we have not investigated the challenges related to creating \acrshortpl{dt} of modular robot systems, such as:
            \begin{itemize}
                \item Synchronization between the manufacturing element and the digital representation.
                \item Identifying methods for controlling the manufacturing element through the digital representation, and the limitations of it.
            \end{itemize}
            Furthermore, \acrshort{dt} platforms can be developed where all technologies required to create a \acrshort{dt} are found.
            Additionally, different types of services can be experimented with, and identification of which services are suitable to use for which types of systems can be performed.

    \item[Interoperable software tools:] 
            As other interoperable software tools, such as \acrshort{fmi} and \acrshort{uml} have shown inconsistency issues within their standards, it may be beneficial to analyze them and determine if there are general methods that can be followed to ensure consistency of such interoperable software tools, and relate these findings to \acrshort{urdf}.

    \item[Efforts for standardization of device interfaces:]
            Most of the areas of this thesis illustrated standardization challenges.
            Documenting robot devices is not standardized, meaning the structure of such documents differs depending on the \acrshort{oem}, and there can be missing information in such documents.
            Researching if it is feasible to create a standard for robot systems, similar to the \acrshort{pc} standard ISO 13066-1~\cite{ISO130661}, that describes interoperability of hardware, software, and their interfaces, and defines the responsibilities of the \acrshortpl{oem}, may decrease costs in robotic systems integration in the long run.

    \item[Ontologies:]
            This thesis has not considered ontologies when configuring and modeling robots, and there may be potential in examining the efforts in this area, as shown in~\cite{Schou&17}.

    \item[Virtual and augmented reality:]
            The potential of virtual and augmented reality is huge in robotic systems integration, and there are already tools that support designing a robot system and demonstrating it operating in a manufacturing line\footnote{For instance, SynergyXR, at \url{https://synergyxr.com/}.}.
            A customer can view the system in virtual or augmented reality before purchasing the system to get a better understanding of its functionality and benefits.
            Furthermore, augmented reality combined with \acrshortpl{dt} may help identify failures on a system and reduce time spent on failure classification.
            
\end{description}

\clearpage
\printbibliography[heading=bibintoc,title={References}]
\clearpage

\end{document}

